\documentclass[letterpaper]{article} 
\usepackage{aaai2026}  
\usepackage{times}  
\usepackage{helvet}  
\usepackage{courier}  
\usepackage[hyphens]{url}  
\usepackage{graphicx} 
\usepackage{natbib}  
\usepackage{caption} 
\usepackage{algorithm}
\usepackage{algorithmic}

\usepackage{newfloat}
\usepackage{listings}
\DeclareCaptionStyle{ruled}{labelfont=normalfont,labelsep=colon,strut=off} 
\floatstyle{ruled}
\newfloat{listing}{tb}{lst}{}
\floatname{listing}{Listing}
\usepackage{microtype}
\usepackage{url}
\usepackage{booktabs}

\usepackage{inconsolata}
\usepackage{amssymb}
\usepackage{amsmath}
\usepackage{booktabs}
\usepackage{multirow}
\usepackage{graphicx}
\usepackage{makecell}
\NewDocumentCommand{\asterisk}{}{\makebox[0pt][l]{$^*$}}

\usepackage{lineno}
\definecolor{darkblue}{rgb}{0, 0, 0.5}

\title{Identifying and Analyzing Performance-Critical Tokens\\ in Large Language Models}

\author {
    Yu Bai\textsuperscript{\rm 1}\thanks{~Work done during the internship at Mila. The full paper, including all technical appendices, is available at \textit{arXiv:2401.11323}.},
    Heyan Huang\textsuperscript{\rm 1,2},
    Cesare Spinoso-Di Piano\textsuperscript{\rm 3,4}, \\
    Sanxing Chen\textsuperscript{\rm 6},
    Marc-Antoine Rondeau\textsuperscript{\rm 3},
    Yang Gao\textsuperscript{\rm 1}\thanks{~Corresponding author.},
    Jackie Chi Kit Cheung\textsuperscript{\rm 3,4,5},
}
\affiliations {
    \textsuperscript{\rm 1} Beijing Institute of Technology, Beijing, China\\
    \textsuperscript{\rm 2} Southeast Academy of Information Technology, Fujian, China\\
    \textsuperscript{\rm 3} Mila - Quebec Artificial Intelligence Institute\\
    \textsuperscript{\rm 4} McGill University\\
    \textsuperscript{\rm 5} Canada CIFAR AI Chair\\
    \textsuperscript{\rm 6} Duke University\\
}

\begin{document}

\maketitle

\begin{abstract}
In-context learning (ICL) has emerged as an effective solution for few-shot learning with large language models (LLMs).
However, how LLMs leverage demonstrations to specify a task and learn a corresponding computational function through ICL is underexplored. 
Drawing from the way humans learn from content-label mappings in demonstrations, we categorize the tokens in an ICL prompt into content, stopword, and template tokens.
Our goal is to identify the types of tokens whose representations directly influence LLM's performance, a property we refer to as being \textit{performance-critical}. 
By ablating representations from the attention of the test example,
we find that the representations of informative content tokens have less influence on performance compared to template and stopword tokens, which contrasts with the human attention to informative words. 
We give evidence that the representations of performance-critical tokens aggregate information from the content tokens.
Moreover, we demonstrate experimentally that lexical meaning, repetition, and structural cues are the main distinguishing characteristics of these tokens. 
Our work sheds light on how LLMs learn to perform tasks from demonstrations and 
deepens our understanding of the roles different types of tokens play in LLMs.
\end{abstract}


\section{Introduction}
\label{sec:intro}


In-context learning (ICL) has become a popular technique employed with large language models~(LLMs) \citep{brown2020language}. However, ICL has been shown to be unstable in that slight changes to the in-context prompts~(e.g., reordering of demonstrations) can lead to substantial differences in performance~\citep{lu-etal-2022-fantastically, zhang2022active}.
This circumstance is difficult to control due to a lack of understanding of the model's working mechanisms, leaving us uncertain about the exact process by which LLMs learn to infer a task specification from demonstrations and produce a computation function to implement that task specification. 
Previous papers explored this issue, focusing on specific aspects such as the label space~\citep{min2022rethinking} and the hidden states of the last prompt token~\citep{hendel2023incontext, todd2023function}, but have been limited in scope.

\begin{figure}[t]
\centering
\includegraphics[width=0.90\linewidth]{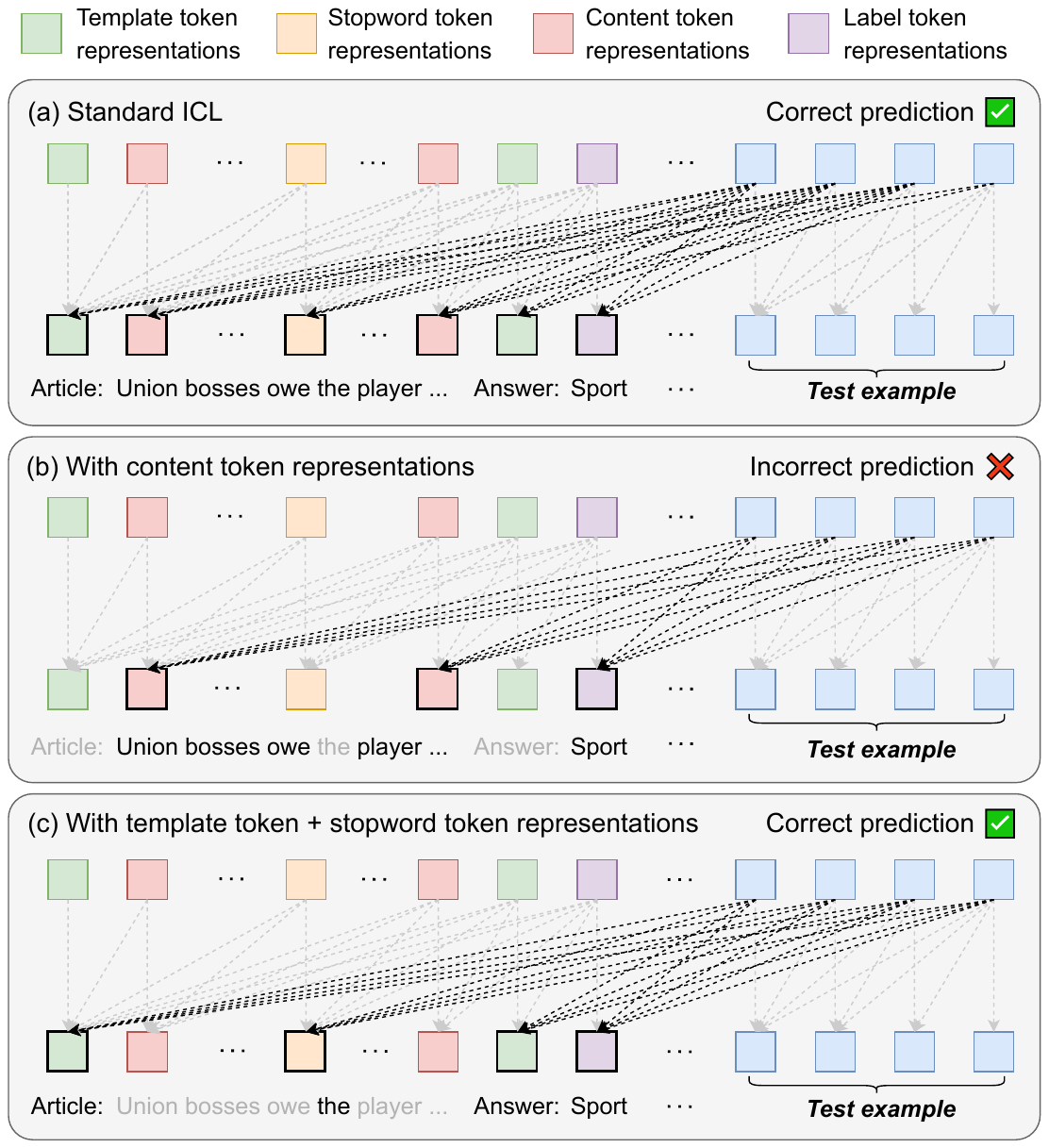} 
\caption{An illustration of 4-way text classification on AGNews with different parts of its 4-shot ICL demonstrations masked with respect to the attention of the test example. Specifically, (b) masking the representations of what we call the template and stopword tokens from~the attention of the test example leads to a significant drop in performance while (c) masking representations of the content tokens leaves the performance relatively unchanged. The dashed lines represent the attention between every pair of tokens while those from the test example to the ICL prompt are unshaded.}
\label{fig:intro}
\end{figure}

In this work, we conduct a comprehensive study on how LLMs extract information that is valuable for improving task performance from demonstrations. 
Drawing from the way humans learn through content-label mappings in demonstrations, we categorize the tokens in an ICL prompt into content, stopword~\citep{wilbur1992automatic}, and template tokens, with the latter two typically ignored by humans due to their uninformative nature~\citep{lenartowicz2014neurophysiological}. 
We use content tokens as a category because they ``are fixated 
(by human) about 85\% of the time''~\citep{rayner1998eye}.
With these categories, we ablate the representations of different token types from the attention of ICL test examples, masking partial information during the model's task-solving process, as shown in Figure~\ref{fig:intro}. 
This ablation is intended to identify the types of tokens whose representations LLMs \textbf{directly} depend on to achieve high-level performance, thereby explaining how LLMs learn from demonstrations. 
We refer to these tokens that are critical for performance as \textbf{performance-critical tokens}.

We show that template and stopword tokens (e.g., ``Answer:'') are the most prone to be performance-critical tokens.
In contrast, content tokens (e.g., ``Union'', etc.) have a negligible impact on performance  when their representations are eliminated from the attention of the test examples.
This finding is counterintuitive since the original text of template and stopword tokens inherently do not possess any information found in the demonstrations.
To explain this, we study the relationship among different types of tokens through ablation experiments that cut off the information flow between different kinds of tokens. 
We show that content tokens are \textbf{indirectly} leveraged by LLMs during ICL through aggregating their information into the representations of performance-critical tokens, raising questions about how LLMs use these tokens in general text processing and reasoning.

Beyond identifying performance-critical tokens, we analyze them to better understand how they are leveraged by LLMs.
Specifically, we investigate the characteristics which differentiate them from other tokens.
We find the following three distinguishing characteristics:
the \textbf{lexical meaning} of tokens as it relates to the task being solved, the \textbf{repetition} of tokens throughout the prompt, and the \textbf{structural cues} which the tokens provide to the prompt.
Our findings indicate that the lexical meaning, repetition, and structural cues of performance-critical tokens contribute to task performance across all model sizes, suggesting that they contribute to the property of tokens being performance-critical.

Our work reveals that we can identify and characterize the types of tokens whose representations are the most important in directly maintaining ICL task performance.
Our results suggest that previous claims about ICL should be more nuanced, in that representations of tokens beyond label words~\citep{wang2023label} may also directly impact the task performance.
We investigate the characteristics of lexical meaning, repetition, and structural cues related to performance-critical tokens which allow us to partially explain their importance 
and help us better understand how to avoid performance instability while using ICL. 
Additionally, our findings reinforce that we cannot assume LLMs and humans solve tasks in similar ways, emphasizing the need to develop better explanations of LLM behaviors.
Overall, our research deepens the understanding of the roles different types of tokens play in LLMs, pointing to future work that explores how specific token representations can be leveraged for particular purposes (e.g., storing compressed information). Code and data are released at \url{https://github.com/ybai-nlp/PCT_ICL}.

\section{Related Work}

\paragraph{Working mechanisms of ICL}
Since the proposal of in-context learning~\citep{brown2020language}, its working mechanisms have been extensively studied by the research community~\citep{min2022rethinking, liu2021makes, bhattamishra2023understanding, zhou2023mystery}. \citet{min2022rethinking} suggest that demonstrations primarily provide the label space, the distribution of the input text, and the format of the sequence for the test example. They argue that the precise ground truth labels do not have significant importance. In contrast, \citet{yoo2022groundtruth} propose a differing view, stating that the impact of the ground truth labels depends on the experimental configuration. \citet{mao2024data} analyze in-context learning from the perspective of data generation. The perspective of supported training data is also leveraged to analyze ICL \citep{han2023understanding}. 
\citet{zhao2024unveiling} propose to use coordinate systems to understand the working mechanism of in-context learning.
\citet{zhou2023mystery} propose a comprehensive survey on the interpretation and analysis of in-context learning. 
\citet{lin2024dual} model the dual modes of in-context learning (task retrieval vs. task learning) and phenomena like early performance drops. 
\citet{lilong} introduce a long-context benchmark revealing degradation with very long demonstrations.

Our work investigates the working of ICL in LLMs at inference time, demonstrating that certain specific tokens are more likely to possess representations that could affect the processing of final test samples, improving the performance. 

\paragraph{Function vectors of in-context learning}
\citet{todd2023function} and \citet{hendel2023incontext} provide evidence that function vectors store information used to solve a task in ICL. They probe and extract the hidden representations of the final tokens in the prompt, which can then be added to or replace the corresponding vectors in a zero-shot example, yielding results comparable to using all demonstrations as context. \citet{liu2023incontext} propose using an in-context vector to represent the target task and apply feature shifting to query examples. They feed each input and its target into an LLM, concatenate the latent states, and apply PCA to derive a vector aligned with the task. Finally, \citet{wang2023label} propose that label words in demonstrations act as information anchors, aggregating information from previous examples. This suggests label tokens may satisfy our definition of performance-critical tokens.
\citet{yang2025task} show that compact task vectors naturally emerge and can be enhanced with an auxiliary loss. \citet{tikhonov2025one} find that multiple vectors are often needed for complex tasks. \citet{dong2025understanding} propose the Linear Combination Conjecture, explaining vector formation but noting limitations for high-rank tasks. \citet{jiangunlocking} introduce function vectors to mitigate catastrophic forgetting in continual instruction tuning.

All these previous studies either solely focus on a single token (i.e., the last prediction prompt token or label token) of the ICL prompt or treat the entire demonstration as a single unit, neglecting the other tokens within it. Our research focuses on all the tokens in the prompt and reveals that there are additional tokens with specific characteristics whose representations significantly affect the final ICL performance.

\section{Preliminaries}
\paragraph{Notation}
\label{sec:notation}
In in-context learning (ICL), task demonstrations (e.g., input-output pairs) are leveraged to construct a structured prompt that guides the model in predicting the final answer. 
Formally, the structural prompt consists of the following components: 
the instruction ${\mathbf I}$, the templates ${\mathbf T}^{\rm in}$, ${\mathbf T}^{\rm out}$, and the demonstrations ${\mathbf D}^{\rm in}_i$, ${\mathbf D}^{\rm out}_i$, where $i$ denotes the $i$\textsuperscript{th} demonstration while ${\rm in}$ and ${\rm out}$ refer to the input text and output labels, respectively. These prompt components are concatenated to form the ICL prompt, $P$, as shown in Table~\ref{tab:ICLexample}. During inference, the templated version of the test example without its answer, ${\mathbf T}^{\rm in} \cdot {\mathbf D}^{\rm in}_{\text{test}} \cdot  {\mathbf T}^{\rm out}$, is appended to the ICL prompt and then sent to the large language model to predict the corresponding answer. We use $\cdot$ to denote the concatenation of token sequences.

\begin{table*}[t]
  \centering
  \resizebox{0.8\linewidth}{!}{
    \begin{tabular}{lp{18cm}}
    \toprule
    \makecell[l]{Component notation} & \multicolumn{1}{c}{Component example} \\
    \midrule     
    ${\mathbf I}$ &    Classify the news articles into the categories of World, Sports, Business, and Technology.\textbackslash n\textbackslash n \\
    ${\mathbf T}^{\rm in}$ & Article: \{${\mathbf D}^{\rm in}$\}\textbackslash n \\
    ${\mathbf T}^{\rm out}$ & Answer: \{${\mathbf D}^{\rm out}$\}\textbackslash n\textbackslash n \\
    ${\mathbf D}^{\rm in}_1$ & Radio veteran Karmazin joins Sirius. Sirius Satellite Radio Inc. named former Viacom Inc. president Mel...\\
    ${\mathbf D}^{\rm out}_1$ & Business \\
    ${\mathbf D}^{\rm in}_2$ & Numbers point to NY. NEW YORK - The New York Yankees can achieve two milestones with one more victory...\\
    ${\mathbf D}^{\rm out}_2$ & Sports \\
    \midrule
    \multirow{7}*{\makecell[l]{ICL\\ Prompt}} & Classify the news articles into the categories of World, Sports, Business, and Technology.
    \\
    \\
    & Article: Radio veteran Karmazin joins Sirius. Sirius Satellite Radio Inc. named former Viacom Inc. president Mel...\\
    & Answer: Business \\

    \\
     & Article: Numbers point to NY. NEW YORK - The New York Yankees can achieve two milestones with one more victory...\\
    & Answer: Sports \\
    \bottomrule
    \end{tabular}
    }

\caption{An example of the components of a 2-shot ICL prompt in the AGNews dataset.}
    \label{tab:ICLexample}
\end{table*}

\paragraph{Definition of performance-critical token}
\label{sec:definition}
In defining the performance-critical tokens, we measure the performance variation before and after incorporating the representations of specific tokens into the attention scope of the test example. Specifically, we define performance-critical tokens as the tokens that lead to both a noticeable performance improvement when their representations are included in the attention of test examples and a noticeable performance degradation when they are excluded. 

Formally, let $M$ be a decoder-only transformer-based large language model (LLM) and $D$ be a classification dataset. 
%
We define $H_P$ as the set of representations of each token in the ICL prompt $P$ and $H_{\text{test}}$ as the set of representations of the test demonstration which is appended to $P$ for prediction (i.e., ${\mathbf T}^{\rm in} \cdot {\mathbf D}^{\rm in}_{\text{test}} \cdot  {\mathbf T}^{\rm out}$). In addition, we let $H_{\text{attend}} \subseteq H_P$ be some set of representations which $M$ may attend to from $H_{\text{test}}$ at inference time while performing ICL. For instance, $H_{\text{attend}} := H_{\mathbf I}$ would imply that, when $M$ is predicting the label of the test demonstration, the attention from the test example is restricted to the prompt's instruction token representations.

Then, in order to provide a practical definition for the performance-critical tokens, we let ${\mathbf{Acc}}(M, D, H_{\text{attend}})$ be the accuracy achieved by a LLM $M$ when performing ICL on the classification dataset $D$ where the only representations which the test example may attend to at inference time are $H_{\text{attend}}$. Given a partition $\mathcal{P}$ of $H_\text{P}$, we say that a set of tokens $H^* \in \mathcal{P}$ is \emph{performance-critical} if
\begin{equation}
    \small
    \begin{aligned}
    {\mathbf{Acc}}(M, D, H^*) &\gg {\mathbf{Acc}}(M, D, \emptyset) \quad \text{and} \\
    {\mathbf{Acc}}(M, D, H_P) &\gg {\mathbf{Acc}}(M, D, H_P - H^*)
    \end{aligned}
\end{equation}

We note that examining the possibility of each token being performance-critical (i.e., $|H^*| = 1$) in an ICL prompt would be computationally intractable. We instead categorize all the tokens based on the role they play in the prompt and identify which types of tokens are more likely to be performance-critical in Section~\ref{sec:identification}.

\section{Experimental settings}
\label{sec:experiment_setting}

\textbf{Datasets. }
We consider the most widely used text classification datasets used by previous studies~\citep{pmlr-v139-zhao21c}. For topic classification, we use the 4-way and 14-way datasets AGNews and DBPedia~\citep{NIPS2015_250cf8b5}. For textual entailment, we use the 3-way CB~\citep{de2019commitmentbank} and 2-way RTE dataset~\citep{dagan2005pascal}. We also use SST2~\citep{socher2013recursive} and TREC~\citep{voorhees2000building} for sentiment and question classification tasks. Besides these classification tasks, we also present results in \textbf{machine translation} and \textbf{question answering} tasks to show that our findings can also be extended to text generation tasks.
Results and analyses of these generation tasks are attached to Appendix. 

 \textbf{Evaluation. }
For each dataset, we randomly select 4 training demonstrations from the training set using 15 seeds, limited by the computational cost of LLM inference. For testing, we evaluate each setting on 500 randomly selected test examples. We show this sample size is sufficient by comparing results with 500 test examples and the full dataset using OpenLlama 3B and Llama 7B models, as shown in Appendix~\ref{app:more_examples}. Instruction prompt ${\mathbf I}$ is retained in all ablations, as it is essential for model performance~\citep{yin-etal-2023-read}. We use a fixed ${\mathbf I}$ for all main results of different models, with additional experiments with different ${\mathbf I}$ provided in Appendix~\ref{app:different_instruction}, demonstrating that changing ${\mathbf I}$ does not affect our main findings.

 \textbf{LLMs. }
We utilize the 7B, 13B, and 33B Llama models~\citep{touvron2023llama} and a 3B OpenLlama model. We also report additional results with Llama 2 7B, Llama 2 13B~\citep{touvron2023llama2}, Mistral 7B~\citep{jiang2023mistral7b}, and Gemma 3 4B~\citep{team2025gemma} models.
Models after supervised fine-tuning are tested in Appendix~\ref{app:sft_models}. All experiments are conducted on a single A100 80G GPU. For the 13B and 33B models, we apply 8-bit quantization to fit them into a single GPU. 


\section{Performance-Critical Token Detection}
\label{sec:identification}
In this section, we aim to identify the performance-critical tokens in the ICL prompt. 
We first structurally categorize all the tokens in the prompt into three types: template, stopword, and content tokens.
Then, we provide supporting evidence from the view of task performance to show that the template and stopword tokens are the most prone to be performance-critical tokens. Finally, we demonstrate that the information of content tokens serve to indirectly contribute to the performance by being propagated into the representations of the performance-critical tokens by LLMs.


\subsection{Token types}
We categorize ICL tokens based on the structure of the ICL prompt, following our notation in Table~\ref{tab:ICLexample}. Firstly, we find it natural to categorize tokens based on the structure of ICL prompts where the tokens from the demonstration examples ${\mathbf D}^{\rm in}$ and the labels ${\mathbf D}^{\rm out}$ are separated by template tokens from ${\mathbf T}^{\rm in}$ and ${\mathbf T}^{\rm out}$. 
Second, ${\mathbf D}^{\rm in}$ can be subdivided into content and stopword tokens, with the latter typically providing less useful information and often being ignored \citep{rayner1998eye} when humans use analogy to learn specific tasks.
Guided by these intuitions, we categorize all the tokens in the ICL prompt into template tokens, stopword tokens, and content tokens as follows:


\textbf{Template tokens (\textsc{temp})}: In defining template tokens, we include all the tokens which serve as templates for the ICL prompt. This 
includes the tokens in ${\mathbf T}^{\rm in}$ and ${\mathbf T}^{\rm out}$.

\textbf{Stopword tokens (\textsc{stop})}: In defining stopword tokens, we include punctuation and conjunction words, such as  [\textbf{,}], [\textbf{.}], etc., in the prompt. We use the stopword tokens appearing in the instructions\footnote{The stopword token list used in the main experiments and the ablation results with the complete NLTK~\citep{loper2002nltk} stopwords list are shown in Appendix~\ref{app:stopword}.}.

\textbf{Content tokens (\textsc{cont})}: In defining content tokens, we include all the tokens from ${\mathbf D}^{\rm in}$ except for the ones that are already stopword tokens. We use the term ``content tokens'' as they convey the meaningful information found in the demonstrations.

Researchers might typically expect content tokens to be critical, as they account for most of the information in demonstrations. However, in our experiments, we find that the representations of template and stopword tokens have the greatest direct impact on performance.

\begin{figure}[t]
\centering
\includegraphics[width=0.9\linewidth]{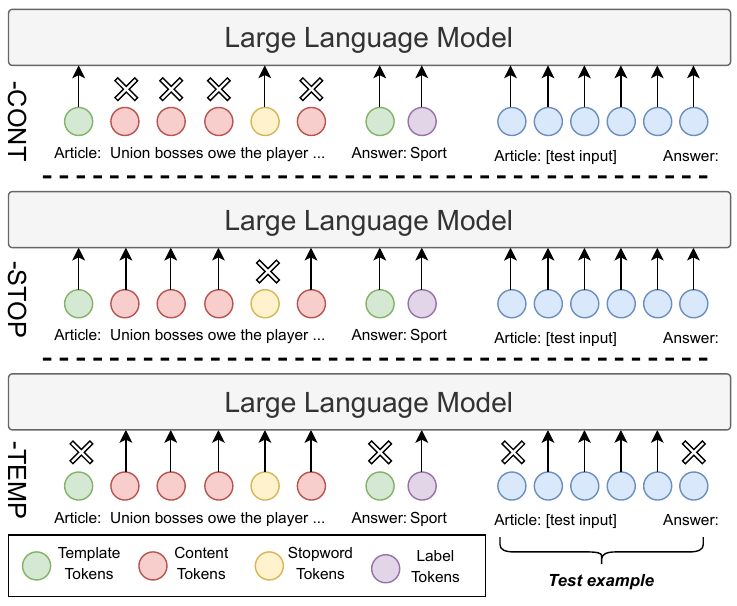} 
\caption{An illustrative example of the token-level ablation we use to analyze the working mechanism of performance-critical tokens.}
\label{fig:ablationmethod}
\end{figure}

\subsection{Ablation on token types}
\label{sec:ablation_all}
To determine which token types are more likely to be performance-critical tokens whose representations directly affect the final performance significantly, we design two experiments which ablate representations or tokens based on token types. The first involves keeping and masking representations of different token types from the attention of the test example. The second involves dropping the various kinds of tokens 
from the ICL prompt. 
The main purpose of the first experiment is to identify the performance-critical tokens defined in Section~\ref{sec:definition}, while the second experiment aims to cut off the information propagation of different types of tokens to 
further explore the functioning of performance-critical tokens.
Illustrations of these two methods which we refer to as representation-level and token-level ablations are shown in Figure~\ref{fig:intro} and Figure~\ref{fig:ablationmethod}. Examples for the representation-level ablation is provided in Appendix~\ref{app:rla_example}. 

\subsubsection{Representation-level ablation}   
\label{sec:ablation_representation}

Our first ablation stems from the intuition that if LLMs rely on the representations of specific token types to achieve high-level performance, they should perform adequately with only these representations. Performance should drop significantly if these tokens are removed from the test example’s attention. Hence, we first pass the entire ICL prompt to the LLM and then restrict the test example’s attention to representations of a particular token type (or types)\footnote{Since ${\mathbf D}^{\rm out}$ tokens have been shown to significantly impact performance~\citep{wang2023label}, we always preserve the attention on the representations of the ${\mathbf D}^{\rm out}$ tokens.} during its solving of the task. 
We compute task performances with every possible ablation combination, removing the representations of one (e.g., Standard ICL $-$ \textsc{temp}) or two token types~(e.g., Zero-shot $+$ \textsc{cont}\footnote{Removing two token types from Standard ICL is equivalent to adding the other type 
to Zero-shot.}) from the attention of the test example. All the task performances and the averaged relative performance changes are reported, shown in Table~\ref{tab:repablation_morellms},~\ref{tab:repablation_final1}, and~\ref{tab:repablation_gemma}.
Results for Llama 7B and 13B models are shown in Appendix~\ref{app:more_models}.
An illustration of this set of experiments is shown in Figure~\ref{fig:intro}.

\begin{table}[t]

  \centering
  \resizebox{0.9\linewidth}{!}{
    \begin{tabular}{llrrrrrr|l}
    \toprule
     Models & Setting & AGNews & SST2  &  TREC  & DBPedia & RTE & CB & $\triangle$Avg. \\
    \midrule
   \multirow{4}*{\makecell[l]{Llama 2\\ 7B}} 
      & Zero-shot  & 50.2 & 50.4 & 57.2 & 6.4 & 51.6 & 0.0 &  36.0\\
    &  \quad$+$ \textsc{cont} & 0.9 & 61.0 & 50.6 & 12.9 & 48.7 & 53.2 & \quad$+$1.9\\
   &  \quad$+$ \textsc{stop}  & 49.0 & 78.1 & 54.4 & 61.6 & \textbf{65.3} & 47.9 & \quad$+$23.4 \\
   &  \quad$+$ \textsc{temp} & \textbf{81.1} & \textbf{82.6} & \textbf{55.2} & \textbf{65.5} & 63.9 & \textbf{55.4} & \textbf{\quad$+$31.3} \\


   \midrule
    \multirow{4}*{\makecell[l]{Llama 2\\ 13B}}  
     & Zero-shot   & 56.2 & 90.8 & 49.0 & 7.6 & 70.0 & 46.4  & 53.3 \\
   
   & \quad$+$ \textsc{cont} & 0.5 & 56.0 & 61.4 & 0.0 & 62.6 & 54.6 & \quad$-$14.1 \\
   & \quad$+$ \textsc{stop}  & 47.2 & 76.8 & \textbf{65.2} & 65.3 & 66.5 & \textbf{56.8} & \quad$+$9.6 \\
   & \quad$+$ \textsc{temp} & \textbf{78.2} & \textbf{93.7} & 62.4 & \textbf{70.4} & \textbf{71.9} & 52.1&  \textbf{\quad$+$18.1}   \\

   \midrule
    \multirow{4}*{\makecell[l]{Mistral\\ 7B}}  
    
     & Zero-shot  & 77.8 & 84.4 & 73.0 & 57.8 & 1.8 & 62.1 & 59.5  \\
    &  \quad$+$ \textsc{cont} & 43.3 & 52.0 & 66.6 & 10.1 & 64.3 & 60.7 & \quad$-$10.0 \\
   &  \quad$+$ \textsc{stop}  & 78.9 & 92.5 & \textbf{71.6} & 81.4 & 69.9 & \textbf{72.9} &   \quad$+$18.4 \\
   &  \quad$+$ \textsc{temp}  & \textbf{81.7} & \textbf{95.9} & 63.9 & \textbf{83.3} & \textbf{77.9} & 72.5&  \textbf{\quad$+$19.7} \\

    \midrule
   \multirow{4}*{\makecell[l]{Llama 2\\ 7B}} 
 
     & Standard ICL   & 85.0 & 93.2 & 58.3 & 66.7 & 66.3 & 55.0  & 70.7\\
   
   & \quad$-$ \textsc{cont}& 82.4 & 85.5 & 54.3 & 64.2 & 59.6 & 55.7  &\quad$-$3.8   \\
   & \quad$-$ \textsc{stop}  & 84.8 & 88.0 & 51.7 & 65.7 & 65.8 & \underline{53.2} &\quad$-$2.5 \\
   & \quad$-$ \textsc{temp}  & \underline{0.9} & \underline{61.0} & \underline{50.6} & \underline{12.9} & \underline{48.5} & 53.6 &\quad\underline{$-$32.8} \\

   \midrule
    \multirow{4}*{\makecell[l]{Llama 2\\ 13B}}  
     & Standard ICL  & 82.8 & 94.9 & 62.8 & 74.6 & 71.2 & 55.4  &  73.6 \\
    &  \quad$-$ \textsc{cont}& 79.0 & 94.1 & 62.7 & 72.4 & 72.1 & 53.6  &\quad$-$1.3  \\
   &  \quad$-$ \textsc{stop}  & 80.1 & 89.4 & 61.5 & 74.1 & 69.6 & \underline{49.3} &\quad$-$2.9 \\
   &  \quad$-$ \textsc{temp} & \underline{0.5} & \underline{56.0} & \underline{61.4} & \underline{0.0} & \underline{68.2} & 54.3 & \quad\underline{$-$33.5} \\

   \midrule
    \multirow{4}*{\makecell[l]{Mistral\\ 7B}}  
    
     & Standard ICL   & 82.2 & 97.0 & 67.4 & 82.4 & 73.6 & 78.6 &  80.2 \\
    &  \quad$-$ \textsc{cont}  & 81.8 & 96.2 & \underline{64.4} & 83.4 & 78.9 & 72.9 &\quad$-$0.6 \\
   &  \quad$-$ \textsc{stop} & 81.3 & 97.0 & 66.5 & 80.5 & 75.5 & 75.7   &\quad$-$0.8 \\
   &  \quad$-$ \textsc{temp} & \underline{78.6} & \underline{52.0} & 66.6& \underline{10.1} & \underline{67.1} & \underline{69.6} &\quad\underline{$-$22.8}  \\

    \bottomrule
    \end{tabular}
    }

\caption{The accuracy results of the representation-level ablation study using Llama 2 and Mistral models where, for example, $+$\textsc{temp} refers to allowing attention only to template tokens and  $-$\textsc{temp} refers to allowing attention only to content and stopword tokens. All values are presented as percentages. Results are averaged over 5 different random seeds. The best results are in bold and the results showing the greatest decrease during the ablation are underlined.}
    \label{tab:repablation_morellms}
\end{table}

\begin{table}[t]
  \centering
  \resizebox{\linewidth}{!}{
    \begin{tabular}{llrrrrrr|l}
    \toprule
     Models & Setting & AGNews & SST2  &  TREC  & DBPedia & RTE & CB & $\triangle$Avg. \\
    \midrule
   \multirow{4}*{\makecell[l]{OpenLlama\\ 3B}} 
 
     & Zero-shot   &22.0 & 20.0 & 23.6 & 5.4 & 44.4 & 1.8& 19.5\\
   
   & \quad $+$ \textsc{cont}& 26.2 & 52.1 & 30.1 & 7.4 & 51.9 & 37.9 & \quad$+$14.8 \\
   & \quad $+$ \textsc{stop} &36.7 & 82.9 &\textbf{32.0}\asterisk & 52.4 & \textbf{58.8}& \textbf{56.2} & \quad$+$33.7\\
   & \quad $+$ \textsc{temp} &\textbf{56.5} &\textbf{86.7}& 27.1 &\textbf{62.2} & 56.4 & 52.3 & \quad\textbf{$+$37.4} \\


    

   \midrule
    \multirow{4}*{\makecell[l]{Llama\\ 33B}}  
     & Zero-shot  &  70.2 & 88.6 & 60.6 & 30.2 & 58.1 & 19.6 & 54.6 \\
    & \quad $+$ \textsc{cont}&  24.4 & 61.7 & 62.1 & 10.5 & 65.2 & 63.6 & \quad$-$6.7 \\
   &\quad $+$ \textsc{stop}  &  72.9 & 92.7 & \textbf{66.7}\asterisk & 69.1 & 69.6 & 63.0 & \quad$+$17.7\\
   &\quad $+$ \textsc{temp} &   \textbf{80.5} & \textbf{95.2} & 65.2 & \textbf{75.2} & \textbf{79.0} & \textbf{80.0}  & \quad\textbf{$+$24.6}\\

    \midrule
    
   \multirow{4}*{\makecell[l]{OpenLlama \\ 3B}} 
 
     & Standard ICL &  63.7 & 91.2 & 21.9 & 61.9 & 57.4 & 52.0 & 58.0 \\
   
   & \quad$-$ \textsc{cont}&58.2 & 86.9 & \underline{27.6} & 61.9 & 56.5 & 51.7 & \quad$-$0.9  \\
   & \quad$-$ \textsc{stop} &51.8 & 78.9 & 28.8 & 30.3 & 53.6 & 45.2 & \quad$-$9.9  \\
   & \quad$-$ \textsc{temp} & \underline{26.2} & \underline{52.1} & 30.1 & \underline{7.4} & \underline{51.9} & \underline{37.9}&\quad\underline{$-$23.8}  \\
 
   


   \midrule
    \multirow{4}*{\makecell[l]{Llama\\ 33B}}  
    
     & Standard ICL & 85.0 & 96.5 & 68.1 & 78.4 & 78.5 & 83.3 & 81.6  \\
    &  \quad$-$ \textsc{cont} & 82.3 & 95.4 & 64.9 & 76.1 & 80.4 & 82.0&\quad$-$1.5 \\
   &  \quad$-$ \textsc{stop} & 84.8 & 94.9 & 62.1 & 77.3 & 70.5 & 74.4& \quad$-$4.3 \\
   &  \quad$-$ \textsc{temp}  & \underline{24.4} & \underline{61.7} & \underline{60.6} & \underline{10.5} & \underline{67.7} & \underline{68.5}& \quad\underline{$-$32.7}\\
    \bottomrule
    \end{tabular}
    }

    \caption{The accuracy results of the representation-level ablation study using Llama models where, for example, $+$\textsc{temp} refers to allowing attention only to template tokens and  $-$\textsc{temp} refers to allowing attention only to content and stopword tokens. All values are presented as percentages. Except where noted with $^*$, all test
statistics reported correspond to p-values $<$ 0.05. The best results are in bold.}
    \label{tab:repablation_final1}
\end{table}

Overall, these results demonstrate that \textbf{template and stopword tokens} are more likely to be performance-critical tokens than content tokens, conforming to our definition in Section~\ref{sec:definition}.
On the one hand, template token representations 
directly influence the task performance in ICL, 
achieving an average performance 39.8\% higher than the zero-shot baseline by only utilizing these representations at inference time.
If the representations of stopword tokens 
are further included (i.e., Standard ICL$-$\textsc{cont}), the performance is nearly equivalent to that of the Standard ICL. 
In contrast, content token representations only bring an average improvement of 10.7\%.
On the other hand, the performance decreases the most with Standard ICL$-$\textsc{temp}, highlighting the direct significance of template tokens again\footnote{Both \textsc{stop} and \textsc{temp} include the ``\textbackslash n'' token; we mask the attention to the ``\textbackslash n'' token as long as one of them is ablated in this set of experiments. Analyses about this experimental setting are shown in Appendix~\ref{app:nxtline_tokens}.}.
Furthermore, considering token counts for each type, as shown in Appendix~\ref{app:representation_number}, content tokens vastly outnumber the other two types. Hence, the averaged impacts of the template and stopword tokens further suggest that they are more prone to be performance-critical. 

\begin{table}[t]

  \centering
  \resizebox{0.6\linewidth}{!}{
    \begin{tabular}{l|rrr|r}
    \toprule
     Setting & AGNews & SST2   & DBPedia & $\triangle$Avg. \\
    \midrule

       Zero-shot  & -& - & - & - \\
      \quad$+$ \textsc{cont} & 1.62 & 	50.08& 2.24 &	17.98 \\
    \quad$+$ \textsc{stop}  & 77.32	&	87.22	& 76.72 & 80.42 \\
    \quad$+$ \textsc{temp} & \textbf{83.66}  & \textbf{92.74}& \textbf{80.50}  & \textbf{85.63} \\

    
    \midrule
 
      Standard ICL   & -& - & - & -\\
   
    \quad$-$ \textsc{cont}& 83.66 & 92.74 & 80.50 & 85.63 \\
    \quad$-$ \textsc{stop}  & 84.22  & 94.14& 79.50 & 85.95 \\
    \quad$-$ \textsc{temp}  & \underline{76.82}	& \underline{90.78} & \underline{76.56} &	\underline{81.39} \\

    \bottomrule
    \end{tabular}
    }

\caption{The accuracy results of the representation-level ablation study using Gemma-3 4B models. All values are presented as percentages. Results are averaged over 5 different random seeds.}
    \label{tab:repablation_gemma}
\end{table}

Rare exceptional cases appear when performance is relatively poor with Standard ICL (e.g., OpenLlama 3B in TREC).
In some cases, masking the representations of the content tokens brings even better performance than the Standard ICL method, which is possibly due to the elimination of noisy information in the demonstration content.
Another interesting observation is that the performance results of Standard ICL$-$\textsc{stop} and Standard ICL$-$\textsc{cont} where the attention to the content and stopword tokens is ablated respectively are similar, with an average difference of only 5.4\%. 
This indicates that the representation of stopword tokens may contain overlapping information with their preceding content tokens. We believe that this could enable LLMs to model long sequences without significant architectural changes (e.g., using stopword representations as synthesis checkpoints) and leave the verification of this hypothesis to future work.

\begin{figure}[t]
    \centering
    \includegraphics[width=0.7\linewidth]{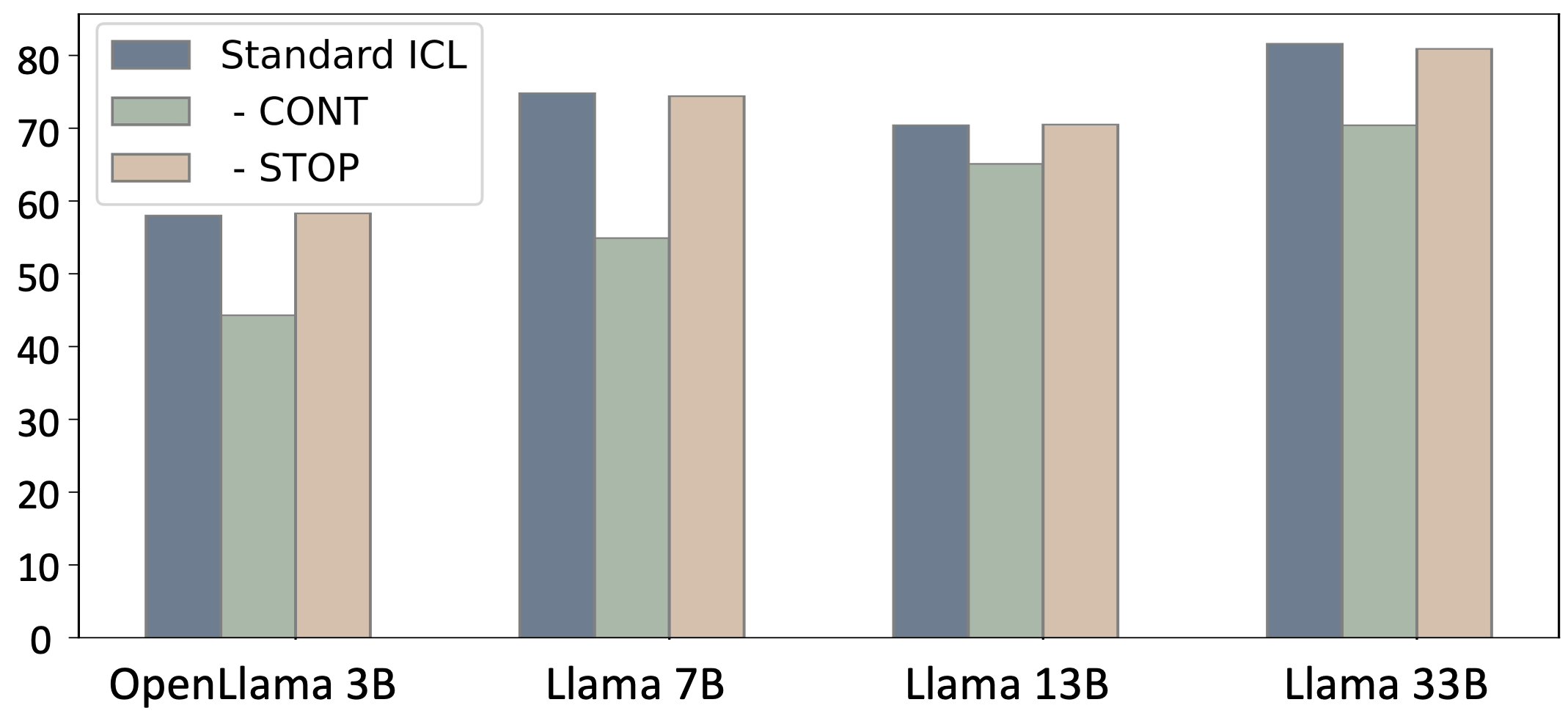} 
    \caption{\small Results of the token-level ablation where, for example, $-$\textsc{stop} refers to the ablation where stopword tokens are dropped from the ICL prompt. 
    Models \textbf{without template tokens} consistently yielded \textbf{an accuracy of 0\%} and are thus omitted from this figure.}
\label{fig:tokenablation}
\end{figure}

\subsubsection{Token-level ablation}
\label{sec:ablation_tokens}
In this section, we modify the ICL prompt by removing certain types of tokens from the ICL prompt\footnote{For template tokens, this includes \emph{both} the tokens in the demonstrations and the test example to maintain their consistency. We included the analyses of only ablating the tokens in the demonstrations in Appendix~\ref{app:token_level_template}.}
to further investigate the relationship between different kinds of tokens, cutting off the information flow between the representations of different tokens, shown in Figure~\ref{fig:ablationmethod}.
When we ablate the template tokens, we preserve the answer and next-line tokens in the templates to maintain a basic separator between the demonstration inputs and outputs. Results averaged on all the datasets are presented in Figure~\ref{fig:tokenablation}. Detailed results on each dataset could be seen in Appendix~\ref{app:token_ablation}.

Our first finding from this ablation is that information is propagated from content token representations to performance-critical token representations, as shown by the contrast between representation-level and token-level ablation in Table~\ref{tab:comparison_token_rep}. The representations of template and stopword tokens alone (i.e., Standard ICL $-$ \textsc{cont} in Figure~\ref{fig:tokenablation}) are less effective at affecting task performance, leading to worse performance than those where content
information is included in their attention
(i.e., Standard ICL $-$ \textsc{cont} in Table~\ref{tab:repablation_final1}).

\begin{table}[t]
\small
\centering
\resizebox{0.9\linewidth}{!}{
\begin{tabular}{l|rrrrr}
\toprule
Ablation($-$ \textsc{cont}) & OpenLlama 3B & Llama 7B & Llama 13B & Llama 33B \\
\midrule
Rep. level & 57.1 & 72.8  & 70.6 & 80.1 \\
Token level & 44.3 & 54.9 & 65.1  & 70.4 \\
$\Delta$ Difference &   12.8 & 17.9 &   5.5 & 9.7      \\
\bottomrule
\end{tabular}
}
\caption{The comparison between the $-$\textsc{cont} performance of different models across two levels of ablations. "Rep. level" refers to representation level.}
\label{tab:comparison_token_rep}
\end{table}

This finding provide us with additional insights about how LLMs leverage different kinds of tokens during ICL.
Firstly, this circumstance means that even though the representations of the content tokens are not directly used when LLMs predict the answer, the encoding of these tokens contribute to the final performance indirectly through being aggregated into the representations of the performance-critical tokens.
Secondly, it also suggests that LLMs prefer to utilize the the performance-critical tokens to aggregate the indirect information from the demonstration rather than others (i.e., content tokens). It is their incorporation of this information that makes them better at encoding tasks, partially explaining the working mechanism of ICL.


In addition, removing template tokens causes the LLMs to completely lose their ability to solve tasks via ICL with an overall task accuracy performance of 0\% for all sizes and all tasks. We hypothesize that this is because the model no longer has an explicit cue to generate the target label, which is further discussed in Section~\ref{sec:text_formattingj}. In this case, if we add back the last prompt token after the next-line token, the results return to their original level due to the introduction of a template token. 
This finding confirms previous claims that the format of ICL prompts plays a significant role in retaining performance~\citep{min2022rethinking}.

\textbf{Roles of different types of tokens. }
To summarize, template and stopword tokens are the most performance-critical. Template token representations significantly improve the task performance, while the representations of stopword tokens play a more supportive role in the spectrum of performance-critical tokens by summarizing the information of content tokens. In contrast, content token representations do not \textbf{directly} contribute to the performance but instead \textbf{indirectly} provide information that is aggregated into the other two types of tokens.
We discuss possible applications of these findings in Appendix~\ref{app:discussion}.
This finding further raises an additional question:
What are the characteristics for a token to be perceived by a LLM as performance-critical?

\begin{table}[t]
  \centering
  \resizebox{0.75\linewidth}{!}{
    \begin{tabular}{lll}
    \toprule
     Settings & Notations & Examples \\
    \midrule     
       \multirow{2}*{Random$_{\rm fixed}$} & ${\mathbf T}^{\rm in}$ & dsafjkldafdsajk: \{${\mathbf D}^{\rm in}$\}\textbackslash n \\
     & ${\mathbf T}^{\rm out}$ & reqwiorewsdafjl: \{${\mathbf D}^{\rm out}$\}\textbackslash n\textbackslash n \\
     \midrule
         \multirow{2}*{Swap} & ${\mathbf T}^{\rm in}$ & Answer: \{${\mathbf D}^{\rm in}$\}\textbackslash n \\
    & ${\mathbf T}^{\rm out}$ & Article: \{${\mathbf D}^{\rm out}$\}\textbackslash n\textbackslash n \\  
    \midrule
     \multirow{6}*{Random$_{\rm nonfixed}$} 
    & ${\mathbf T}^{\rm in}_1$ & dsafjkldaasdfjkl: \{${\mathbf D}^{\rm in}$\}\textbackslash n \\
    & ${\mathbf T}^{\rm out}_1$ & xiadfjdsalgfweqrjl: \{${\mathbf D}^{\rm out}$\}\textbackslash n\textbackslash n \\
    & ${\mathbf T}^{\rm in}_2$ & ewqroudajfsdafq: \{${\mathbf D}^{\rm in}$\}\textbackslash n \\
    & ${\mathbf T}^{\rm out}_2$ & yufoufgaddavfdnsl: \{${\mathbf D}^{\rm out}$\}\textbackslash n\textbackslash n \\

        & ${\mathbf T}^{\rm in}_t$ & vcxnkfgahvczxkl: \{${\mathbf D}^{\rm in}$\}\textbackslash n \\
    & ${\mathbf T}^{\rm out}_t$ & dafhglajfdvcaol: \{${\mathbf D}^{\rm out}$\}\textbackslash n\textbackslash n \\
    \bottomrule
    \end{tabular}
    }
\caption{An example of the ICL template with random strings used in AGNews.}
    \label{tab:randomICLtemplate_agnews}
\end{table}

\section{Performance-Critical Token Analyses}
\label{sec:analysis}
\label{sec:experiment_id}
To answer the above question, we provide analyses of the tokens whose representations we believe mainly store information that directly affects the performance of a task drastically. We focus on the template tokens since, as evidenced by the findings in Section~\ref{sec:ablation_representation}, \textbf{their representations are the most important to maintaining task performance}. 
Our analysis focuses on the distinguishing characteristics of performance-critical tokens, while we also examine the effects of each part of template tokens in Appendix~\ref{app:different_task_encoding}.

\begin{table}
  \centering
  \resizebox{0.9\linewidth}{!}{
    \begin{tabular}{llrrrrrr|rrrrrrrrrrrr}
    \toprule
     Models& Settings & AGNews & SST2    &TREC  &  DBPedia & RTE  & CB & Avg. \\
    \midrule
    \multirow{3}*{\makecell[l]{Open\\Llama\\ 3B}}
  & 
   Standard ICL & 63.7 & 91.2 & 21.9 & 61.9 & 57.4 & 52.0 & 58.0 \\
    &Swap &  64.4 & 86.8 & \underline{21.7} & 58.7 & 60.6 & 54.6 & 57.8  \\
    & Random$_{\rm fixed}$ &  \underline{57.5} & \underline{71.4} & 32.4 & \underline{51.2} & \underline{53.3} & \underline{49.8} & \underline{52.6} \\
    \midrule
    \multirow{3}*{\makecell[l]{Llama\\ 7B}}
& Standard ICL &  82.4 & 94.3 & 63.5 & 68.7 & 68.6 & 71.3 & 74.8 \\
    &Swap & 70.2 & \underline{11.4} & 44.3 & 58.2 & 64.5 & 50.1 & 49.8  \\
    & Random$_{\rm fixed}$ &   \underline{19.5} & 11.4 & \underline{13.2} & \underline{7.4} & \underline{19.7} & \underline{21.7} & \underline{15.5} \\
    \midrule
    \multirow{3}*{\makecell[l]{Llama\\ 13B}}
& Standard ICL &  81.6 & 94.3 & 60.0 & 76.1 & 70.6 & 39.9 & 70.4 \\
    &Swap & 81.5 & \underline{67.4} & 36.4 & 75.9 & 69.1 & 52.1 & 63.7 \\
    & Random$_{\rm fixed}$ & \underline{52.1} & 76.8 & \underline{27.7} & \underline{48.9} & \underline{55.7} & \underline{34.5} & \underline{49.3}  \\
    \midrule
    \multirow{3}*{\makecell[l]{Llama\\ 33B}}
& Standard ICL & 85.0 & 96.5 & 68.1 & 78.4 & 78.5 & 83.3 & 81.6 \\
    &Swap &  84.5 & 94.9 & 60.8 & \underline{75.5} & \underline{68.0} & 55.5  & 73.2 \\
    & Random$_{\rm fixed}$ &   \underline{78.7} & \underline{92.5} & \underline{52.2} & 75.8 & 68.9 & \underline{41.1} & \underline{68.2} \\
    \bottomrule
    \end{tabular}
    }

\caption{Results validating the effect of lexical meanings of template tokens, presented as percentages. The results showing the greatest decrease during the disruption are underlined.}
    \label{tab:randomtemplate_lexical}
\end{table}



By better understanding what characteristics of performance-critical tokens lead them to affect task performance, we provide insights on how to best leverage LLMs for ICL (e.g., What principles should practitioners be using when designing prompt templates?). 
We hypothesize that the following characteristics are critical for a token to be leveraged as performance-critical tokens: 
\textbf{lexical meaning} referring to the task-related lexical meaning of a performance-critical token, \textbf{repetition} referring to the multiple appearances of the performance-critical tokens in the prompt, and \textbf{structural cues} referring to how performance-critical tokens format the ICL prompt, shown in Table~\ref{tab:ICLexample}, into structured text. 

We design several experiments to test if these characteristics affect the impact of performance-critical tokens, by disrupting each characteristic in the ICL prompts.
A characteristic is related if there is a performance drop after the disruption. The disruption is achieved by replacing the template tokens with different kinds of random string templates, shown in Table~\ref{tab:randomICLtemplate_agnews}. We use 5 different random string templates which are attached to Appendix~\ref{app:random} and average all the results for each setting.

\textbf{Lexical meaning. }
A performance-critical token might be more impactful on the performance based on its lexical meaning.
One hypothesis is that if the token carries task-related meanings like ``Answer'', it is more likely to serve as a performance-critical token.

To verify if lexical meanings could affect the formation of performance-critical tokens, we 1)~Replace the tokens from ${\mathbf T}^{\rm in}$ and ${\mathbf T}^{\rm out}$ with the same random strings across the different demonstrations~(\textbf{Random$_{\rm fixed}$}), thus completely disrupting the lexical characteristic of these tokens; 
2) Swap ${\mathbf T}^{\rm in}$  and ${\mathbf T}^{\rm out}$ (\textbf{Swap}), thus partially disrupting the lexical characteristic of these tokens. 
Table~\ref{tab:randomtemplate_lexical} shows that disrupting the lexical meaning of tokens slightly impacts task performance in smaller models (OpenLlama 3B), while larger models experience more significant drops. Llama 7B, in particular, is highly sensitive to lexical meaning and performs worse when semantics are disturbed. Thus, the lexical meaning of tokens likely influences their performance-critical nature.

\textbf{Repetition. }
\label{sec:repeition}
The impact of performance-critical tokens could also be influenced by their repetition throughout the prompt. 
Intuitively, via the attention mechanism, repetitive patterns are more likely to propagate information through the processing of text.
\citet{yan2023understanding} propose self-reinforcement in in-context learning, also suggesting that repetition could be a significant factor in ICL.

We explore the repetition characteristic by comparing the results of the previously discussed \textbf{Random$_{\rm fixed}$} experiment with an experiment replacing ${\mathbf T}^{\rm in}$ and ${\mathbf T}^{\rm out}$ with different random strings (\textbf{Random$_{\rm nonfixed}$}), thus breaking the repetition of template tokens present in ICL demonstrations. 
We further conduct experiments \textbf{using template tokens with specific lexical meanings for comparison}, as detailed in Appendix~\ref{app:repetitive}.

\begin{table}[t]
  \centering
  \resizebox{\linewidth}{!}{
    \begin{tabular}{llrrrrrr|rrrr}
    \toprule
     Models& Settings & AGNews & SST2    &TREC  &  DBPedia & RTE  & CB & Avg. \\

    \midrule
    \multirow{2}*{\makecell[l]{OpenLlama\\ 3B}}& Random$_{\rm fixed}$ & \textbf{57.5} & \textbf{71.4} & \textbf{32.4} & \textbf{51.2} & \textbf{53.3} & \textbf{49.8} & \textbf{52.6}  \\
    
    & Random$_{\rm nonfixed}$ &  30.2 & 71.4 & 17.1 & 18.6 & 47.9 & 47.7 &38.8 \\

    \midrule
    \multirow{2}*{\makecell[l]{Llama\\ 7B}}& Random$_{\rm fixed}$ &   \textbf{19.5} & 11.4 & \textbf{13.2} & \textbf{7.4} & \textbf{19.7} & 21.7 & \textbf{15.5} \\
   
    & Random$_{\rm nonfixed}$ &  15.5 & \textbf{11.6} & 10.4 & 1.8 & 4.6 & \textbf{25.6} & 11.6 \\

    \midrule
    \multirow{2}*{\makecell[l]{Llama\\ 13B}}& Random$_{\rm fixed}$ & \textbf{52.1} & \textbf{76.8} & \textbf{27.7} & \textbf{48.9} & \textbf{55.7} & \textbf{34.5} & \textbf{49.3}  \\

    & Random$_{\rm nonfixed}$ &  32.1 & 34.5 & 19.2 & 6.0 & 21.0 & 32.8  & 24.3 \\

    \midrule
    \multirow{2}*{\makecell[l]{Llama\\ 33B}}& Random$_{\rm fixed}$ &  \textbf{78.7} & \textbf{92.5} &\textbf{ 52.2} &\textbf{ 75.8} & \textbf{68.9} & 41.1 & \textbf{68.2}  \\
  
    & Random$_{\rm nonfixed}$ &  78.5 & 87.5 & 46.3 & 63.1 & 63.6 & \textbf{46.1} & 64.2 \\

    \bottomrule
    \end{tabular}
    }
\caption{Results validating the effect of repetitive patterns, presented as percentages. We bold the highest accuracy for each classification task and model size.}
    \label{tab:randomtemplate_repetitive}
\end{table}
Table~\ref{tab:randomtemplate_repetitive} shows that without consistent repetition of performance-critical tokens, performance decreases for most models. This suggests that necessary information may not have been properly accumulated in the template token representations. These experiments demonstrate that repetition significantly influences the impact of performance-critical tokens.
The results align with previous findings, reinforcing our claim that repetition is a key characteristic.

\textbf{Structural cues. }
\label{sec:text_formattingj}
Beyond lexical meaning and repetition, the influence of performance-critical tokens may also depend on how ICL prompts are formatted. ICL prompts often include structural cues to assist the model to differentiate between elements with distinct roles, such as task inputs and target labels, within a demonstration. For example, template tokens (i.e., ${\mathbf T}^{\rm in}$ and ${\mathbf T}^{\rm out}$) delimit demonstration examples and labels, while stopword tokens (e.g., ,'', .’’, ``:’’, etc.) structure content words into sentence components. Examples of how performance-critical tokens delimit ICL prompts are shown in Appendix~\ref{app:text_formatting}. These structural cues are similar to those in LLM pretraining data (e.g., column names in SQL tables), suggesting that pretraining on such data enables the model to recognize the structuring role of performance-critical tokens, allowing the representations to store higher-level information.

\begin{table}[t]
  \centering
  \resizebox{\linewidth}{!}{
    \begin{tabular}{llrrrrrr|rrrr}
    \toprule
     Models& Settings & AGNews & SST2    &TREC  &  DBPedia & RTE  & CB & Avg. \\

    \midrule
    \multirow{2}*{\makecell[l]{OpenLlama\\ 3B}}& Standard ICL &  70.7 & 51.7 & 40.4 & 53.5 & 50.2 & 48.6 & 53.3  \\
    
    & Random$_{\rm fixed}$ & 47.5 & 51.8 & 32.6 & 19.4 & 51.8 & 42.4 & 40.9 \\

    \midrule
    \multirow{2}*{\makecell[l]{Llama\\ 7B}}& Standard ICL &  72.3 & 77.4 & 54.1 & 64.7 & 53.0 & 64.4 & 64.3 \\
   
    &  Random$_{\rm fixed}$ & 3.9 & 16.9 & 3.5 & 9.6 & 16.9 & 10.4 & 10.2  \\

    \midrule
    \multirow{2}*{\makecell[l]{Llama\\ 13B}}& Standard ICL  & 82.0 & 72.0 & 60.1 & 75.9 & 60.4 & 18.8 & 70.1   \\

    & Random$_{\rm fixed}$ &  46.1 & 47.5 & 25.0 & 50.8 & 47.5 & 21.4 & 39.7 \\

    \midrule
    \multirow{2}*{\makecell[l]{Llama\\ 33B}}& Standard ICL & 85.3 & 88.3 & 71.2 & 75.5 & 64.1 & 45.5 & 76.9 \\
  
    & Random$_{\rm fixed}$ &  69.7 & 53.0 & 37.8 & 72.8 & 53.0 & 37.6 & 54.0 \\

    \bottomrule
    \end{tabular}
    }
        \caption{One-shot experimental results validating the effect of structural cues, presented as percentages. Models \textbf{without template tokens} consistently yielded \textbf{an accuracy of 0\%} and are thus omitted from this table. }
    \label{tab:new_structural}
\end{table}

To assess the structuring characteristic of performance-critical tokens, we perturb the structure of one-shot ICL prompts in two stages, where the one-shot setting could eliminate repetition as a confounding factor. First, we disrupt the lexical meaning of template tokens, since token meaning helps LLMs distinguishing the different parts of a prompt. Then, we remove all template tokens to eliminate any source of structure cues.
Table~\ref{tab:new_structural} shows that disrupting structural cues decreases performance, highlighting their importance. Consistent with Section~\ref{sec:ablation_tokens}, removing all template tokens results in 0\% performance due to the complete elimination of structural cues. Supplemental experiments in Appendix~\ref{app:text_formatting0} further support this from a representation-level perspective.

\section{Conclusion}

In this paper, we have provided a fine-grained characterization of performance-critical tokens, whose representations LLMs directly depend on to achieve high-level performance in ICL. Through a series of 
experiments, we have examined the roles of template tokens and stopword tokens within ICL as potential performance-critical tokens. Our findings add nuance to previous claims made about ICL, for example, that tokens other than label words could also provide valuable information directly affecting the performance. Overall, our results demonstrate that model performance depends directly on the presence of these tokens and that their lexical meaning, their repetition throughout the ICL prompt, and their structural formatting of ICL demonstrations are likely to play a role in how effectively they allow an LLM to recover the critical information needed to perform a task.



\section*{Ethics Statement}
\label{app:ethics}
This work focuses on analyzing the working mechanisms of large language models and, as such, does not present any increased risks of harm beyond the existing norms of natural language processing or computational linguistics research. The associated risks include using a model trained on vast amounts of text, which may inadvertently contain biases. Another concern is the potential misuse of the model for generating misleading or harmful content. However, such a scenario is unlikely in our work, as we concentrate on classification tasks with fixed outputs.

\section*{Acknowledgements}
\label{app:ethics}
The authors thank all the reviewers for their suggestions and comments. This work is supported by
National Natural Science Foundation of China (No. U21B2009). 
Jackie Chi Kit Cheung is supported by Canada CIFAR AI Chair program.

\bibliography{AAAI_conference, custom, custom_2}

@article{team2025gemma,
  title={Gemma 3 technical report},
  author={Team, Gemma and Kamath, Aishwarya and Ferret, Johan and Pathak, Shreya and Vieillard, Nino and Merhej, Ramona and Perrin, Sarah and Matejovicova, Tatiana and Ram{\'e}, Alexandre and Rivi{\`e}re, Morgane and others},
  journal={arXiv preprint arXiv:2503.19786},
  year={2025}
}

@article{touvron2023llama,
  title={Llama: Open and efficient foundation language models},
  author={Touvron, Hugo and Lavril, Thibaut and Izacard, Gautier and Martinet, Xavier and Lachaux, Marie-Anne and Lacroix, Timoth{\'e}e and Rozi{\`e}re, Baptiste and Goyal, Naman and Hambro, Eric and Azhar, Faisal and others},
  journal={arXiv preprint arXiv:2302.13971},
  year={2023}
}

@article{touvron2023llama2,
  title={Llama 2: Open foundation and fine-tuned chat models},
  author={Touvron, Hugo and Martin, Louis and Stone, Kevin and Albert, Peter and Almahairi, Amjad and Babaei, Yasmine and Bashlykov, Nikolay and Batra, Soumya and Bhargava, Prajjwal and Bhosale, Shruti and others},
  journal={arXiv preprint arXiv:2307.09288},
  year={2023}
}

@misc{jiang2023mistral7b,
      title={Mistral 7B}, 
      author={Albert Q. Jiang and Alexandre Sablayrolles and Arthur Mensch and Chris Bamford and Devendra Singh Chaplot and Diego de las Casas and Florian Bressand and Gianna Lengyel and Guillaume Lample and Lucile Saulnier and Lélio Renard Lavaud and Marie-Anne Lachaux and Pierre Stock and Teven Le Scao and Thibaut Lavril and Thomas Wang and Timothée Lacroix and William El Sayed},
      year={2023},
      eprint={2310.06825},
      archivePrefix={arXiv},
      primaryClass={cs.CL},
      url={https://arxiv.org/abs/2310.06825}, 
}

@inproceedings{zhang2022active,
  title={Active Example Selection for In-Context Learning},
  author={Zhang, Yiming and Feng, Shi and Tan, Chenhao},
  booktitle={Proc. of EMNLP 2022},
  pages={9134--9148},
  year={2022}
}

@inproceedings{yin-etal-2023-read,
    title = "Did You Read the Instructions? Rethinking the Effectiveness of Task Definitions in Instruction Learning",
    author = "Yin, Fan  and
      Vig, Jesse  and
      Laban, Philippe  and
      Joty, Shafiq  and
      Xiong, Caiming  and
      Wu, Chien-Sheng",
    booktitle = "Proc. of ACL 2023",
    month = jul,
    year = "2023",
    address = "Toronto, Canada",
    publisher = "Association for Computational Linguistics",
    url = "https://aclanthology.org/2023.acl-long.172",
    doi = "10.18653/v1/2023.acl-long.172",
    pages = "3063--3079",
}

@inproceedings{lu-etal-2022-fantastically,
    title = "Fantastically Ordered Prompts and Where to Find Them: Overcoming Few-Shot Prompt Order Sensitivity",
    author = "Lu, Yao  and
      Bartolo, Max  and
      Moore, Alastair  and
      Riedel, Sebastian  and
      Stenetorp, Pontus",
    booktitle = "Proc. of ACL 2022",
    month = may,
    year = "2022",
    address = "Dublin, Ireland",
    publisher = "Association for Computational Linguistics",
    url = "https://aclanthology.org/2022.acl-long.556",
    doi = "10.18653/v1/2022.acl-long.556",
    pages = "8086--8098",
}

@InProceedings{pmlr-v139-zhao21c,
  title = 	 {Calibrate Before Use: Improving Few-shot Performance of Language Models},
  author =       {Zhao, Zihao and Wallace, Eric and Feng, Shi and Klein, Dan and Singh, Sameer},
  booktitle = 	 {ICML 2021},
  pages = 	 {12697--12706},
  year = 	 {2021},
  editor = 	 {Meila, Marina and Zhang, Tong},
  volume = 	 {139},
  series = 	 {Proceedings of Machine Learning Research},
  month = 	 {18--24 Jul},
  publisher =    {PMLR},
  url = 	 {https://proceedings.mlr.press/v139/zhao21c.html}
}

@inproceedings{NIPS2015_250cf8b5,
 author = {Zhang, Xiang and Zhao, Junbo and LeCun, Yann},
 booktitle = {NeurIPS},
 editor = {C. Cortes and N. Lawrence and D. Lee and M. Sugiyama and R. Garnett},
 pages = {},
 publisher = {Curran Associates, Inc.},
 title = {Character-level Convolutional Networks for Text Classification},
 url = {https://proceedings.neurips.cc/paper_files/paper/2015/file/250cf8b51c773f3f8dc8b4be867a9a02-Paper.pdf},
 volume = {28},
 year = {2015}
}

@inproceedings{de2019commitmentbank,
  title={The commitmentbank: Investigating projection in naturally occurring discourse},
  author={De Marneffe, Marie-Catherine and Simons, Mandy and Tonhauser, Judith},
  booktitle={proceedings of Sinn und Bedeutung},
  volume={23},
  number={2},
  pages={107--124},
  year={2019}
}

@inproceedings{dagan2005pascal,
  title={The pascal recognising textual entailment challenge},
  author={Dagan, Ido and Glickman, Oren and Magnini, Bernardo},
  booktitle={Machine learning challenges workshop},
  pages={177--190},
  year={2005},
  organization={Springer}
}

@inproceedings{socher2013recursive,
  title={Recursive deep models for semantic compositionality over a sentiment treebank},
  author={Socher, Richard and Perelygin, Alex and Wu, Jean and Chuang, Jason and Manning, Christopher D and Ng, Andrew Y and Potts, Christopher},
  booktitle={Proceedings of the 2013 conference on empirical methods in natural language processing},
  pages={1631--1642},
  year={2013}
}

@inproceedings{voorhees2000building,
  title={Building a question answering test collection},
  author={Voorhees, Ellen M and Tice, Dawn M},
  booktitle={Proc. of SIGIR 2000},
  pages={200--207},
  year={2000}
}

@article{brown2020language,
  title={Language models are few-shot learners},
  author={Brown, Tom and Mann, Benjamin and Ryder, Nick and Subbiah, Melanie and Kaplan, Jared D and Dhariwal, Prafulla and Neelakantan, Arvind and Shyam, Pranav and Sastry, Girish and Askell, Amanda and others},
  journal={NeurIPS},
  volume={33},
  pages={1877--1901},
  year={2020}
}

@misc{hendel2023incontext,
      title={In-Context Learning Creates Task Vectors}, 
      author={Roee Hendel and Mor Geva and Amir Globerson},
      year={2023},
      eprint={2310.15916},
      archivePrefix={arXiv},
      primaryClass={cs.CL}
}

@misc{todd2023function,
      title={Function Vectors in Large Language Models}, 
      author={Eric Todd and Millicent L. Li and Arnab Sen Sharma and Aaron Mueller and Byron C. Wallace and David Bau},
      year={2023},
      eprint={2310.15213},
      archivePrefix={arXiv},
      primaryClass={cs.CL}
}

@article{yan2023understanding,
  title     = {Understanding In-Context Learning from Repetitions},
  author    = {Jianhao Yan and Jin Xu and Chiyu Song and Chenming Wu and Yafu Li and Yue Zhang},
  journal   = {ICLR},
  year      = {2024},
  doi       = {10.48550/arXiv.2310.00297},
  bibSource = {Semantic Scholar https://www.semanticscholar.org/paper/a865d04c266fd2b3ea820b5741b7420779db9f73}
}

@article{liu2023incontext,
  title   = {In-context Vectors: Making In Context Learning More Effective and Controllable Through Latent Space Steering},
  author  = {Sheng Liu and Lei Xing and James Zou},
  year    = {2023},
  journal = {arXiv preprint arXiv: 2311.06668}
}

@misc{guo2023transformers,
      title={How Do Transformers Learn In-Context Beyond Simple Functions? A Case Study on Learning with Representations}, 
      author={Tianyu Guo and Wei Hu and Song Mei and Huan Wang and Caiming Xiong and Silvio Savarese and Yu Bai},
      year={2023},
      eprint={2310.10616},
      archivePrefix={arXiv},
      primaryClass={cs.LG}
}

@article{bhattamishra2023understanding,
  title={Understanding In-Context Learning in Transformers and LLMs by Learning to Learn Discrete Functions},
  author={Bhattamishra, Satwik and Patel, Arkil and Blunsom, Phil and Kanade, Varun},
  journal={arXiv preprint arXiv:2310.03016},
  year={2023}
}

@article{min2022rethinking,
  title   = {Rethinking the Role of Demonstrations: What Makes In-Context Learning Work?},
  author  = {Sewon Min and Xinxi Lyu and Ari Holtzman and Mikel Artetxe and Mike Lewis and Hannaneh Hajishirzi and Luke Zettlemoyer},
  year    = {2022},
  journal = {arXiv preprint arXiv: 2202.12837},
  url     = {https://arxiv.org/abs/2202.12837v2}
}

@article{liu2021makes,
  title     = {What Makes Good In-Context Examples for GPT-$3$?},
  author    = {Jiachang Liu and Dinghan Shen and Yizhe Zhang and Bill Dolan and L. Carin and Weizhu Chen},
  journal   = {Workshop on Knowledge Extraction and Integration for Deep Learning Architectures; Deep Learning Inside Out},
  year      = {2021},
  doi       = {10.18653/v1/2022.deelio-1.10},
  bibSource = {Semantic Scholar https://www.semanticscholar.org/paper/59641c10ed7431a3cf841f308367dc2dc0281b74},
  url       = {https://arxiv.org/abs/2101.06804v1}
}

@article{yoo2022groundtruth,
  title   = {Ground-Truth Labels Matter: A Deeper Look into Input-Label Demonstrations},
  author  = {Kang Min Yoo and Junyeob Kim and Hyuhng Joon Kim and Hyunsoo Cho and Hwiyeol Jo and Sang-Woo Lee and Sang-goo Lee and Taeuk Kim},
  year    = {2022},
  journal = {arXiv preprint arXiv: 2205.12685}
}

@article{xie2021explanation,
  title     = {An Explanation of In-context Learning as Implicit Bayesian Inference},
  author    = {Sang Michael Xie and Aditi Raghunathan and Percy Liang and Tengyu Ma},
  journal   = {International Conference on Learning Representations},
  year      = {2021},
  bibSource = {Semantic Scholar https://www.semanticscholar.org/paper/10bd4160b44803ada6a3d2e366c44b7e2a4ffe90}
}

@article{akyrek2022learning,
  title     = {What learning algorithm is in-context learning? Investigations with linear models},
  author    = {Ekin Akyürek and D. Schuurmans and Jacob Andreas and Tengyu Ma and Denny Zhou},
  journal   = {ICLR},
  year      = {2022},
  doi       = {10.48550/arXiv.2211.15661},
  bibSource = {Semantic Scholar https://www.semanticscholar.org/paper/7aa801b907b59b8ee4cfb1296d9dac22c5164c5d}
}

@article{bai2023transformers,
  title   = {Transformers as Statisticians: Provable In-Context Learning with In-Context Algorithm Selection},
  author  = {Yu Bai and Fan Chen and Huan Wang and Caiming Xiong and Song Mei},
  year    = {2023},
  journal = {arXiv preprint arXiv: 2306.04637},
  url     = {https://arxiv.org/abs/2306.04637v2}
}

@inproceedings{pmlr-v202-li23l,
  title     = {Transformers as Algorithms: Generalization and Stability in In-context Learning},
  author    = {Li, Yingcong and Ildiz, Muhammed Emrullah and Papailiopoulos, Dimitris and Oymak, Samet},
  booktitle = {Proc. of ICML 2023},
  pages     = {19565-19594},
  year      = {2023},
  editor    = {Krause, Andreas and Brunskill, Emma and Cho, Kyunghyun and Engelhardt, Barbara and Sabato, Sivan and Scarlett, Jonathan},
  volume    = {202},
  series    = {Proceedings of Machine Learning Research},
  month     = {23-29 Jul},
  publisher = {PMLR},
  url       = {https://proceedings.mlr.press/v202/li23l.html}
}

@article{wang2023label,
  title     = {Label Words are Anchors: An Information Flow Perspective for Understanding In-Context Learning},
  author    = {Lean Wang and Lei Li and Damai Dai and Deli Chen and Hao Zhou and Fandong Meng and Jie Zhou and Xu Sun},
  journal   = {Proc. of EMNLP 2023},
  year      = {2023},
  doi       = {10.48550/arXiv.2305.14160},
  bibSource = {Semantic Scholar https://www.semanticscholar.org/paper/7bd4ca8706a79983d31ab74e6c79bfdfd949602e}
}

@inproceedings{DBLP:conf/acl/PanG0C23,
  author    = {Jane Pan and Tianyu Gao and Howard Chen and Danqi Chen},
  editor    = {Anna Rogers and Jordan L. Boyd{-}Graber and Naoaki Okazaki},
  title     = {What In-Context Learning "Learns" In-Context: Disentangling Task Recognition and Task Learning},
  booktitle = {Findings of ACL 2023},
  pages     = {8298-8319},
  publisher = {Association for Computational Linguistics},
  year      = {2023},
  url       = {https://doi.org/10.18653/v1/2023.findings-acl.527},
  doi       = {10.18653/V1/2023.FINDINGS-ACL.527},
  bibsource = {dblp computer science bibliography, https://dblp.org}
}

@article{loper2002nltk,
  title={Nltk: The natural language toolkit},
  author={Loper, Edward and Bird, Steven},
  journal={arXiv preprint cs/0205028},
  year={2002}
}

@inproceedings{NEURIPS2023_a452a7c6,
  author    = {Liu, Zichang and Desai, Aditya and Liao, Fangshuo and Wang, Weitao and Xie, Victor and Xu, Zhaozhuo and Kyrillidis, Anastasios and Shrivastava, Anshumali},
  booktitle = {NeurIPS 2023},
  editor    = {A. Oh and T. Neumann and A. Globerson and K. Saenko and M. Hardt and S. Levine},
  pages     = {52342-52364},
  publisher = {Curran Associates, Inc.},
  title     = {Scissorhands: Exploiting the Persistence of Importance Hypothesis for LLM KV Cache Compression at Test Time},
  url       = {https://proceedings.neurips.cc/paper_files/paper/2023/file/a452a7c6c463e4ae8fbdc614c6e983e6-Paper-Conference.pdf},
  volume    = {36},
  year      = {2023}
}

@article{costa2022no,
  title={No language left behind: Scaling human-centered machine translation},
  author={Costa-juss{\`a}, Marta R and Cross, James and {\c{C}}elebi, Onur and Elbayad, Maha and Heafield, Kenneth and Heffernan, Kevin and Kalbassi, Elahe and Lam, Janice and Licht, Daniel and Maillard, Jean and others},
  journal={arXiv preprint arXiv:2207.04672},
  year={2022}
}

@inproceedings{papineni2002bleu,
  title={Bleu: a method for automatic evaluation of machine translation},
  author={Papineni, Kishore and Roukos, Salim and Ward, Todd and Zhu, Wei-Jing},
  booktitle={Proc. of ACL 2002},
  pages={311--318},
  year={2002}
}

@article{madaan2022text,
  title={Text and patterns: For effective chain of thought, it takes two to tango},
  author={Madaan, Aman and Yazdanbakhsh, Amir},
  journal={arXiv preprint arXiv:2209.07686},
  year={2022}
}

@article{chen2024parallel,
  title={Parallel structures in pre-training data yield in-context learning},
  author={Chen, Yanda and Zhao, Chen and Yu, Zhou and McKeown, Kathleen and He, He},
  journal={arXiv preprint arXiv:2402.12530},
  year={2024}
}

@article{zhang2023h_2o,
  title   = {H$_2$O: Heavy-Hitter Oracle for Efficient Generative Inference of Large Language Models},
  author  = {Zhenyu Zhang and Ying Sheng and Tianyi Zhou and Tianlong Chen and Lianmin Zheng and Ruisi Cai and Zhao Song and Yuandong Tian and Christopher Ré and Clark Barrett and Zhangyang Wang and Beidi Chen},
  year    = {2023},
  journal = {arXiv preprint arXiv: 2306.14048}
}

@article{bai2024citrus,
  title   = {CItruS: Chunked Instruction-aware State Eviction for Long Sequence Modeling},
  author  = {Yu Bai and Xiyuan Zou and Heyan Huang and Sanxing Chen and Marc-Antoine Rondeau and Yang Gao and Jackie Chi Kit Cheung},
  year    = {2024},
  journal = {arXiv preprint arXiv: 2406.12018}
}

@article{zhou2023mystery,
  title   = {The Mystery of In-Context Learning: A Comprehensive Survey on Interpretation and Analysis},
  author  = {Yuxiang Zhou and Jiazheng Li and Yanzheng Xiang and Hanqi Yan and Lin Gui and Yulan He},
  year    = {2023},
  journal = {arXiv preprint arXiv: 2311.00237}
}

@article{mao2024data,
  title   = {A Data Generation Perspective to the Mechanism of In-Context Learning},
  author  = {Haitao Mao and Guangliang Liu and Yao Ma and Rongrong Wang and Kristen Johnson and Jiliang Tang},
  year    = {2024},
  journal = {arXiv preprint arXiv: 2402.02212}
}

@inproceedings{han2023understanding,
  title={Understanding In-Context Learning via Supportive Pretraining Data},
  author={Han, Xiaochuang and Simig, Daniel and Mihaylov, Todor and Tsvetkov, Yulia and Celikyilmaz, Asli and Wang, Tianlu},
  booktitle={Proc. of ACL 2023, Vol. 1: Long Papers},
  pages={12660--12673},
  year={2023}
}

@article{zhao2024unveiling,
  title={Unveiling In-Context Learning: A Coordinate System to Understand Its Working Mechanism},
  author={Zhao, Anhao and Ye, Fanghua and Fu, Jinlan and Shen, Xiaoyu},
  journal={arXiv preprint arXiv:2407.17011},
  year={2024}
}

@article{lenartowicz2014neurophysiological,
  title={Neurophysiological signals of ignoring and attending are separable and related to performance during sustained intersensory attention},
  author={Lenartowicz, Agatha and Simpson, Gregory V and Haber, Catherine M and Cohen, Mark S},
  journal={Journal of cognitive neuroscience},
  volume={26},
  number={9},
  pages={2055--2069},
  year={2014},
  publisher={MIT Press One Rogers Street, Cambridge, MA 02142-1209, USA journals-info~…}
}

@article{sarica2021stopwords,
  title={Stopwords in technical language processing},
  author={Sarica, Serhad and Luo, Jianxi},
  journal={Plos one},
  volume={16},
  number={8},
  pages={e0254937},
  year={2021},
  publisher={Public Library of Science San Francisco, CA USA}
}

@inproceedings{talmor-etal-2019-commonsenseqa,
    title = "{C}ommonsense{QA}: A Question Answering Challenge Targeting Commonsense Knowledge",
    author = "Talmor, Alon  and
      Herzig, Jonathan  and
      Lourie, Nicholas  and
      Berant, Jonathan",
    booktitle = "Proc. of NAACL-HLT 2019, Vol. 1 (Long and Short Papers)",
    month = jun,
    year = "2019",
    address = "Minneapolis, Minnesota",
    publisher = "Association for Computational Linguistics",
    url = "https://aclanthology.org/N19-1421",
    doi = "10.18653/v1/N19-1421",
    pages = "4149--4158",
    archivePrefix = "arXiv",
    eprint        = "1811.00937",
    primaryClass  = "cs",
}

@article{li2023incontext,
  title   = {In-Context Learning with Many Demonstration Examples},
  author  = {Mukai Li and Shansan Gong and Jiangtao Feng and Yiheng Xu and Jun Zhang and Zhiyong Wu and Lingpeng Kong},
  year    = {2023},
  journal = {arXiv preprint arXiv: 2302.04931}
}

@article{hao2022structured,
  title   = {Structured Prompting: Scaling In-Context Learning to 1,000 Examples},
  author  = {Yaru Hao and Yutao Sun and Li Dong and Zhixiong Han and Yuxian Gu and Furu Wei},
  year    = {2022},
  journal = {arXiv preprint arXiv: 2212.06713}
}

@article{bertsch2024incontext,
  title   = {In-Context Learning with Long-Context Models: An In-Depth Exploration},
  author  = {Amanda Bertsch and Maor Ivgi and Uri Alon and Jonathan Berant and Matthew R. Gormley and Graham Neubig},
  year    = {2024},
  journal = {arXiv preprint arXiv: 2405.00200}
}

@article{wilbur1992automatic,
  title={The automatic identification of stop words},
  author={Wilbur, W John and Sirotkin, Karl},
  journal={Journal of information science},
  volume={18},
  number={1},
  pages={45--55},
  year={1992},
  publisher={Sage Publications Sage CA: Thousand Oaks, CA}
}

@article{rayner1998eye,
  title={Eye movements in reading and information processing: 20 years of research.},
  author={Rayner, Keith},
  journal={Psychological bulletin},
  volume={124},
  number={3},
  pages={372},
  year={1998},
  publisher={American Psychological Association}
}

@inproceedings{lin2024dual,
  title={Dual operating modes of in-context learning},
  author={Lin, Ziqian and Lee, Kangwook},
  booktitle={ICML 2024},
  year={2024}
}

@article{lilong,
  title={Long-context LLMs Struggle with Long In-context Learning},
  author={Li, Tianle and Zhang, Ge and Do, Quy Duc and Yue, Xiang and Chen, Wenhu},
  journal={TMLR},
  year={2024}
}

@article{yang2025task,
  title={Task vectors in in-context learning: Emergence, formation, and benefit},
  author={Yang, Liu and Lin, Ziqian and Lee, Kangwook and Papailiopoulos, Dimitris and Nowak, Robert},
  journal={arXiv preprint arXiv:2501.09240},
  year={2025}
}

@article{tikhonov2025one,
  title={One Task Vector is not Enough: A Large-Scale Study for In-Context Learning},
  author={Tikhonov, Pavel and Oseledets, Ivan and Tutubalina, Elena},
  journal={arXiv preprint arXiv:2505.23911},
  year={2025}
}

@article{dong2025understanding,
  title={Understanding Task Vectors in In-Context Learning: Emergence, Functionality, and Limitations},
  author={Dong, Yuxin and Jiang, Jiachen and Zhu, Zhihui and Ning, Xia},
  journal={arXiv preprint arXiv:2506.09048},
  year={2025}
}

@inproceedings{jiangunlocking,
  title={Unlocking the Power of Function Vectors for Characterizing and Mitigating Catastrophic Forgetting in Continual Instruction Tuning},
  author={Jiang, Gangwei and JIANG, Caigao and Li, Zhaoyi and Xue, Siqiao and ZHOU, JUN and Song, Linqi and Lian, Defu and Wei, Ying},
  booktitle={ICLR 2025},
  year={2025}
}


\clearpage
\appendix

\section{Reproducibility Statement}
\label{app:repro}
To ensure the reproducibility of our work, we have made several efforts that are documented throughout this paper. Our experiments utilize the open-source models described in Section~\ref{sec:experiment_setting}. The prompts and templates used in our experiments are detailed in Section~\ref{sec:notation}, Section~\ref{sec:experiment_id} of the main text and in Appendix~\ref{app:ICL_templates}, Appendix~\ref{app:rla_example}, Appendix~\ref{app:different_instruction}, Appendix~\ref{app:random}, Appendix~\ref{app:repetitive}. The stopword token list used in our experiments is shown in Appendix~\ref{app:stopword}. The complete code for our implementation, including all inference processes, is provided in the supplementary materials. We employed random seeds ranging from 1 to 15 to ensure consistent results across experiments, as specified in Section~\ref{sec:experiment_setting} and the supplementary code. All datasets used in our experiments are described comprehensively in Section~\ref{sec:experiment_setting}, and the supplementary code includes all data processing steps and any preprocessing applied. We encourage other researchers to consult these references for replicating our findings.
\section{Limitations}
\label{app:limitations}
In this paper, the token categorization is performed manually, leaving room for further refinement and the exploration of other specific content tokens as performance-critical tokens in certain contexts to future work. While the results provide robust support to our categorization, the identification process itself lacks precision. For instance, stopwords may only represent a subset of all in-context performance-critical tokens. The manual nature of our categorization limits our ability to comprehensively track these tokens. 

Moreover, our experiments are limited to classification, machine translation, and question answewring datasets, suggesting that our conclusions should be further validated for other tasks. Additionally, our focus on performance-critical tokens, whose representations could impact task performance, may overlook other tokens responsible for other possible functions. Another limitation of our study is that we focus exclusively on in-context learning scenarios, meaning that our findings may not be directly applicable to zero-shot learning scenarios.

\begin{table}[t]
  \centering

  \resizebox{0.9\linewidth}{!}{
    \begin{tabular}{llp{7cm}}
    \toprule
    Datasets & Notations & Examples \\
    \midrule     
   
    \multirow{4}*{\makecell[l]{AGNews}} &  ${\mathbf I}$ &    Classify the news articles into the categories of World, Sports, Business, and Technology.\textbackslash n\textbackslash n \\
    & ${\mathbf T}^{\rm in}$ & Article: \{${\mathbf D}^{\rm in}$\}\textbackslash n \\
    & ${\mathbf T}^{\rm out}$ & Answer: \{${\mathbf D}^{\rm out}$\}\textbackslash n\textbackslash n \\
    \midrule
    \multirow{4}*{\makecell[l]{SST2}}&  ${\mathbf I}$ &    Classify the reviews into the categories of Positive and Negative.\textbackslash n\textbackslash n \\
    & ${\mathbf T}^{\rm in}$ & Review: \{${\mathbf D}^{\rm in}$\}\textbackslash n \\
    & ${\mathbf T}^{\rm out}$ & Sentiment: \{${\mathbf D}^{\rm out}$\}\textbackslash n\textbackslash n \\
    \midrule
    \multirow{5}*{\makecell[l]{RTE}}&  ${\mathbf I}$ &   Classify the entailment of the hypothesis and the premise into the categories of True and False.\textbackslash n\textbackslash n \\
    & ${\mathbf T}^{\rm in}$ & Hypothesis: \{${\mathbf D}^{\rm inA}$\}\textbackslash n Premise: \{${\mathbf D}^{\rm inB}$\}\textbackslash n \\
    & ${\mathbf T}^{\rm out}$ & Answer: \{${\mathbf D}^{\rm out}$\}\textbackslash n\textbackslash n \\
    \midrule
    \multirow{5}*{\makecell[l]{CB}}&  ${\mathbf I}$ &   Classify the entailment of the hypothesis and the premise into the categories of true, neither and false.\textbackslash n\textbackslash n \\
    & ${\mathbf T}^{\rm in}$ & Hypothesis: \{${\mathbf D}^{\rm inA}$\}\textbackslash n Premise: \{${\mathbf D}^{\rm inB}$\}\textbackslash n \\
    & ${\mathbf T}^{\rm out}$ & Answer: \{${\mathbf D}^{\rm out}$\}\textbackslash n\textbackslash n \\
    \midrule
    \multirow{5}*{\makecell[l]{TREC}}&  ${\mathbf I}$ &   Classify the questions based on whether their answer type is a Number, Location, Person, Description, Entity, or Abbreviation.\textbackslash n\textbackslash n \\
    & ${\mathbf T}^{\rm in}$ & Question: \{${\mathbf D}^{\rm in}$\}\textbackslash n \\
    & ${\mathbf T}^{\rm out}$ & Answer Type: \{${\mathbf D}^{\rm out}$\}\textbackslash n\textbackslash n \\
    \midrule
    \multirow{7}*{\makecell[l]{DBPedia}}&  ${\mathbf I}$ &   Classify the documents based on whether they are about a Company, School, Artist, Athlete, Politician, Transportation, Building, Nature, Village, Animal, Plant, Album, Film, or Book.\textbackslash n\textbackslash n \\
    & ${\mathbf T}^{\rm in}$ & Article: \{${\mathbf D}^{\rm in}$\}\textbackslash n \\
    & ${\mathbf T}^{\rm out}$ & Answer: \{${\mathbf D}^{\rm out}$\}\textbackslash n\textbackslash n \\
    
    \bottomrule
    \end{tabular}
    }
    \caption{An example of the ICL template used in our experiments.}
    \label{tab:ICLtemplates}
\end{table}

\begin{table}[t]
  \centering

  \resizebox{0.99\linewidth}{!}{
    \begin{tabular}{llp{8cm}}
    \toprule
    Datasets & Stopwords \\
    \midrule     
   
    AGNews  &   ``the'', ``into'', ``of'', ``and'', ``,'' , ``.'', ``\textbackslash n'' \\
    SST2 &    ``the'',  ``into'', ``of'',  ``and'', ``.'', ``\textbackslash n'' \\
    RTE &    ``the'',  ``of'', ``into'', ``and'', ``into'', ``.'', ``\textbackslash n'' \\
    CB &   ``the'', ``of'',  ``and'',  ``into'', ``,'' ,``.'', ``\textbackslash n'' \\
    TREC  &   ``the'',  ``based'', ``on'',  ``whether'',  ``their'',  ``is'', ``a'', ``,'', ``or'', ``.'', ``\textbackslash n'' \\
    DBPedia&   ``the'',  ``based'',  ``on'', ``whether'', ``they'', ``are'', ``about'', ``a'', ``,'', ``or'', ``.'', ``\textbackslash n'' \\
    
    \bottomrule
    \end{tabular}
    }
  \caption{The stopwords used in our experiments.}
    \label{tab:stopwords}
\end{table}

\begin{table}[t]
\small
  \centering

  \resizebox{0.99\linewidth}{!}{
    \begin{tabular}{p{8cm}}
    \toprule
    NLTK Stopwords List\\
    \midrule     
``i'', ``me'', ``my'', ``myself'', ``we'', ``our'', ``ours'', ``ourselves'', ``you'', ``your'', ``yours'', ``yourself'', ``yourselves'', ``he'', ``him'', ``his'', ``himself'', ``she'', ``her'', ``hers'', ``herself'', ``it'', ``its'', ``itself'', ``they'', ``them'', ``their'', ``theirs'', ``themselves'', ``what'', ``which'', ``who'', ``whom'', ``this'', ``that'', ``these'', ``those'', ``am'', ``is'', ``are'', ``was'', ``were'', ``be'', ``been'', ``being'', ``have'', ``has'', ``had'', ``having'', ``do'', ``does'', ``did'', ``doing'', ``a'', ``an'', ``the'', ``and'', ``but'', ``if'', ``or'', ``because'', ``as'', ``until'', ``while'', ``of'', ``at'', ``by'', ``for'', ``with'', ``about'', ``against'', ``between'', ``into'', ``through'', ``during'', ``before'', ``after'', ``above'', ``below'', ``to'', ``from'', ``up'', ``down'', ``in'', ``out'', ``on'', ``off'', ``over'', ``under'', ``again'', ``further'', ``then'', ``once'', ``here'', ``there'', ``when'', ``where'', ``why'', ``how'', ``all'', ``any'', ``both'', ``each'', ``few'', ``more'', ``most'', ``other'', ``some'', ``such'', ``no'', ``nor'', ``not'', ``only'', ``own'', ``same'', ``so'', ``than'', ``too'', ``very'', ``s'', ``t'', ``can'', ``will'', ``just'', ``don'', ``should'', ``now'',  ``\!'', ``\@'',  ``\#'', ``\$'', ``\%'', ``\^{}'', ``\&'', ``*'', ``('', ``)'', ``-'', ``\_'', ``+'', ``='', ``['', ``]'', ``\{'', ``\}'',  ``|'', ``\textbackslash'', ``;'', ``\'', ``\'', ``<'', ``>'', ``,'', ``.'' , ``?'', ``/'', ``\textbackslash n'' \\
    \bottomrule
    \end{tabular}
    }
    \caption{The whole stopword list from NLTK. We add the punctuation tokens in this case.}
    \label{tab:stopwords_nltk}
\end{table}

\begin{table}[t]
  \centering
  \resizebox{0.99\linewidth}{!}{
\begin{tabular}{llrrlrr}
\toprule
  \multirow{2}*{Models} & \multirow{2}*{Settings} & \multicolumn{2}{c}{RTE} & \multirow{2}*{Settings} & \multicolumn{2}{c}{RTE} \\
  \cmidrule(r{4pt}){3-4} \cmidrule(r{4pt}){6-7}
   & & \textbf{with ``:''} & \textbf{w/o ``:''} & & \textbf{with ``:''} & \textbf{w/o ``:''}  \\
\midrule

\multirow{3}*{\makecell[l]{Llama\\ 7B}}
& \textsc{temp} with ${\mathbf D}^{\rm out}$ & 66.3 & 59.5 & \textsc{temp} w/o ${\mathbf D}^{\rm out}$ & \underline{40.7} & \underline{42.5} \\
& \quad $- {\mathbf T}^{\rm in}$ & 58.9 & \underline{56.5} & \quad $- {\mathbf T}^{\rm in}$ & 43.7 & 49.9 \\
& \quad $- {\mathbf T}^{\rm out}$ & \underline{56.7} & 56.7 & \quad $- {\mathbf T}^{\rm out}$ & 56.0 & 55.6 \\
\midrule
\multirow{3}*{\makecell[l]{Llama\\ 13B}}
& \textsc{temp} with ${\mathbf D}^{\rm out}$ & 68.5 & 59.8 & \textsc{temp} w/o ${\mathbf D}^{\rm out}$ & 57.5 & 53.7 \\
& \quad $- {\mathbf T}^{\rm in}$ & 65.5 & 59.0 & \quad $- {\mathbf T}^{\rm in}$ & \underline{53.6} & \underline{52.8} \\
& \quad $- {\mathbf T}^{\rm out}$ & \underline{61.2} & \underline{58.4} & \quad $- {\mathbf T}^{\rm out}$ & 54.8 & 53.7 \\
\midrule
\multirow{3}*{\makecell[l]{Llama\\ 33B}}
& \textsc{temp} with ${\mathbf D}^{\rm out}$ & 79.0 & 77.1 & \textsc{temp} w/o ${\mathbf D}^{\rm out}$ & 71.8 & 65.8 \\
& \quad $- {\mathbf T}^{\rm in}$ & 77.4 & 75.4 & \quad $- {\mathbf T}^{\rm in}$ & 70.6 & 67.8 \\
& \quad $- {\mathbf T}^{\rm out}$ & \underline{72.8} & \underline{70.0} & \quad $- {\mathbf T}^{\rm out}$ & \underline{67.0} & \underline{61.3} \\
\bottomrule
\end{tabular}

    }
\caption{Ablation for different template token representations with and without ${\mathbf D}^{\rm out}$, presented as percentages. The results showing the greatest impact from ablation are underlined. }
    \label{tab:partial_combine}
\end{table}

\section{More Related Work}
\label{app:more_related_work}

\subsection{Working mechanisms of ICL}
Since the proposal of in-context learning~\citep{brown2020language}, its working mechanisms have been extensively studied by the research community~\citep{min2022rethinking, liu2021makes, bhattamishra2023understanding}. \citet{min2022rethinking} suggest that demonstrations primarily provide the label space, the distribution of the input text, and the format of the sequence for the test example. They argue that the precise ground truth labels do not have significant importance. In contrast, \citet{yoo2022groundtruth} propose a differing view, stating that the impact of the ground truth labels depends on the experimental configuration. \citet{xie2021explanation} explain ICL as implicit Bayesian inference, while \citet{akyrek2022learning} explore ICL learning process using linear models. Theoretical explanations~\citep{guo2023transformers, bai2023transformers, pmlr-v202-li23l} and gradient descent explanations have also been proposed. \citet{mao2024data} analyze in-context learning from the perspective of data generation. The perspective of supported training data is also leveraged to analyze ICL \citep{han2023understanding}. 
\citet{zhao2024unveiling} propose to use coordinate systems to understand the working mechanism of in-context learning.
\citet{zhou2023mystery} propose a comprehensive survey on the interpretation and analysis of in-context learning. 

Additional analyses exploring different aspects of ICL have also been studied. For instance, order sensitivity where task performance fluctuates based on the order of the same ICL demonstrations has been identified as a limitation of ICL~\citep{lu-etal-2022-fantastically}. 
\citet{yan2023understanding} propose that repetitive patterns in the prompt could affect the ICL performance in both positive and negative ways. 
\citet{DBLP:conf/acl/PanG0C23} analyze the ICL process by disentangling it into task recognition and task learning. 
\citet{madaan2022text} propose to define text and patterns while using counterfactual prompting for attributing token importance in chain-of-thought techniques. 

Our work investigates the working process of ICL in LLMs at inference time, demonstrating that certain specific tokens are more likely to possess representations that could affect the processing of the final test sample, improving the task performance.

\subsection{Function vectors of in-context learning}
\citet{todd2023function} and \citet{hendel2023incontext} provide evidence of function vectors that store information used to solve a task in ICL. They probe and extract the hidden representations of the final tokens in the prompt. These vectors can then be added to, or used to replace, the corresponding vectors in a zero-shot example, achieving results comparable to those obtained when the model uses all demonstrations as context. In addition, \citet{liu2023incontext} also propose using an in-context vector to represent the target task and applying feature shifting to query examples. They first feed each input and its corresponding target separately into an LLM, then concatenate all the latent states. A PCA method is applied to derive a vector that is more closely aligned with the task. Finally, \citet{wang2023label} propose that label words in the demonstration examples function as information anchors by aggregating the information from previous demonstrations and providing it to the test example. This finding suggests that we may view label tokens as satisfying our definition of performance-critical tokens.

All these previous studies either solely focus on a single token (i.e., the last prediction prompt token or label token) of the ICL prompt or treat the entire demonstration as a single unit, neglecting the other tokens within it. Our research focuses on all the tokens in the prompt and reveals that there are additional tokens with specific characteristics whose representations significantly affect the final ICL performance.

\section{Effects of different performance-critical tokens}
\label{app:different_task_encoding}
In this section, we aim at examining the relationship among the representations of different performance-critical tokens. To achieve this, we test the effectiveness (i.e., how much they could affect the downstream task performance) of each part of performance-critical tokens to see if they could work without each other.

To achieve this, we ablate the representations of each performance-critical token, similar to Section~\ref{sec:ablation_representation}.
In Section~\ref{sec:ablation_representation}, we assume that the label token ${\mathbf D}^{\rm out}$ is needed for ICL to achieve performance results on par with Standard ICL, as suggested by previous work~\citep{wang2023label}. 
However, it is still not known how the other performance-critical tokens affect the performance without ${\mathbf D}^{\rm out}$.
Hence, we divide our experiments by including or excluding the label tokens ${\mathbf D}^{\rm out}$ to further specifically investigate their effectiveness. We present the results on RTE datasets in Table~\ref{tab:partial_combine} while full results are shown in Appendix~\ref{app:full_results_individual}.

Overall, the above experiments show that the performance-critical tokens \textbf{should be utilized together} to provide the best performance and that removing some of them would cause \textbf{performance degeneration} or \textbf{instability} issues.
From the results with ${\mathbf D}^{\rm out}$, it is observed that all the template tokens (i.e., ${\mathbf T}^{\rm in}$, ${\mathbf T}^{\rm out}$, and ``:'') contribute to the final performance. Removing one of them would cause a performance degradation. From the results without ${\mathbf D}^{\rm out}$, the performance becomes less predictable, where adding back a template token (e.g., ``:'') does not always bring performance improvements. Moreover, in some datasets, models without ${\mathbf D}^{\rm out}$ can still achieve relatively high performance.
These results show that representations of other template tokens may also be seen as information anchors whose representations  aggregate and serve information to the final prediction of LLMs, broadening the conclusions of \citet{wang2023label} who claim that only answer tokens serve as information anchors.



\section{In-context Learning Templates}
\label{app:ICL_templates}
 In this section, we present all the in-context learning templates used in this paper. For the RTE and CB datasets, there are two distinct inputs in the demonstrations (i.e., the hypothesis and the premise), which we denote as ${\mathbf D}^{\rm inA}$ and ${\mathbf D}^{\rm inB}$, respectively. The examples are provided in Table~\ref{tab:ICLtemplates}. All the notations are consistent with the notations in Table~\ref{tab:ICLexample}. All the next-line tokens are represented as ``\textbackslash n ''

\begin{table*}[t]

  \centering
  \resizebox{0.7\linewidth}{!}{
    \begin{tabular}{lrrrrrr}
    \midrule
        Dataset & AGNews & DBPedia & SST2 & TREC & RTE & CB  \\ \midrule
        Test set number of the dataset & 7,601 & 70,000 & 1,821 & 500 & 277 & 250 \\ 
        Test set number we scaled up & 5,000 & 5,000 & 1,821 & - & - & -   \\ \midrule
    \end{tabular}
    }
\caption{Statistics for the original test set and the test set number we scaled up of each dataset we used.}
    \label{tab:more_example_statistics}
\end{table*}

\begin{table*}[t]

  \centering
  \resizebox{0.7\linewidth}{!}{
    \begin{tabular}{lrrrlrrr}
    \midrule
        \multicolumn{8}{c}{\textbf{OpenLlama 3B}} \\ \midrule
        Setting & AGNews & SST2 & DBPedia & Setting & AGNews & SST2 & DBPedia    \\ \midrule
        Zero-shot$+$\textsc{cont} & 25.7 & 52.4 & 7.8 & Standard ICL$-$\textsc{cont} &  57.6 & 86.4 & 62.3  \\ 
        Zero-shot$+$\textsc{stop} &  37.2 & 82.6 & 53.1 & Standard ICL$-$\textsc{stop} & 62.0 & 91.0 & 63.2   \\ 
        Zero-shot$+$\textsc{temp} & \textbf{56.2} & \textbf{86.3} & \textbf{62.5} & Standard ICL$-$\textsc{temp} & \underline{41.5} & \underline{87.2} & \underline{57.3} \\ \midrule
        \multicolumn{8}{c}{\textbf{Llama 7B}} \\ \midrule
        Setting & AGNews & SST2 & DBPedia & Setting & AGNews & SST2 & DBPedia  \\ \midrule
        Zero-shot$+$\textsc{cont} &  32.0 & 58.1 & 13.4 & Standard ICL$-$\textsc{cont} &77.0 & 91.9 & 67.4  \\ 
        Zero-shot$+$\textsc{stop} & 58.0 & 84.6 & 43.9 & Standard ICL$-$\textsc{stop} & 79.5 & 94.2 & 69.0 \\ 
        Zero-shot$+$\textsc{temp} & \textbf{71.4} & \textbf{90.7} & \textbf{67.4} & Standard ICL$-$\textsc{temp} & \underline{64.3} & \underline{84.9} & \underline{58.6} \\ \midrule
    \end{tabular}
    }
\caption{Results of the representation-level ablation experiments with more text examples. The best results are in bold while the results showing the greatest
decrease from ablation are underlined.}
    \label{tab:more_example_ablationrep}
\end{table*}


\section{Representation-level Ablation Examples}
\label{app:rla_example}
We provide a one-shot demonstration of all ablation cases for the representation-level ablation experiments, as shown in Table~\ref{tab:rla_example}. In these demonstrations, representations of specific token types are masked in the attention mechanism, where (m) denotes that all representations of the token are removed from the attention scope from the test example.
\begin{table*}[t]

  \centering
  \resizebox{0.8\linewidth}{!}{
    \begin{tabular}{p{18.5cm}}
    \toprule
 
        \midrule
\multicolumn{1}{c}{Zero-shot$+$\textsc{temp}}
    \\
    \midrule     
\textbf{(s) Classify the news articles into the categories of World, Sports, Business, and Technology.} \\
\\
\textbf{Article:} (m) (m) (m) (m) (m) (m) (m) (m) (m) (m) (m) (m) (m) (m) (m) (m) (m) (m) (m) (m) (m) (m) (m) (m) (m) (m) (m) (m) (m) (m) (m) (m) (m) (m) (m) (m) (m) (m) (m) (m) (m) (m) (m) (m) (m) (m) (m) (m) (m) (m) (m) (m) (m) (m) (m) (m) \\

\textbf{Answer: Technology} \\
    \midrule
        \midrule
\multicolumn{1}{c}{Zero-shot$+$\textsc{stop}}
    \\
    \midrule     
\textbf{(s) Classify the news articles into the categories of World, Sports, Business, and Technology.} \\
\\

(m) (m) (m) (m) (m) (m) \textbf{the} (m) \textbf{.} (m) (m) (m) (m) (m) (m) (m) (m) (m) (m) (m) (m) (m) (m) (m) (m) \textbf{the} (m) (m) (m) (m) (m) (m) (m) (m) \textbf{,} (m) (m) (m) (m) (m) (m) \textbf{of} (m) (m) (m) (m) (m) (m) (m) (m) (m) (m) \textbf{into} (m) (m) \textbf{the} (m) (m) \textbf{. }\\

(m) (m) \textbf{Technology} \\

    \midrule
       \midrule
    \multicolumn{1}{c}{Zero-shot$+$\textsc{cont}}
    \\
    \midrule     
    \textbf{(s) Classify the news articles into the categories of World, Sports, Business, and Technology. }\\
    \\
(m) (m) (m) \textbf{First class to} (m) \textbf{moon} (m) \textbf{London - British airline magnate Richard Branson announced a plan on Monday for} (m) \textbf{world's first commercial space flights} (m) \textbf{saying ``thousands''} (m) \textbf{fee-paying astronauts could be sent} (m) \textbf{orbit in} (m) \textbf{near future} (m) (m) (m) (m) \textbf{Technology} (m) (m) \\

     \midrule
        \midrule
\multicolumn{1}{c}{Standard ICL$-$\textsc{temp}}
    \\
    \midrule     
\textbf{(s) Classify the news articles into the categories of World, Sports, Business, and Technology.} \\
\\

(m) (m) (m) \textbf{First class to the moon. London - British airline magnate Richard Branson announced a plan on Monday for the world's first commercial space flights, saying "thousands" of fee-paying astronauts could be sent into orbit in the near future.} 
(m) (m) (m) \textbf{Technology} (m) (m) \\

    \midrule
        \midrule

 \multicolumn{1}{c}{Standard ICL$-$\textsc{stop}}    \\

         \midrule

\textbf{(s) Classify the news articles into the categories of World, Sports, Business, and Technology.} \\
\\

\textbf{Article: First class to} (m) \textbf{moon} (m) \textbf{London - British airline magnate Richard Branson announced a plan on Monday for} (m) \textbf{world's first commercial space flights} (m) \textbf{saying ``thousands''} (m) \textbf{fee-paying astronauts could be sent} (m) \textbf{orbit in} (m) \textbf{near future} (m) (m) 
\textbf{Answer: Technology} (m) (m) \\

            \midrule    
                        \midrule    
\multicolumn{1}{c}{Standard ICL$-$\textsc{cont}}
    \\

    \midrule     
\textbf{(s) Classify the news articles into the categories of World, Sports, Business, and Technology.} \\
\\

\textbf{Article:} (m) (m) (m) \textbf{the} (m) \textbf{.} (m) (m) (m) (m) (m) (m) (m) (m) (m) (m) (m) (m) (m) (m) (m) (m) \textbf{the} (m) (m) (m) (m) (m) (m) (m) (m) \textbf{,} (m) (m) (m) (m) (m) (m) \textbf{of} (m) (m) (m) (m) (m) (m) (m) (m) (m) (m) \textbf{into} (m) (m) \textbf{the} (m) (m) \textbf{.}\\

\textbf{Answer: Technology}
 \\
    \midrule

    \bottomrule
    \end{tabular}
    }
    \caption{An example of the masked tokens from the attention of the test example in the representation-level ablation, where (s) represents the start of sentence token and (m) means that this token is masked. Tokens that are not masked are bold for clarity.}
    \label{tab:rla_example}
\end{table*}

\section{Experiment Results with More Large Language Models}
\label{app:more_models}
Table~\ref{tab:repablation_final1_app} and \ref{tab:repablation_final2_app} show the full results containing Llama 7B and Llama 13B models.
Meanwhile, experimental results using the Llama 2 and Mistral models are shown in Table~\ref{tab:repablation_morellms_app}. The trends observed in these experiments are consistent with those involving the Llama models. These results further reinforce the findings of this paper, indicating that template tokens and stopword tokens are the most prone to serving as performance-critical tokens.

\begin{table}[t]
  \centering
  \resizebox{\linewidth}{!}{
    \begin{tabular}{llrrrrrr|l}
    \toprule
     Models & Setting & AGNews & SST2  &  TREC  & DBPedia & RTE & CB & $\triangle$Avg. \\
    \midrule
   \multirow{4}*{\makecell[l]{OpenLlama\\ 3B}} 
 
     & Zero-shot   &22.0 & 20.0 & 23.6 & 5.4 & 44.4 & 1.8& 19.5\\
   
   & \quad $+$ \textsc{cont}& 26.2 & 52.1 & 30.1 & 7.4 & 51.9 & 37.9 & \quad$+$14.8 \\
   & \quad $+$ \textsc{stop} &36.7 & 82.9 &\textbf{32.0}\asterisk & 52.4 & \textbf{58.8}& \textbf{56.2} & \quad$+$33.7\\
   & \quad $+$ \textsc{temp} &\textbf{56.5} &\textbf{86.7}& 27.1 &\textbf{62.2} & 56.4 & 52.3 & \quad\textbf{$+$37.4} \\

   \midrule
    \multirow{4}*{\makecell[l]{Llama\\ 7B}}  
     & Zero-shot  & 25.0 & 29.2 & 41.4 & 0.0 & 54.2 & 3.6& 25.6 \\
    &  \quad $+$ \textsc{cont}& 32.4 & 57.9 & 42.5 & 12.5 & 55.5 & 46.1 &\quad$+$15.6  \\
   &  \quad $+$ \textsc{stop} & 57.3 & 83.7 & 49.8 & 43.0 & 55.9 & 50.7 &\quad$+$31.1\\
   &  \quad $+$ \textsc{temp}& \textbf{70.8} & \textbf{90.2} & \textbf{58.4} & \textbf{66.2} & \textbf{66.3} & \textbf{73.5}  & \quad\textbf{$+$45.3}\\

   \midrule
    \multirow{4}*{\makecell[l]{Llama\\ 13B}}  
    
     & Zero-shot  &  59.0 & 18.0 & 37.0 & 0.0 & 0.0 & 0.0& 19.0 \\
    &  \quad $+$ \textsc{cont} &  27.7 & 52.4 & 33.5 & 10.9 & 61.7 & 41.7& \quad$+$19.0 \\
   &  \quad $+$ \textsc{stop} &72.2 & 73.5 & 46.8 & 50.7 & 58.6 & 30.6& \quad$+$36.4  \\
   &  \quad $+$ \textsc{temp} & \textbf{80.0} & \textbf{92.3} & \textbf{58.6} & \textbf{76.9} & \textbf{68.5} & \textbf{47.7}& \quad\textbf{$+$51.7}  \\

   \midrule
    \multirow{4}*{\makecell[l]{Llama\\ 33B}}  
     & Zero-shot  &  70.2 & 88.6 & 60.6 & 30.2 & 58.1 & 19.6 & 54.6 \\
    & \quad $+$ \textsc{cont}&  24.4 & 61.7 & 62.1 & 10.5 & 65.2 & 63.6 & \quad$-$6.7 \\
   &\quad $+$ \textsc{stop}  &  72.9 & 92.7 & \textbf{66.7}\asterisk & 69.1 & 69.6 & 63.0 & \quad$+$17.7\\
   &\quad $+$ \textsc{temp} &   \textbf{80.5} & \textbf{95.2} & 65.2 & \textbf{75.2} & \textbf{79.0} & \textbf{80.0}  & \quad\textbf{$+$24.6}\\
    \bottomrule
    \end{tabular}
    }

    \caption{The accuracy results of the representation-level ablation study using Llama models where, for example, + \textsc{temp} refers to allowing attention only to template tokens. All values are presented as percentages. Except where noted with $^*$, all test
statistics reported correspond to p-values $<$ 0.05. The best results are in bold.}
    \label{tab:repablation_final1_app}
\end{table}
\begin{table}[t]
    \centering
  \resizebox{0.99\linewidth}{!}{
    \begin{tabular}{llrrrrrr|l}
    \toprule
     Models & Setting & AGNews & SST2  &  TREC  & DBPedia & RTE & CB & $\triangle$Avg. \\
    \midrule
    
   \multirow{4}*{\makecell[l]{Open\\Llama \\ 3B}} 
 
     & Standard ICL &  63.7 & 91.2 & 21.9 & 61.9 & 57.4 & 52.0 & 58.0 \\
   
   & \quad$-$ \textsc{cont}&58.2 & 86.9 & \underline{27.6} & 61.9 & 56.5 & 51.7 & \quad$-$0.9  \\
   & \quad$-$ \textsc{stop} &51.8 & 78.9 & 28.8 & 30.3 & 53.6 & 45.2 & \quad$-$9.9  \\
   & \quad$-$ \textsc{temp} & \underline{26.2} & \underline{52.1} & 30.1 & \underline{7.4} & \underline{51.9} & \underline{37.9}&\quad\underline{$-$23.8}  \\
         \midrule
   \multirow{4}*{\makecell[l]{Llama \\ 7B}} 
 
     & Standard ICL    & 82.4 & 94.3 & 63.5 & 68.7 & 68.6 & 71.3&74.8  \\
   
   & \quad$-$ \textsc{cont}& 77.9 & 91.5 & 58.5 & 66.5 & 67.8 & 74.4&\quad$-$2.0  \\
   & \quad$-$ \textsc{stop}  & 78.5 & 88.7 & 39.3 & 66.7 & 60.6 & 60.4 &\quad$-$9.1\\
   & \quad$-$ \textsc{temp}  & \underline{20.8} & \underline{58.2} & \underline{32.4} & \underline{11.6} & \underline{54.4} & \underline{46.0} &\quad\underline{$-$37.6} \\

   \midrule
    \multirow{4}*{\makecell[l]{Llama \\ 13B}}  
     & Standard ICL   & 81.6 & 94.3 & 60.0 & 76.1 & 70.6 & 39.9 &70.4  \\
    &  \quad$-$ \textsc{cont}& 81.4 & 93.1 & 58.9 & 75.7 & 69.6 & 45.1& \quad$+$0.2 \\
   &  \quad$-$ \textsc{stop}  & 79.8 & 85.8 & 64.4 & 73.6 & 64.5 & \underline{40.2}&\quad$-$2.4\\
   &  \quad$-$ \textsc{temp} & \underline{27.8} & \underline{52.4} & \underline{33.5} & \underline{10.9} & \underline{63.1} & 45.6 &\quad\underline{$-$31.5}\\

   \midrule
    \multirow{4}*{\makecell[l]{Llama\\ 33B}}  
    
     & Standard ICL & 85.0 & 96.5 & 68.1 & 78.4 & 78.5 & 83.3 & 81.6  \\
    &  \quad$-$ \textsc{cont} & 82.3 & 95.4 & 64.9 & 76.1 & 80.4 & 82.0&\quad$-$1.5 \\
   &  \quad$-$ \textsc{stop} & 84.8 & 94.9 & 62.1 & 77.3 & 70.5 & 74.4& \quad$-$4.3 \\
   &  \quad$-$ \textsc{temp}  & \underline{24.4} & \underline{61.7} & \underline{60.6} & \underline{10.5} & \underline{67.7} & \underline{68.5}& \quad\underline{$-$32.7}\\
    \bottomrule
    \end{tabular}
    }
      \caption{ The accuracy results of the representation-level ablation study using Llama models where, for example, $-$ \textsc{temp} refers to allowing attention only to content and stopword tokens. All values are presented as percentages. 
Results showing the greatest decrease from ablation are underlined.}
    \label{tab:repablation_final2_app}
    \vspace{-1em}
\end{table}

\begin{table}[t]

  \centering
  \resizebox{0.99\linewidth}{!}{
    \begin{tabular}{ll|l|rrrrrr}
    \toprule
     Models & Setting & $\triangle$Avg. & AGNews & SST2  &  TREC  & DBPedia & RTE & CB  \\
    \midrule
   \multirow{4}*{\makecell[l]{Llama 2\\ 7B}} 
      & Zero-shot  &  36.0 & 50.2 & 50.4 & 57.2 & 6.4 & 51.6 & 0.0\\
    &  \quad$+$ \textsc{cont}& \quad$+$1.9 & 0.9 & 61.0 & 50.6 & 12.9 & 48.7 & 53.2 \\
   &  \quad$+$ \textsc{stop} & \quad$+$23.4 & 49.0 & 78.1 & 54.4 & 61.6 & 65.3 & 47.9  \\
   &  \quad$+$ \textsc{temp}& \textbf{\quad$+$31.3} & 81.1 & 82.6 & 55.2 & 65.5 & 63.9 & 55.4  \\


   \midrule
    \multirow{4}*{\makecell[l]{Llama 2\\ 13B}}  
     & Zero-shot   & 53.3 & 56.2 & 90.8 & 49.0 & 7.6 & 70.0 & 46.4 \\
   
   & \quad$+$ \textsc{cont}& \quad$-$14.1 & 0.5 & 56.0 & 61.4 & 0.0 & 62.6 & 54.6  \\
   & \quad$+$ \textsc{stop} & \quad$+$9.6 & 47.2 & 76.8 & 65.2 & 65.3 & 66.5 & 56.8 \\
   & \quad$+$ \textsc{temp} &  \textbf{\quad$+$18.1} & 78.2 & 93.7 & 62.4 & 70.4 & 71.9 & 52.1 \\

   \midrule
    \multirow{4}*{\makecell[l]{Mistral\\ 7B}}  
    
     & Zero-shot  & 59.5 & 77.8 & 84.4 & 73.0 & 57.8 & 1.8 & 62.1  \\
    &  \quad$+$ \textsc{cont} & \quad$-$10.0 & 43.3 & 52.0 & 66.6 & 10.1 & 64.3 & 60.7 \\
   &  \quad$+$ \textsc{stop} &   \quad$+$18.4 & 78.9 & 92.5 & 71.6 & 81.4 & 69.9 & 72.9 \\
   &  \quad$+$ \textsc{temp} &  \textbf{\quad$+$19.7} & 81.7 & 95.9 & 63.9 & 83.3 & 77.9 & 72.5 \\

    \midrule
   \multirow{4}*{\makecell[l]{Llama 2\\ 7B}} 
 
     & Standard ICL   & 70.7 & 85.0 & 93.2 & 58.3 & 66.7 & 66.3 & 55.0\\
   
   & \quad$-$ \textsc{cont}&\quad$-$3.8 & 82.4 & 85.5 & 54.3 & 64.2 & 59.6 & 55.7    \\
   & \quad$-$ \textsc{stop} &\quad$-$2.5 & 84.8 & 88.0 & 51.7 & 65.7 & 65.8 & 53.2  \\
   & \quad$-$ \textsc{temp} &\quad\underline{$-$32.8} & 0.9 & 61.0 & 50.6 & 12.9 & 48.5 & 53.6\\

   \midrule
    \multirow{4}*{\makecell[l]{Llama 2\\ 13B}}  
     & Standard ICL  &  73.6 & 82.8 & 94.9 & 62.8 & 74.6 & 71.2 & 55.4 \\
    &  \quad$-$ \textsc{cont}&\quad$-$1.3 & 79.0 & 94.1 & 62.7 & 72.4 & 72.1 & 53.6 \\
   &  \quad$-$ \textsc{stop} &\quad$-$2.9 & 80.1 & 89.4 & 61.5 & 74.1 & 69.6 & 49.3 \\
   &  \quad$-$ \textsc{temp}& \quad\underline{$-$33.5} & 0.5 & 56.0 & 61.4 & 0.0 & 68.2 & 54.3 \\

   \midrule
    \multirow{4}*{\makecell[l]{Mistral\\ 7B}}  
    
     & Standard ICL  &  80.2 & 82.2 & 97.0 & 67.4 & 82.4 & 73.6 & 78.6 \\
    &  \quad$-$ \textsc{cont} &\quad$-$0.6 & 81.8 & 96.2 & 64.4 & 83.4 & 78.9 & 72.9 \\
   &  \quad$-$ \textsc{stop} &\quad$-$0.8 & 81.3 & 97.0 & 66.5 & 80.5 & 75.5 & 75.7 \\
   &  \quad$-$ \textsc{temp} &\quad\underline{$-$22.8} & 78.6 & 52.0 & 66.6 & 10.1 & 67.1 & 69.6 \\

    \bottomrule
    \end{tabular}
    }

\caption{The accuracy results of the representation-level ablation study using Llama 2 and Mistral models where, for example, $+$\textsc{temp} refers to allowing attention only to template tokens and  $-$\textsc{temp} refers to allowing attention only to content and stopword tokens. All values are presented as percentages. Results are averaged over 5 different random seeds. The best results are in bold and the results showing the greatest decrease during the ablation are underlined.}
    \label{tab:repablation_morellms_app}
\end{table}

\section{Experiment Results with Instruction-tuned Large Language Models}
\label{app:sft_models}
We present the representation-level ablation results of large language models after instruction tuning to confirm that our findings remain consistent. Specifically, we use Llama 2 7B Chat and Llama 2 13B Chat in our experiments. As shown in Table~\ref{tab:representation_level_sft_models}, the results align with the findings discussed in Section~\ref{sec:ablation_representation}, with only one exception: the TREC dataset. In this dataset, the input data is structured in a question-answering format (e.g., Who/What/When did ...). We hypothesize that, during the supervised fine-tuning process, tokens associated with these question formats may also serve as performance-critical tokens, although they are currently categorized as content tokens in our experiments. Overall, these supplemental results prove further evidence to our findings in the main paper.

\begin{table}[t]

  \centering
  \resizebox{0.99\linewidth}{!}{
    \begin{tabular}{ll|r|rrrrrr}
    \toprule
     Models & Setting & Avg. & AGNews & SST2  &  TREC  & DBPedia & RTE & CB  \\
    \midrule
   \multirow{4}*{\makecell[l]{Llama 2\\ 7B Chat}} 
      & Zero-shot  &  -& -&-&-&-&-&- \\
    &  \quad$+$ \textsc{cont}&  27.9 & 0.7 & 52.6 & 52.2 & 7.9 & 24.8 & 28.9 \\
   &  \quad$+$ \textsc{stop} &72.9 & 77.1 & 90.6 & 60.2 & 75.4 & 66.7 & 67.5  \\
   &  \quad$+$ \textsc{temp}& \textbf{75.6} & \textbf{80.1} & \textbf{92.6} & \textbf{62.6} & \textbf{76.6} & \textbf{69.5} & \textbf{72.1}  \\


   \midrule
    \multirow{4}*{\makecell[l]{Llama 2\\ 13B Chat}}  
     & Zero-shot   &   -& -&-&-&-&-&- \\
   
   & \quad$+$ \textsc{cont}&  38.4 & 0.0 & 55.7 & \textbf{67.3} & 0.5 & 63.5 & 43.2  \\
   & \quad$+$ \textsc{stop} & 67.1 & 78.5 & 87.6 & 66.9 & 71.4 & 67.2 & 31.1 \\
   & \quad$+$ \textsc{temp} &  \textbf{72.3} & \textbf{82.0} & \textbf{93.7} & 65.6 & \textbf{72.3} & \textbf{72.3}& \textbf{47.9} \\

   \midrule

   \multirow{4}*{\makecell[l]{Llama 2\\ 7B Chat}} 
 
     & Standard ICL   &   -& -&-&-&-&-&- \\
   
   & \quad$-$ \textsc{cont}&76.1 & 80.7 & 93.1 & 62.9 & 76.8 & 70.7 & 72.5    \\
   & \quad$-$ \textsc{stop} &76.4 & 81.7 & 94.4 & 61.9 & 74.9 & 71.0 & 74.3 \\
   & \quad$-$ \textsc{temp} &\underline{31.4} & \underline{0.7} & \underline{52.6} & \underline{52.2} & \underline{7.9} & \underline{24.1} & \underline{51.1}  \\

   \midrule
    \multirow{4}*{\makecell[l]{Llama 2\\ 13B Chat}}  
     & Standard ICL  &  -& -&-&-&-&-&-  \\
    &  \quad$-$ \textsc{cont}&74.7 & 83.4 & 93.6 & 66.4 & 74.3 & 74.6 & 56.1\\
   &  \quad$-$ \textsc{stop} &69.8 & 78.9 & 94.1 & \underline{57.7} & 72.8 & 70.1 & 45.4 \\
   &  \quad$-$ \textsc{temp}& \underline{38.7} & \underline{0.0} & \underline{55.7} & 67.3 & \underline{0.5} & \underline{65.1} & \underline{43.6} \\


    \bottomrule
    \end{tabular}
    }

\caption{The accuracy results of the representation-level ablation study using supervised finetuned (SFT) version of Llama 2 models where, for example, $+$\textsc{temp} refers to allowing attention only to template tokens and  $-$\textsc{temp} refers to allowing attention only to content and stopword tokens. All values are presented as percentages. Results are acquired with 15 different random seeds. The best results are in bold and the results showing the greatest decrease during the ablation are underlined.}
    \label{tab:representation_level_sft_models}
\end{table}

\section{Stopword Tokens}
\label{app:stopword}
For the results shown in the main paper, we used the stopword token \citep{sarica2021stopwords} list shown in Table~\ref{tab:stopwords}. This list only includes the stopword tokens from the task instruction, aiming to minimize their presence. We made this choice under the assumption that task-affecting information should be stored densely in a few tokens. Hence, the number of tokens whose representations affect the final task performance significantly  should be small.

Nevertheless, one might be curious about the results if we used a more complete stopword list. In this case, we utilize a more comprehensive stopword token list of NLTK\footnote{\url{https://gist.github.com/sebleier/554280}} shown in Table~\ref{tab:stopwords_nltk} and conduct the representation-level ablation once more. The results are presented in Table~\ref{tab:repablation_morestop}. It can be observed that all the conclusions from Section~\ref{sec:ablation_representation} are still well established. A few results are different from Table~\ref{tab:repablation_final1} because we masked the representations of the ``(/s)'' token in this set of experiments. We claim that this masking does not impact the main findings of these experiments.

\begin{table}[t]

  \centering
  \resizebox{0.99\linewidth}{!}{
    \begin{tabular}{llrrrrrrrrrrrrrrr}
    \toprule
     Models & Setting & AGNews & SST2  &  TREC  & DBPedia & RTE & CB  \\
    \midrule
   \multirow{7}*{\makecell[l]{OpenLlama\\ 3B}}

     & Zero-shot &22.0 & 20.0 & 23.6 & 5.4 & 44.4 & 1.8 \\
   & \quad + \textsc{cont} & 26.2 & 52.1 & 30.1 & 7.4 & 51.9 & 37.9  \\
   &\quad + \textsc{stop} &38.0 & 85.1 & \textbf{31.6} & 54.6 & \textbf{58.8} & \textbf{55.7}  \\
   &\quad + \textsc{temp} &\textbf{56.5} & \textbf{86.7} & 27.1 & \textbf{62.2} & 56.4 & 52.3 \\
   \cmidrule(r{4pt}){2-8}
   & Standard ICL & 63.7 & 91.2 & 21.9 & 61.9 & 57.4 & 52.0 \\
   &   \quad - \textsc{temp} & 42.1 & 87.2 & 25.9 & 56.3 & \textbf{58.3} & \textbf{57.4} \\
    &   \quad - \textsc{cont} & 57.1 & 88.4 & \textbf{27.1} & \textbf{62.6} & 56.8 & 52.4\\
   & \quad - \textsc{stop} &  \textbf{61.6} & \textbf{90.7} & 24.8 & 62.2 & 56.7 & 51.9\\

   \midrule
    \multirow{7}*{\makecell[l]{Llama\\ 7B}}  
     & Zero-shot & 25.0 & 29.2 & 41.4 & 0.0 & 54.2 & 3.6 \\
    & \quad + \textsc{cont} & 32.4 & 57.9 & 42.5 & 12.5 & 55.5 & 46.1   \\
   &\quad + \textsc{stop} & 59.9 & 85.9 & 51.7 & 28.9 & 56.0 & 52.7 \\
   &\quad + \textsc{temp} & \textbf{70.8} & \textbf{90.2} & \textbf{58.4} & \textbf{66.2} & \textbf{66.3} & \textbf{73.5}  \\
   \cmidrule(r{4pt}){2-8}
& Standard ICL &  82.4 & 94.3 & 63.5 & 68.7 & 68.6 & 71.3 \\
   &  \quad - \textsc{temp} &  64.7 & 84.1 & 54.0 & 56.7 & 56.1 & 48.2  \\
    &  \quad - \textsc{cont}& 75.4 & 93.8 & 59.8 & 67.5 & 66.8 & \textbf{74.8}  \\
   & \quad - \textsc{stop}  & \textbf{81.4} & \textbf{94.2} & \textbf{60.5} & \textbf{67.9} & \textbf{67.6} & 72.1 \\

   \midrule
    \multirow{7}*{\makecell[l]{Llama\\ 13B}}  
     & Zero-shot &  59.0 & 18.0 & 37.0 & 0.0 & 0.0 & 0.0  \\
    &\quad +  \textsc{cont} &  30.6 & 52.4 & 43.8 & 13.0 & 60.2 & 45.5  \\
   &\quad + \textsc{stop} &  72.7 & 78.7 & 49.2 & 27.4 & 58.5 & 27.1\\
   &\quad + \textsc{temp} & \textbf{78.5} & \textbf{92.3} & \textbf{59.0} & \textbf{74.2} & \textbf{67.4} & \textbf{52.3}  \\
   \cmidrule(r{4pt}){2-8}
    & Standard ICL &  81.6 & 94.3 & 60.0 & 76.1 & 70.6 & 39.9 \\
   &  \quad - \textsc{temp} & 71.7 & 80.1 & 56.2 & 8.7 & 56.5 & 29.3 \\
    &  \quad - \textsc{cont}&  \textbf{79.3} & 93.4 & \textbf{60.1} & \textbf{74.1} & 68.4 & \textbf{47.6} \\
   & \quad - \textsc{stop} & 79.2 & \textbf{94.1} & 59.3 & 73.8 & \textbf{68.9} & 44.6 \\

   \midrule
    \multirow{7}*{\makecell[l]{Llama\\ 33B}}  
     & Zero-shot  &  70.2 & 88.6 & 60.6 & 30.2 & 58.1 & 19.6 \\
    & \quad + \textsc{cont} & 27.8 & 61.7 & 61.9 & 10.8 & 64.2 & 68.1 \\
   &\quad + \textsc{stop} &  74.7 & 93.6 & \textbf{66.9} & 70.8 & 69.1 & 63.8 \\
   &\quad + \textsc{temp} &  \textbf{80.6} & \textbf{95.2} & 63.1 & \textbf{71.9} &\textbf{78.7} & \textbf{84.0}  \\
   \cmidrule(r{4pt}){2-8}
& Standard ICL & 85.0 & 96.5 & 68.1 & 78.4 & 78.5 & 83.3  \\
   & \quad - \textsc{temp} &  79.5 & 93.8 & 58.5 & 62.8 & 68.0 & 68.0  \\
    &  \quad - \textsc{cont}&  82.7 & 95.9 & \textbf{62.9} & \textbf{74.1} & \textbf{79.6} & \textbf{83.1} \\
   & \quad - \textsc{stop}   &  \textbf{84.4} & \textbf{96.1} & 61.8 & 72.8 & 79.4 & 82.1 \\
    \bottomrule
    \end{tabular}
    }
\caption{The accuracy results of the representation level ablation study where we use the more complete stopword token list of NLTK. All values are presented as percentages. The best results presented by the number of ablated token types are in bold.}
    \label{tab:repablation_morestop}
\end{table}
\section{Analysis of the Next-line Tokens}
\label{app:nxtline_tokens}
In this section, we analyze the next-line token, which is ablated whenever any type of the stopword tokens or template tokens are ablated in the representation-level ablation experiments. We analyze this token by not ablating it when the these types of tokens are ablated. Results presented in Table~\ref{tab:repablation_nxt_always} demonstrate that the next-line token is an important performance-critical token, due to the fact that they improved the performance by a large margin compared to the results in Table~\ref{tab:repablation_final1}.

\begin{table}[t]

  \centering
  \resizebox{0.99\linewidth}{!}{
    \begin{tabular}{llrrrrrrr}
    \toprule
     Models & Setting & AGNews & SST2  &  TREC  & DBPedia & RTE & CB & $\triangle$Avg. \\
    \midrule
   \multirow{3}*{\makecell[l]{OpenLlama\\ 3B}} 

   & Standard ICL & 63.7 & 91.2 & 21.9 & 61.9 & 57.4 & 52.0 & 58.0 \\
   & \quad $-$ \textsc{stop} &  62.3& 91.0 & \underline{24.8}\asterisk & 62.9 & 57.1 & \underline{51.1}\asterisk& \quad$+$0.2 \\
   & \quad $-$ \textsc{temp}   &  \underline{41.9} & 87.2 & 26.0 & \underline{56.3} & 58.5 & 57.4& \quad\underline{$-$5.4} \\

   \midrule
    \multirow{3}*{\makecell[l]{Llama\\ 7B}}  

    & Standard ICL &  82.4 & 94.3 & 63.5 & 68.7 & 68.6 & 71.3& 74.8  \\
   & \quad $-$ \textsc{stop} & 80.4 & 94.6 & 61.1 & 68.0 & 67.2 & 72.0 & \quad$-$0.9  \\
   & \quad $-$ \textsc{temp} &  \underline{64.5} & \underline{84.1} & \underline{54.0} & \underline{58.0} & \underline{56.8} & \underline{54.3} &\quad \underline{$-$12.8}\\

   \midrule
    \multirow{3}*{\makecell[l]{Llama\\ 13B}}  

& Standard ICL &  81.6 & 94.3 & 60.0 & 76.1 & 70.6 & 39.9 & 70.4 \\
   &\quad $-$ \textsc{stop}  &  81.2 & 94.1 & 59.3 & 76.9 & 69.2 & 40.6& \quad$-$0.2  \\
   & \quad $-$ \textsc{temp} & \underline{74.1} & \underline{80.0} & \underline{46.5} & \underline{30.6} & \underline{58.3} & \underline{25.4} & \quad \underline{$-$17.9} \\

   \midrule
    \multirow{3}*{\makecell[l]{Llama\\ 33B}}  

& Standard ICL & 85.0 & 96.5 & 68.1 & 78.4 & 78.5 & 83.3 & 81.6 \\
   &\quad $-$ \textsc{stop} &  84.3 & 95.6& 65.7 & 77.6 & 78.6 & 81.8 & \quad$-$1.0  \\
   &\quad $-$  \textsc{temp} & \underline{76.6} & \underline{93.9}\asterisk  & \underline{61.2} & \underline{72.7} & \underline{70.3} & \underline{59.6}  & \quad\underline{$-$9.2} \\
    \midrule
   \multirow{3}*{\makecell[l]{Llama 2\\ 7B}} 
 
     & Standard ICL   &70.7 & 85.0 & 93.2 & 58.3 & 66.7 & 66.3 & 55.0 \\
   
   & \quad$-$ \textsc{stop}  & 85.5 & 92.7 & 56.4 & 66.6 & 63.6 & 57.1& \quad$-$0.4 \\
   & \quad$-$ \textsc{temp}  & 69.8 & 82.8 & 56.3 & 58.8 & 67.5 & 42.9 & \quad\underline{$-$7.7}\\

   \midrule
    \multirow{3}*{\makecell[l]{Llama 2\\ 13B}}  
     & Standard ICL  &  73.6 & 82.8 & 94.9 & 62.8 & 74.6 & 71.2 & 55.4 \\
   &  \quad$-$ \textsc{stop} & 81.2 & 94.5 & 61.2 & 73.7 & 72.0 & 53.2  & \quad$-$1.0\\
   &  \quad$-$ \textsc{temp}& 71.1 & 95.6 & 61.0 & 72.4 & 72.9 & 54.3&  \quad\underline{$-$2.4}  \\

   \midrule
    \multirow{3}*{\makecell[l]{Mistral\\ 7B}}  
    
     & Standard ICL  & 80.2 & 82.2 & 97.0 & 67.4 & 82.4 & 73.6 & 78.6  \\
   &  \quad$-$ \textsc{stop}& 81.2 & 97.3 & 65.5 & 82.0 & 77.6 & 73.9   &\quad$-$0.6 \\
   &  \quad$-$ \textsc{temp}  & 78.6 & 89.7 & 67.6 & 79.4 & 70.8 & 72.5 & \quad\underline{$-$3.8} \\

    \bottomrule
    \end{tabular}
    }

\caption{The accuracy results of the representation-level ablation study where, for example, $-$ \textsc{temp} refers to allowing attention only to content and stopword tokens. The next-line tokens are always ablated in this set of experiments. All values are presented as percentages. 
      Except where noted with $^*$, all test statistics reported correspond to p-values < 0.05. 
The results showing the greatest decrease from ablation are underlined.}
    \label{tab:repablation_nxt_always}
\end{table}

\section{Comparison Experiments with More Text Examples}
\label{app:more_examples}
In the ideal scenario, our experiments would have been conducted on the full test set. However, in practice, this is infeasible for any of the models studied in our paper due to computational resource constraints. For instance, it took 42 hours for the OpenLlama 3B model to run one round of the representation ablation experiment on the whole test set of DBPedia (i.e., one cell in Table~\ref{tab:repablation_final1} for the DBPedia column). To verify our number of test examples decision, we provide additional results where we scale up the number of test examples and observe no difference with our original experimental setup. Thus, we believe that limiting our test set sample size to 500 is a reasonable setup. We provide the test set statistics and the experiment results in Table~\ref{tab:more_example_statistics} and Table~\ref{tab:more_example_ablationrep}.

For TREC, RTE, and CB, using 500 test examples won’t affect the final results at all since their test set size is smaller than 500. We provide the results of experiments using all test examples in SST2, and 5000 test examples in AGNews and DBPedia here to prove our point that limiting our test set sample size to 500 is a reasonable compromise.
Shown in Table~\ref{tab:more_example_ablationrep}, compared to the results we show in Table~\ref{tab:repablation_final1}, the numbers are changed less than 1\% for all the results. 

\section{Results using different instruction prompts}
\label{app:different_instruction}
We conducted experiments on AGNews and DBPedia with 3 other different instructions to show that the and show the results in Table~\ref{tab:different_instruction_res1} and Table~\ref{tab:different_instruction_res2}. Based on these additional results, our conclusions remain the same, which shows that our findings are not sensitive to variations of the instruction prompt. The different instruction prompts ${\mathbf I}$  we used are shown in Table~\ref{tab:different_instruction}.
\begin{table*}[t]
  \centering

  \resizebox{0.7\linewidth}{!}{
    \begin{tabular}{lp{15cm}}
    \toprule
    Datasets & Stopwords \\
    \midrule     
   
    AGNews Ins. 1 &   Classify the text into World, Sports, Business, and Technology. \\
    \midrule
    AGNews Ins. 2 &   Classify the articles based on whether they are in the categories of World, Sports, Business, and Technology. \\
    \midrule
    AGNews Ins. 3 &  Classify the news to World, Sports, Business, and Technology. \\
    \midrule
    DBPedia Ins. 1 &  Classify the text into Company, School, Artist, Athlete, Politician, Transportation, Building, Nature, Village, Animal, Plant, Album, Film, and Book. \\
    \midrule
     DBPedia Ins. 2  &   Classify the documents into the categories of Company, School, Artist, Athlete, Politician, Transportation, Building, Nature, Village, Animal, Plant, Album, Film, and Book. \\
     DBPedia Ins. 3&  Classify the articles based on whether they are in the categories of Company, School, Artist, Athlete, Politician, Transportation, Building, Nature, Village, Animal, Plant, Album, Film, and Book. \\

    \bottomrule
    \end{tabular}
    }
  \caption{The different instruction prompts used in our experiments. ``Ins.'' represents ``Instruction''.}
    \label{tab:different_instruction}
\end{table*}
\begin{table*}[t]

  \centering
  \resizebox{0.7\linewidth}{!}{
    \begin{tabular}{lrrrlrrr}
    \midrule
        \multicolumn{8}{c}{\textbf{OpenLlama 3B}} \\ \midrule
        Setting & Ins. 1 & Ins. 2 & Ins. 3 & Setting & Ins. 1 & Ins. 2 & Ins. 3     \\ \midrule
        Zero-shot$+$\textsc{cont} & 26.5 & 26.9 & 22.1  & Standard ICL$-$\textsc{cont} &  53.1 & 55.6 & 67.9 \\ 
        Zero-shot$+$\textsc{stop} &  40.6 & 38.8 & 49.1 & Standard ICL$-$\textsc{stop} & 57.5 & 59.8 & 72.0    \\ 
        Zero-shot$+$\textsc{temp} & \textbf{51.1} & \textbf{53.7} & \textbf{67.6} & Standard ICL$-$\textsc{temp} & \underline{43.3} & \underline{42.6} & \underline{53.9} \\ \midrule
        \multicolumn{8}{c}{\textbf{Llama 7B}} \\ \midrule
        Setting & Ins. 1 & Ins. 2 & Ins. 3  & Setting & Ins. 1 & Ins. 2 & Ins. 3  \\ \midrule
        Zero-shot$+$\textsc{cont} &  31.1 & 30.2 & 35.2& Standard ICL$-$\textsc{cont} &70.6 & 78.2 & 75.2 \\ 
        Zero-shot$+$\textsc{stop} &51.4 & 63.5 & 61.8& Standard ICL$-$\textsc{stop} & 73.9 & 80.1 & 79.4\\ 
        Zero-shot$+$\textsc{temp} & \textbf{62.5} & \textbf{73.2} & \textbf{71.1} & Standard ICL$-$\textsc{temp} & \underline{59.9} & \underline{69.1} & \underline{74.0} \\ \midrule
    \end{tabular}
    }
\caption{Results of the representation-level ablation experiments with different instruction prompts for AGNews dataset. ``Ins.'' represents ``Instruction''. The best results are in bold while the results showing the greatest
decrease from ablation are underlined.}
    \label{tab:different_instruction_res1}
\end{table*}

\begin{table*}[t]

  \centering
  \resizebox{0.7\linewidth}{!}{
    \begin{tabular}{lrrrlrrr}
    \midrule
        \multicolumn{8}{c}{\textbf{OpenLlama 3B}} \\ \midrule
        Setting & Ins. 1 & Ins. 2 & Ins. 3 & Setting & Ins. 1 & Ins. 2 & Ins. 3     \\ \midrule
        Zero-shot$+$\textsc{cont} &  6.7 & 6.3 & 7.2   & Standard ICL$-$\textsc{cont} &  56.1 & 60.4 & 58.1 \\ 
        Zero-shot$+$\textsc{stop} &  43.1 & 48.3 & 40.2  & Standard ICL$-$\textsc{stop} & 58.1 & 61.4 & 59.6    \\ 
        Zero-shot$+$\textsc{temp} & \textbf{55.7} & \textbf{59.9} & \textbf{57.7}  & Standard ICL$-$\textsc{temp} & \underline{48.6} & \underline{54.0} & \underline{47.8} \\ \midrule
        \multicolumn{8}{c}{\textbf{Llama 7B}} \\ \midrule
        Setting & Ins. 1 & Ins. 2 & Ins. 3  & Setting & Ins. 1 & Ins. 2 & Ins. 3  \\ \midrule
        Zero-shot$+$\textsc{cont} &  15.0 & 15.9 & 6.8 & Standard ICL$-$\textsc{cont} &64.9 & 66.1 & 68.7 \\ 
        Zero-shot$+$\textsc{stop} &49.7 & 48.1 & 48.6& Standard ICL$-$\textsc{stop} & 66.8 & 67.6 & 69.6 \\ 
        Zero-shot$+$\textsc{temp} & \textbf{66.1} & \textbf{66.5 }& \textbf{69.0}  & Standard ICL$-$\textsc{temp} & \underline{58.9} & \underline{59.2} & \underline{61.7} \\ \midrule
    \end{tabular}
    }
\caption{Results of the representation-level ablation experiments with different instruction prompts for DBPedia dataset. ``Ins.'' represents ``Instruction''. The best results are in bold while the results showing the greatest
decrease from ablation are underlined.}
    \label{tab:different_instruction_res2}
\end{table*}

\begin{table*}[t]

  \centering
  \resizebox{0.7\linewidth}{!}{
    \begin{tabular}{lrrrlrrr}
    \midrule
        \multicolumn{8}{c}{\textbf{OpenLlama 3B}} \\ \midrule
        Settings & En-Fr & En-De & En-Da & Settings & En-Fr & En-De & En-Da   \\ \midrule
        Zero-shot$+$\textsc{cont} & 0.13 & 0.38 & 0.28 & Standard ICL$-$\textsc{cont} & 26.07 & 12.53 & 17.43  \\ 
        Zero-shot$+$\textsc{stop} & 16.68 & 9.17 & 13.09 & Standard ICL$-$\textsc{stop} & 26.19 & 12.52 & 17.29  \\ 
        Zero-shot$+$\textsc{temp} & \textbf{26.06} & \textbf{12.92} & \textbf{17.17} & Standard ICL$-$\textsc{temp} & \underline{17.38} & \underline{8.88} & \underline{12.9} \\ \midrule
        \multicolumn{8}{c}{\textbf{Llama 7B}} \\ \midrule
        Settings & En-Fr & En-De & En-Da & Settings & En-Fr & En-De & En-Da  \\ \midrule
        Zero-shot$+$\textsc{cont} & 11.76 & 13.83 & 10.18 & Standard ICL$-$\textsc{cont} & 35.39 & 24.23 & 30.08 \\ 
        Zero-shot$+$\textsc{stop} & 30.23 & 21.76 & 23.34 & Standard ICL$-$\textsc{stop} & 35.36 & 24.33 & 29.99 \\ 
        Zero-shot$+$\textsc{temp} & \textbf{35.47} & \textbf{24.34} & \textbf{30.12} & Standard ICL$-$\textsc{temp} & \underline{31.09} & \underline{21.98} & \underline{24.88} \\ \midrule
    \end{tabular}
    }
\caption{Results of the representation-level ablation for machine translation tasks. The best results are in bold while the results showing the greatest
decrease from ablation are underlined.}
    \label{tab:mtresults}
\end{table*}

\section{Representation-level ablation on Machine Translation Tasks}
\label{app:mt}
Besides the classification tasks, we also show results in the machine translation tasks to show that our findings could also be extended in text generation tasks. We used the Flores MT dataset \citep{costa2022no} to conduct this set of 4-shot machine translation experiments. The results are reported with the BLEU metric~\citep{papineni2002bleu}. We investigated three different language directions: English-to-French, English-to-Danish, and English-to-German. We used 10 random seeds for En-Fr and En-De and 15 random seeds for En-Da to randomly choose the demonstrations. 100 test examples are sampled in this set of the experiments as a computational compromise. Similar to the classification tasks, we keep the answer (i.e., target language) unablated for all the settings and ablate different kinds of tokens. Results in Table~\ref{tab:mtresults} show the consistent finding to those in Section~\ref{sec:ablation_representation}.

\section{Representation-level ablation on Question Answering Tasks}
\label{app:qa}
We show the representation-level experimental results of the question answering (QA) tasks in this section. We used Commonsense QA~\citep{talmor-etal-2019-commonsenseqa} dataset to test if the template and stopword tokens would directly affect the downstream task performance. We applied the settings of 4 in-context examples and 15 random seeds in this set of experiments. We frame the task as directly answering the questions instead of choosing one answer from the choices because the token types in this scenario are easier to be categorized.

Results shown in Table~\ref{tab:qaresults} demonstrate that our main findings, that template and stopword tokens are more likely to serve as performance-critical tokens, still hold in the QA tasks.
\begin{table*}[t]

  \centering
  \resizebox{0.7\linewidth}{!}{
    \begin{tabular}{lrrrrrrr}
    \midrule
        
        \multirow{2}*{\makecell[l]{Setting}} & \multirow{2}*{\makecell[r]{OpenLlama\\3B}} & \multirow{2}*{\makecell[r]{Llama\\7B}} & \multirow{2}*{\makecell[r]{Llama\\13B}} & \multirow{2}*{\makecell[r]{Llama\\33B}} &\multirow{2}*{\makecell[r]{Llama 2\\7B}} & \multirow{2}*{\makecell[r]{Llama 2\\13B}} & \multirow{2}*{\makecell[r]{Mistral\\7B}}  \\ 
        ~ & \\ \midrule
        Zero-shot$+$\textsc{cont} &7.42 & 16.62 & 14.49 & 19.47 & 17.51 & 17.78 & 16.64 \\ 
        Zero-shot$+$\textsc{stop} &11.71 & 21.96 & 18.98 & 23.38 & 22.13 & 22.31 & 19.16  \\ 
        Zero-shot$+$\textsc{temp} & \textbf{13.24} & 24.38 & \textbf{25.73} & \textbf{27.20} & \textbf{25.42} & \textbf{25.11} & \textbf{25.56}  \\ 
        \midrule

        Standard ICL$-$\textsc{cont} & 14.40 & 24.07 & 26.22 & 27.73 & 25.89 & 24.71 & 25.47  \\ 
        Standard ICL$-$\textsc{stop} & 11.96 & 21.84 & 21.44 & 26.93 & 24.62 & 23.09 & 23.69 \\ 
        Standard ICL$-$\textsc{temp} & \underline{6.89} & \underline{16.29} & \underline{15.31} & \underline{19.78} & \underline{18.51} & \underline{16.64} & \underline{15.51} \\ \midrule
    \end{tabular}
    }
\caption{Results of the representation-level ablation for question answering tasks. 15 random seeds are used to acquire all the experimental results. The best results are in bold while the results showing the greatest
decrease from ablation are underlined.}
    \label{tab:qaresults}
\end{table*}

\section{Representation-level Ablation based on the Token Count}
\label{app:representation_number}
One possible explanation for the performance variation when different types of tokens are ablated at the representation level is the simple fact that the number of tokens being ablated may vary. Intuitively, template and stopword tokens are far fewer in number compared to content tokens. In this section, we show the statistics of the number count of each type of tokens and include a supplementary experiment that only let the LLM attend to certain number of token representations of each type of the tokens. 

We first present the average token count for each type of token across the datasets. Token counts may vary depending on the tokenizer used by the large language models, and all statistics are shown in Table~\ref{tab:token_count}. The results indicate that the number of template and stopword tokens is much smaller than the number of content tokens, suggesting that performance variation during ablation is not solely due to differences in token type counts.

\begin{table*}[t]

  \centering
  \resizebox{0.65\linewidth}{!}{
    \begin{tabular}{llrrrrrrrrrrrrrrr}
    \toprule
     Tokenizer & Setting & Avg. & AGNews & DBPedia  &  SST2  & TREC & CB & RTE  \\
    \midrule
   \multirow{3}*{\makecell[l]{OpenLlama\\ 3B}} 
   & \textsc{cont} & 204.1 & 207.5 & 278.8 & 116.3 & 48.7 & 295.5 & 278.0  \\
   & \textsc{stop} &43.0 & 43.2 & 45.1 & 27.7 & 21.8 & 66.7 & 53.7  \\
   &\textsc{temp} &56.5 & 43.3 & 49.9 & 42.4 & 48.8 & 78.5 & 76.3 \\

   \midrule
    \multirow{3}*{\makecell[l]{Llama \& \\ Llama 2}}  
    & \textsc{cont} & 220.6 & 238.5 & 288.8 & 127.8 & 51.3 & 312.0 & 305.1   \\
   &\textsc{stop} & 45.0 & 43.1 & 46.1 & 29.9 & 21.7 & 71.1 & 57.9  \\
   & \textsc{temp} & 52.7 & 41.3 & 50.5 & 48.7 & 36.7 & 70.9 & 68.1 \\

   \midrule
    \multirow{3}*{\makecell[l]{Mistral}}  
    & \textsc{cont} &   213.6 & 225.0 & 284.7 & 123.7 & 50.1 & 304.2 & 293.7  \\
   &\textsc{stop} &   44.7 & 43.2 & 45.0 & 29.9 & 21.7 & 70.7 & 57.8 \\
   & \textsc{temp} & 54.6 & 45.3 & 53.5 & 40.5 & 40.7 & 75.0 & 72.5 \\

    \bottomrule
    \end{tabular}
    }
\caption{The token count statistics of different types of tokens. Avg. stands for the average token count for each type of tokens.}
    \label{tab:token_count}
\end{table*}

We then conduct an additional ablation experiment in which the model attends to representations from a specific number of tokens of a given type. We also include a baseline where a random subset of token representations from all prompt tokens is unmasked to the test examples. In this set of experiments, the label tokens are always included and are not counted as part of the token numbers.

Results in Table~\ref{tab:repablation_random} demonstrate that when the model is exposed to an equal number of each type of token representation, the performance consistently improves with template and stopword tokens, outperforming both content tokens and the random baseline. In contrast, models attending to the same number of content tokens consistently underperform relative to the random baseline. Additionally, all results improve when more tokens are included in the attention of test examples. This experiment further supports our claim that template and stopword tokens are more likely to serve as performance-critical tokens.

\begin{table}[t]

  \centering
  \resizebox{0.99\linewidth}{!}{
    \begin{tabular}{llrrrrrr|rrrrrrrrr}
    \toprule
     Models & Setting  &  AGNews& DBPedia & SST2  &  TREC  & CB  & RTE & Avg. \\
    \midrule
    \multicolumn{9}{c}{\textbf{10 Random Tokens}} \\
        \midrule
   \multirow{4}*{\makecell[l]{Llama 2\\ 7B}}

     & From \textsc{all}  &14.5 & 16.9 & 71.7 & 47.3 & 60.2 & 57.5 & 44.7   \\
   & From \textsc{cont} & 3.0 & 9.2 & 60.9 & 61.9 & 65.2 & 60.4 & 43.4 \\
   & From \textsc{stop} & 15.2 & 35.6 & 75.2 & 68.5 & 62.0 & 64.1 & 53.4 \\
   & From \textsc{temp} &  53.0 & 50.1 & 59.1 & 56.2 & 64.4 & 60.2 & \textbf{57.2}  \\

    \midrule
   \multirow{4}*{\makecell[l]{Mistral\\ 7B}}

     & From \textsc{all}  & 67.8 & 67.2 & 75.0 & 70.9 & 63.8 & 60.7 & 67.6 \\
   & From \textsc{cont} &  36.6 & 64.0 & 63.1 & 67.5 & 59.3 & 57.3 & 58.0 \\
   & From \textsc{stop} & 73.5 & 68.6 & 86.3 & 72.8 & 61.8 & 61.6 & 70.8  \\
   & From \textsc{temp} & 78.7 & 78.7 & 92.1 & 68.9 & 71.2 & 68.4 & \textbf{76.3} \\

    \midrule
    \multicolumn{9}{c}{\textbf{20 Random Tokens}} \\
        \midrule
   \multirow{4}*{\makecell[l]{Llama 2\\ 7B}}

     & From \textsc{all}  & 12.6 & 24.0 & 77.2 & 54.4 & 60.6 & 58.9 & 47.9   \\
   & From \textsc{cont} &  1.3 & 8.6 & 65.3 & 63.5 & 61.7 & 62.2 & 43.8 \\
   & From \textsc{stop} & 28.3 & 45.5 & 84.5 & 66.8 & 58.0 & 67.1 & 58.4\\
   & From \textsc{temp} &  75.0 & 63.7 & 60.3 & 59.0 & 65.4 & 60.7 & \textbf{64.0}   \\

    \midrule
   \multirow{4}*{\makecell[l]{Mistral\\ 7B}}

     & From \textsc{all}  & 78.4 & 67.8 & 74.0 & 70.1 & 65.4 & 62.0 & 69.6 \\
   & From \textsc{cont} & 69.1 & 63.3 & 59.3 & 68.2 & 60.5 & 56.8 & 62.9 \\
   & From \textsc{stop} & 78.4 & 74.3 & 89.9 & 74.2 & 68.3 & 69.1 & 75.7  \\
   & From \textsc{temp} &81.9 & 78.7 & 93.0 & 68.9 & 72.4 & 71.4 & \textbf{77.7} \\

       \midrule
    \multicolumn{9}{c}{\textbf{30 Random Tokens}} \\
        \midrule
   \multirow{4}*{\makecell[l]{Llama 2\\ 7B}}

     & From \textsc{all}  & 27.4 & 27.7 & 77.9 & 54.9 & 63.9 & 62.2 & 52.3   \\
   & From \textsc{cont} &  1.3 & 7.6 & 77.9 & 66.3 & 64.2 & 63.5 & 46.8 \\
   & From \textsc{stop} & 44.8 & 59.8 & 92.8 & 68.3 & 53.2 & 69.3 & 64.7  \\
   & From \textsc{temp} & 76.9 & 66.1 & 61.0 & 59.6 & 67.9 & 59.8 & \textbf{65.2}  \\

    \midrule
   \multirow{4}*{\makecell[l]{Mistral\\ 7B}}

     & From \textsc{all}  &79.9 & 74.6 & 83.0 & 67.8 & 69.0 & 64.6 & 73.2 \\
   & From \textsc{cont} & 71.0 & 62.6 & 58.2 & 67.2 & 59.2 & 59.4 & 62.9 \\
   & From \textsc{stop} &  79.8 & 76.9 & 91.0 & 74.2 & 71.1 & 68.8 & 77.0  \\
   & From \textsc{temp} &84.0 & 79.3 & 93.8 & 68.9 & 75.6 & 75.4 & \textbf{79.5} \\

       \midrule
    \multicolumn{9}{c}{\textbf{40 Random Tokens}} \\
        \midrule
   \multirow{4}*{\makecell[l]{Llama 2\\ 7B}}

     & From \textsc{all}  & 45.2 & 27.2 & 77.9 & 54.9 & 58.9 & 62.3 & 54.4   \\
   & From \textsc{cont} &  0.9 & 9.1 & 89.5 & 66.1 & 62.3 & 62.4 & 48.4  \\
   & From \textsc{stop} & 49.2 & 63.7 & 94.0 & 68.7 & 53.7 & 70.8 & \textbf{66.7}  \\
   & From \textsc{temp} &77.9 & 66.1 & 62.9 & 61.5 & 68.1 & 60.1 & 66.1  \\

    \midrule
   \multirow{4}*{\makecell[l]{Mistral\\ 7B}}

     & From \textsc{all}  &78.8 & 73.4 & 86.9 & 69.2 & 69.3 & 67.4 & 74.2 \\
   & From \textsc{cont} & 74.1 & 59.4 & 55.8 & 68.0 & 61.0 & 59.9 & 63.0\\
   & From \textsc{stop} &  80.1 & 76.4 & 91.0 & 74.2 & 72.0 & 69.6 & 77.2  \\
   & From \textsc{temp} &85.0 & 80.7 & 93.8 & 69.0 & 76.4 & 76.2 & \textbf{80.2}\\
    \bottomrule
    \end{tabular}
    }
\caption{The accuracy results of the representation level ablation study where we only include fixed number of certain type of tokens. All values are presented as percentages. The best results presented by the number of ablated token types are in bold. Avg. stands for the average performance. \textsc{all} represents all types of tokens}
    \label{tab:repablation_random}
\end{table}

\section{Results of the Token-level Ablation}
\label{app:token_ablation}
Detailed results of the token-level ablation are shown in Table~\ref{tab:tokenablation}. We omited the $-$\textsc{temp} case from here since it constantly yields an accuracy of 0\% when both the template token in the demonstrations and the test examples are ablated. Since the setting for the template tokens are not aligned with the ones for the stopword and content tokens, we included another set of experiments where only the the template tokens in the demonstrations are ablated at the token level in Appendix~\ref{app:token_level_template}. We want to emphasize that this experimental design choice does not affect the main findings in Section~\ref{sec:ablation_tokens}, where information is being propagated from the representations of content tokens to the representations of the performance-critical tokens and this incorporation of the information makes them better at affecting task performance, partially explaining the working mechanism of in-context learning.

\begin{table}[t]

  \centering
  \resizebox{0.99\linewidth}{!}{
    \begin{tabular}{llrrrrrr|rrrrrrrrrrrr}
    \toprule
     Models & Settings & AGNews  & SST2  & TREC  &  DBPedia  &  RTE  & CB & Avg. \\
    \midrule
    \multirow{3}*{\makecell[l]{OpenLlama\\ 3B}}
    & Standard ICL & 63.7 & 91.2 & 21.9 & 61.9 & 57.4 & 52.0 & 58.0 \\
    & \quad $-$ \textsc{cont} & 31.5 & 63.0 & 40.6 & 25.4 & 56.1 & 48.9 & 44.3  \\
    & \quad $-$ \textsc{stop} &64.4 & 91.5 & 20.9 & 62.3 & 57.8 & 52.6 & 58.3 \\
    \midrule
    \multirow{3}*{\makecell[l]{Llama\\ 7B}}
    & Standard ICL &  82.4 & 94.3 & 63.5 & 68.7 & 68.6 & 71.3  & 74.8 \\
    & \quad $-$ \textsc{cont} &55.2 & 67.2 & 42.6 & 50.8 & 57.4 & 56.3 & 54.9 \\
    &\quad $-$ \textsc{stop} &82.3 & 93.8 & 64.1 & 69.7 & 66.5 & 70.0 & 74.4\\
    \midrule
    \multirow{3}*{\makecell[l]{Llama\\ 13B}}
    & Standard ICL &  81.6 & 94.3 & 60.0 & 76.1 & 70.6 & 39.9 & 70.4 \\
    &\quad $-$ \textsc{cont} & 78.8 & 81.7 & 45.3 & 75.1 & 55.1 & 54.5 & 65.1  \\
    & \quad $-$ \textsc{stop} & 82.5 & 92.5 & 61.5 & 76.5 & 69.6 & 40.5 & 70.5\\
    \midrule
    \multirow{3}*{\makecell[l]{Llama\\ 33B}}
    & Standard ICL & 85.0 & 96.5 & 68.1 & 78.4 & 78.5 & 83.3  & 81.6\\
    &\quad $-$ \textsc{cont} &  74.0 & 89.6 & 67.0 & 73.0 & 69.8 & 49.0 & 70.4 \\
    &\quad $-$ \textsc{stop} & 85.3 & 96.4 & 66.9 & 77.9 & 77.7 & 81.3 & 80.9 \\
    \bottomrule
    \end{tabular}
    }
\caption{Results of the token-level ablation where, for example, $-$\textsc{stop} refers to the ablation where stopword tokens are dropped from the ICL prompt. 
Models without template tokens consistently yielded an accuracy of 0\% and are thus omitted from this table.}
    \label{tab:tokenablation}
\end{table}

\section{Token-level Ablation for Template Tokens}
\label{app:token_level_template}
To maintain consistency of the templates across both demonstrations and test examples, we choose to ablate the template tokens at the token level in both in Section~\ref{sec:ablation_tokens}. This experimental design differs from the other two token-level ablations. This inconsistency does not impact the main findings in Section~\ref{sec:ablation_tokens}, which show that information is propagated from the representations of content tokens to the representations of performance-critical tokens and this information aggregation enhances the ability of performance-critical tokens to improve the final task performance, partially explaining the mechanism of in-context learning. For completeness, we provide a supplemental experiment in this section where only the template tokens in the demonstrations are ablated.

Results in Table~\ref{tab:tokenablation_template} demonstrate that, although not all values reduce to 0\%, large language models perform significantly worse than in the standard in-context learning case and the other two ablation scenarios after the removal of template tokens from the demonstrations except for a few rare cases. This further supports the finding that template tokens are likely important as performance-critical tokens.

\begin{table}[t]

  \centering
  \resizebox{0.8\linewidth}{!}{
    \begin{tabular}{llr|rrrrrrrrrrrrrrrrr}
    \toprule
     Models & Settings& Avg. & AGNews  & SST2  & TREC  &  DBPedia  &  RTE  & CB  \\
    \midrule
    \multirow{2}*{\makecell[l]{OpenLlama\\ 3B}}
    & Standard ICL& 58.0 & 63.7 & 91.2 & 21.9 & 61.9 & 57.4 & 52.0  \\
    & \quad $-$ \textsc{temp} & 17.4& 0.0 & 24.2 & 40.5 & 0.4 & 35.6 & 3.6   \\
    \midrule
    \multirow{2}*{\makecell[l]{Llama\\ 7B}}
    & Standard ICL& 74.8 &  82.4 & 94.3 & 63.5 & 68.7 & 68.6 & 71.3   \\
    & \quad $-$ \textsc{temp}& 29.0  &41.0 & 11.4 & 39.9 & 0.9 & 49.7 & 31.3 \\
        \midrule
    \multirow{2}*{\makecell[l]{Llama\\ 13B}}
    & Standard ICL &70.4 &  81.6 & 94.3 & 60.0 & 76.1 & 70.6 & 39.9  \\
    &\quad $-$ \textsc{temp}& 36.9 & 82.1 & 17.5 & 51.4 & 8.6 & 39.5 & 22.0   \\
    \midrule
    \multirow{2}*{\makecell[l]{Llama\\ 33B}}
    & Standard ICL  & 81.6 & 85.0 & 96.5 & 68.1 & 78.4 & 78.5 & 83.3 \\
    &\quad $-$ \textsc{temp}& 63.0 & 73.5  & 82.8 & 58.8& 39.9 & 66.4 &56.7    \\
    \bottomrule
    \end{tabular}
    }
\caption{Results of the token-level ablation where $-$\textsc{temp} refers to the ablation where template tokens are dropped from the ICL demonstration prompt. }
    \label{tab:tokenablation_template}
\end{table}
\section{Possible Applications}
\label{app:discussion}
In this section, we discuss several potential applications that could benefit from the findings in our work. These include \textbf{long sequence processing}, where our insights can help models handle longer contexts more efficiently; \textbf{in-context learning with more demonstrations}, enabling the inclusion of additional examples without compromising performance; \textbf{better ICL prompt designing and engineering}, improving the creation of more effective prompts; and \textbf{improving model robustness}, ensuring consistent performance despite prompt variations. Each of these areas can be enhanced by understanding the role of performance-critical tokens in large language models.

\paragraph{Long sequence processing} As discussed in our paper, we hypothesize that stopword tokens tend to function as performance-critical tokens by encapsulating the semantics of preceding tokens. This finding suggests an opportunity to improve the efficiency of modeling longer sequences by selectively deleting or compressing certain hidden states during the encoding and generation stages of large language models (LLMs). Specifically, by retaining only the essential performance-critical representations while reducing unnecessary content from less informative tokens, models could manage longer inputs and outputs without compromising performance. This approach not only conserves computational resources but also addresses token length limitations in LLMs, allowing for extended sequence processing and potentially more nuanced learning from longer contexts.

This area is indeed attracting increased research attention, and our findings could contribute valuable insights into ongoing work on efficient sequence modeling and memory management in LLMs~\citep{NEURIPS2023_a452a7c6, zhang2023h_2o, bai2024citrus}. By identifying which tokens retain critical task-related information, our work aligns with and can inform methods focused on compressing intermediate states and improving long-context processing for applications ranging from summarization and document understanding to interactive dialogue systems.

\paragraph{In-context learning with more demonstrations} Given our findings, there is a promising avenue for improving in-context learning (ICL) performance by including a greater number of examples in ICL prompts~\citep{li2023incontext, hao2022structured, bertsch2024incontext}. Our results suggest that only a subset of token representations, specifically performance-critical tokens, play a critical role in determining ICL performance, while the representations of other tokens are less impactful. This observation opens up the possibility of selectively compressing or omitting unimportant token representations after the initial encoding of a demonstration. By doing so, it becomes feasible to maximize the use of the model’s fixed-length capacity, potentially enabling the inclusion of a higher number of examples within the same prompt length constraints. This approach may enhance the effectiveness of ICL in tasks where the availability of diverse examples contributes to improved model accuracy and stability.

\paragraph{Better ICL prompt designing and engineering} Our investigation into which components of ICL prompts are most critical for task performance is worthwhile and useful for directing where to put effort into tuning or improving prompts. Furthermore, the exploration on the characteristics of performance-critical tokens are useful for future design choices in ICL prompting, and help the field understand why some prompts work better than others for ICL. For instance, knowing that template and stopword tokens are particularly performance-critical allows developers to optimize prompts by focusing on specific token structures or repetitions that are most influential. This insight can improve task performance consistency across variations in prompt phrasing and structure, ultimately making prompt creation more efficient and predictable.

\paragraph{Improving Model Robustness} The findings in our study can also inform techniques to enhance the robustness of large language models (LLMs). Since prompt sensitivity (e.g., to token arrangement) can often lead to fluctuations in performance, understanding performance-critical tokens helps mitigate these vulnerabilities. By aligning model training and prompt engineering to leverage performance-critical token characteristics, it becomes possible to minimize performance drops due to minor prompt alterations, thereby enhancing the stability and reliability of LLMs in production environments.

\section{Results of Representation-level Partial performance-critical Token Ablation}
\label{app:full_results_individual}
The full results on all the six datasets are shown in Table~\ref{tab:partial} and Table~\ref{tab:partial_noanswer}. Most of the results align with our descriptions in Section~~\ref{app:different_task_encoding}, where the performance-critical tokens \textbf{should be utilized together} to provide the best performance and that removing some of them would cause performance degeneration, demonstrated by the performance decrease from Table~\ref{tab:partial}, or instability issues, shown by Table~\ref{tab:partial_noanswer}.

\begin{table*}[t]
\vspace{0em}
  \centering

  \resizebox{0.99\linewidth}{!}{
    \begin{tabular}{llrrrrrrrrrrrrrrrrrr}
    \toprule
      \multirow{2}*{Models} & \multirow{2}*{Settings} & \multicolumn{2}{c}{AGNews} & \multicolumn{2}{c}{SST2} & \multicolumn{2}{c}{TREC} & \multicolumn{2}{c}{DBPedia}  &\multicolumn{2}{c}{RTE} & \multicolumn{2}{c}{CB}  \\ 
      \cmidrule(r{4pt}){3-4} \cmidrule(r{4pt}){5-6} \cmidrule(r{4pt}){7-8}\cmidrule(r{4pt}){9-10}\cmidrule(r{4pt}){11-12} \cmidrule(r{4pt}){13-14}
       & & \textbf{with ``:''} & \textbf{w/o ``:''}& \textbf{with ``:''} & \textbf{w/o ``:''}  & \textbf{with ``:''} & \textbf{w/o ``:''} & \textbf{with ``:''} & \textbf{w/o ``:''} & \textbf{with ``:''} & \textbf{w/o ``:''} & \textbf{with ``:''} & \textbf{w/o ``:''}  \\
    \midrule
        \multirow{3}*{\makecell[l]{OpenLlama\\ 3B}}
    & \textsc{temp} with $ {\mathbf D}^{\rm out}$  &  56.5 & 47.4 & 86.7 & 83.7 & \underline{27.1} & 26.5 & 62.2 & 59.8 & \underline{56.4} & \underline{56.0} & \underline{52.3} & 56.1  \\
    & \quad $- {\mathbf T}^{\rm in}$  &  50.3 & 47.1 & \underline{85.7} & 84.4 & 28.9 & 24.4 & 57.7 & 57.7 & 56.5 & 56.1 & 53.2 & \underline{55.2}  \\
    & \quad $- {\mathbf T}^{\rm out}$ &  \underline{34.6} & \underline{32.7} & 86.9 & \underline{82.3} & 28.2 & \underline{31.2} & \underline{55.5} & \underline{54.1} & 58.3 & 59.2 & 55.4 & 58.3  \\
    \midrule
    \multirow{3}*{\makecell[l]{Llama\\ 7B}}
    
    & \textsc{temp} with ${\mathbf D}^{\rm out}$  &  70.8 & 57.3 & 90.2 & 87.1 & 58.4 & 46.7 & 66.2 & 63.8 & 66.3 & 59.5 & 73.5 & 69.6  \\
    & \quad $- {\mathbf T}^{\rm in}$ & 62.7 & 55.1 & 91.6 & 87.1 & 52.8 & \underline{43.3} & 61.6 & 61.8 & 58.9 & \underline{56.5} & \underline{59.2} & \underline{55.7} \\
    &  \quad $- {\mathbf T}^{\rm out}$  & \underline{50.8} & \underline{48.6} & \underline{84.9} & \underline{82.8} & \underline{46.0} & 50.2 & \underline{57.9} & \underline{55.2} & \underline{56.7} & 56.7 & 66.2 & 64.5  \\
    \midrule
    \multirow{3}*{\makecell[l]{Llama\\ 13B}}
    & \textsc{temp} with ${\mathbf D}^{\rm out}$  &  80.0 & 76.2 & 92.3 & 89.1 & 58.6 & 54.0 & 76.9 & 71.4 & 68.5 & 59.8 & 47.7 & 35.0 \\
    &   \quad $- {\mathbf T}^{\rm in}$ &  79.9 & 76.3 & 91.5 & 88.9 & 55.1 & \underline{47.8} & 75.8 & 70.7 & 65.5 & 59.0 & \underline{35.7} & \underline{24.5} \\
    &  \quad $- {\mathbf T}^{\rm out}$  & \underline{72.0} & \underline{72.1} & \underline{81.1} & \underline{75.9} & \underline{47.1} & 48.3 & \underline{60.3} & \underline{35.5} & \underline{61.2} & \underline{58.4} & 36.2 & 36.0  \\
    \midrule
    \multirow{3}*{\makecell[l]{Llama\\ 33B}}
    & \textsc{temp} with ${\mathbf D}^{\rm out}$ &  80.5 & 75.0 & 95.2 & 93.3 & 65.2 & 66.7 & 75.2 & 73.5 & 79.0 & 77.1 & 80.0 & 70.7  \\
    &  \quad $- {\mathbf T}^{\rm in}$ &  78.7 & 71.5 & 95.2 & \underline{92.8} & 68.1 & 67.7 & 75.1 & 73.8 & 77.4 & 75.4 & 73.3 & \underline{62.3}  \\
    & \quad $- {\mathbf T}^{\rm out}$   &   \underline{69.2} & \underline{69.5} & \underline{93.9} & 92.9 & \underline{62.1} & \underline{66.2} & \underline{71.3} & \underline{70.1} & \underline{72.8} & \underline{70.0} & \underline{67.4} & 63.5   \\

    \bottomrule
    \end{tabular}
    }
        \caption{Ablation for different template token representations with the answer label token representations, presented as percentages. The results showing the greatest impact from ablation are underlined. }
    \label{tab:partial}
\end{table*}

\begin{table*}[t]

  \centering
  \resizebox{0.99\linewidth}{!}{
    \begin{tabular}{llrrrrrrrrrrrrrrrrrr}
    \toprule
      \multirow{2}*{Models} & \multirow{2}*{Settings} & \multicolumn{2}{c}{AGNews} & \multicolumn{2}{c}{SST2} & \multicolumn{2}{c}{TREC} & \multicolumn{2}{c}{DBPedia}  &\multicolumn{2}{c}{RTE} & \multicolumn{2}{c}{CB}  \\ 
      \cmidrule(r{4pt}){3-4} \cmidrule(r{4pt}){5-6} \cmidrule(r{4pt}){7-8}\cmidrule(r{4pt}){9-10}\cmidrule(r{4pt}){11-12} \cmidrule(r{4pt}){13-14}
       & & \textbf{with ``:''} & \textbf{w/o ``:''}& \textbf{with ``:''} & \textbf{w/o ``:''}  & \textbf{with ``:''} & \textbf{w/o ``:''} & \textbf{with ``:''} & \textbf{w/o ``:''} & \textbf{with ``:''} & \textbf{w/o ``:''} & \textbf{with ``:''} & \textbf{w/o ``:''}  \\
    \midrule
        \multirow{3}*{\makecell[l]{OpenLlama\\ 3B}}
    & \textsc{temp} w/o ${\mathbf D}^{\rm out}$ &   41.5 & 54.6 & \underline{14.3} & \underline{73.2} & \underline{36.5} & 42.0 & 29.4 & 21.7 & \underline{24.7} & \underline{45.7} & \underline{0.7} & \underline{3.5}  \\
    &  \quad $- {\mathbf T}^{\rm in}$  & 42.2 & 52.2 & 18.5 & 79.9 & 39.7 & 42.2 & 22.6 & 22.5 & 49.8 & 57.1 & 3.1 & 6.5   \\
    &  \quad $- {\mathbf T}^{\rm out}$ &  \underline{36.3} & \underline{35.5} & 83.0 & 83.4 & 43.2 & \underline{41.9} & \underline{16.3} & \underline{18.7} & 54.4 & 56.8 & 1.2 & 4.2   \\
    \midrule
    \multirow{3}*{\makecell[l]{Llama\\ 7B}}
    &\textsc{temp} w/o ${\mathbf D}^{\rm out}$  &  50.4 & 56.6 & 68.2 & 56.1 & 55.3 & 48.5 & 0.2 & 1.3 & \underline{40.7} & \underline{42.5} & 28.5 & \underline{18.8} \\
    &  \quad $- {\mathbf T}^{\rm in}$ &  46.1 & 50.6 & \underline{61.9} & \underline{55.1} & \underline{43.5} & \underline{44.7} & 0.0 & 0.2 & 43.7 & 49.9 & \underline{27.1} & 26.5  \\
    &  \quad $- {\mathbf T}^{\rm out}$ & \underline{21.4} & \underline{12.7} & 86.2 & 66.5 & 54.4 & 55.6 & 0.0 & 0.0 & 56.0 & 55.6 & 39.4 & 35.1   \\
    \midrule
    \multirow{3}*{\makecell[l]{Llama\\ 13B}}
    & \textsc{temp} w/o ${\mathbf D}^{\rm out}$  &  66.9 & 77.0 & \underline{65.6} & 87.9 & 51.8 & 53.1 & 0.1 & 0.1 & 57.5 & 53.7 & 16.7 & 21.9  \\
    & \quad $- {\mathbf T}^{\rm in}$  & \underline{72.9} & \underline{76.6} & 83.0 & 89.5 & \underline{45.5} & 48.4 & 0.0 & 0.0 & \underline{53.6} & \underline{52.8} & 16.0 & 20.1 \\
    & \quad $- {\mathbf T}^{\rm out}$ &    79.2 & 77.7 & 77.5 & \underline{47.5} & 56.8 & \underline{43.2} & 0.0 & 0.0 & 54.8 & 53.7 & \underline{4.3} & \underline{2.4}  \\
    \midrule
    \multirow{3}*{\makecell[l]{Llama\\ 33B}}
    & \textsc{temp} w/o ${\mathbf D}^{\rm out}$ &   77.3 & 78.2 & \underline{17.3} & 88.9 & \underline{65.4} & \underline{69.3} & 31.0 & 41.7 & 71.8 & 65.8 & 23.8 & 23.0  \\
    &  \quad $- {\mathbf T}^{\rm in}$   &  72.9 & \underline{72.4} & 29.2 & \underline{87.4} & 65.6 & 70.9 & \underline{14.9} & 37.9 & 70.6 & 67.8 & \underline{19.9} & 21.1 \\
    & \quad $- {\mathbf T}^{\rm out}$  &   \underline{69.5} & 74.3 & 92.6 & 92.8 & 70.0 & 70.8 & 42.0 & \underline{20.3} & \underline{67.0} & \underline{61.3} & 23.1 & \underline{18.5}  \\

    \bottomrule
    \end{tabular}
    }
\caption{Ablation for different template token representations without the answer label token representations. All values are presented as percentages. The results showing the greatest decrease during the ablation are underlined.}
    \label{tab:partial_noanswer}
\end{table*}

\section{Significance test for the representation-level ablation}
\label{app:significance}
In this section, we report the p-value of all the pair-wise comparisons in the representation-level ablation experiments in Table~\ref{tab:repablation_final1}. Results are shown in Table~\ref{tab:t_test}. Most of the ablation results show significant difference among different ablation scenarios.

\begin{table*}[t]
\small

  \centering
  \resizebox{0.99\linewidth}{!}{
    \begin{tabular}{llrrrrrrrrrrrrrrrrrr}
    \toprule
      \multirow{2}*{Models} & \multirow{2}*{Settings} & \multicolumn{2}{c}{AGNews} & \multicolumn{2}{c}{SST2} & \multicolumn{2}{c}{TREC} & \multicolumn{2}{c}{DBPedia}  &\multicolumn{2}{c}{RTE} & \multicolumn{2}{c}{CB}  \\ 
      \cmidrule(r{4pt}){3-4} \cmidrule(r{4pt}){5-6} \cmidrule(r{4pt}){7-8}\cmidrule(r{4pt}){9-10}\cmidrule(r{4pt}){11-12} \cmidrule(r{4pt}){13-14}
       & & P-value & p < 0.05 & P-value & p < 0.05 & P-value & p < 0.05 & P-value & p < 0.05& P-value & p < 0.05 & P-value & p < 0.05 \\
    \midrule
        \multirow{6}*{\makecell[l]{OpenLlama\\ 3B}} & temp (-) cont  & 0.0000581 & T & 0.0000000 & T & 0.1952165 & F & 0.0000000 & T & 0.0042427 & T & 0.0027293 & T  \\ 
        ~ & temp (-) stop  & 0.0001605 & T & 0.0571278 & F & 0.0242797 & T & 0.0000663 & T & 0.0319815 & T & 0.0985942 & F  \\ 
        ~ & cont (-) stop & 0.0023957 & T & 0.0000001 & T & 0.1940792 & F & 0.0000000 & T & 0.0000073 & T & 0.0000698 & T  \\ 
        ~ & temp\_cont (-) cont\_stop  & 0.0000065 & T & 0.0385760 & T & 0.3206221 & F & 0.0000000 & T & 0.1237570 & F & 0.0544049 & F  \\ 
        ~ & temp\_stop (-) cont\_stop & 0.0001166 & T & 0.4514005 & F & 0.2549225 & F & 0.0000001 & T & 0.0545474 & F & 0.0534521 & F  \\ 
        ~ & temp\_cont (-) temp\_stop & 0.0000507 & T & 0.0096752 & T & 0.0005775 & T & 0.0000000 & T & 0.1208140 & F & 0.3193696 & F  \\ 
        \midrule
        \multirow{6}*{\makecell[l]{Llama\\ 7B}}& temp (-) cont  & 0.0000000 & T & 0.0000001 & T & 0.0000020 & T & 0.0000001 & T & 0.0000004 & T & 0.0000001 & T  \\ 
        ~ & temp (-) stop  & 0.0000083 & T & 0.0101283 & T & 0.0002883 & T & 0.1438193 & F & 0.0000000 & T & 0.0000031 & T  \\ 
        ~ & cont (-) stop & 0.0000060 & T & 0.0000001 & T & 0.0019529 & T & 0.0000001 & T & 0.3392237 & F & 0.0016487 & T  \\ 
        ~ & temp\_cont (-) cont\_stop  & 0.0000115 & T & 0.0030175 & T & 0.0005950 & T & 0.0000000 & T & 0.0000000 & T & 0.0001649 & T  \\ 
        ~ & temp\_stop (-) cont\_stop & 0.0002004 & T & 0.0227328 & T & 0.0015468 & T & 0.0000000 & T & 0.0000001 & T & 0.0000094 & T  \\ 
        ~ & temp\_cont (-) temp\_stop & 0.0086396 & T & 0.0003632 & T & 0.0089932 & T & 0.0000007 & T & 0.1637608 & F & 0.1553081 & F  \\ 
        \midrule
        \multirow{6}*{\makecell[l]{Llama\\ 13B}} & temp (-) cont  & 0.0000000 & T & 0.0000000 & T & 0.0000082 & T & 0.0000001 & T & 0.0006445 & T & 0.1060226 & F  \\ 
        ~ & temp (-) stop  & 0.0003841 & T & 0.0000012 & T & 0.0034370 & T & 0.0002018 & T & 0.0000000 & T & 0.0010178 & T  \\ 
        ~ & cont (-) stop & 0.0000000 & T & 0.0000202 & T & 0.0002820 & T & 0.0000000 & T & 0.0098209 & T & 0.0022848 & T  \\ 
        ~ & temp\_cont (-) cont\_stop  & 0.0010838 & T & 0.0000730 & T & 0.0004557 & T & 0.0000048 & T & 0.0000001 & T & 0.0002364 & T  \\ 
        ~ & temp\_stop (-) cont\_stop & 0.0007763 & T & 0.0000310 & T & 0.0016544 & T & 0.0000000 & T & 0.0000000 & T & 0.0000888 & T  \\ 
        ~ & temp\_cont (-) temp\_stop & 0.4411518 & F & 0.1158895 & F & 0.3323328 & F & 0.0000000 & T & 0.3148253 & F & 0.0144961 & T  \\ 
        \midrule
        \multirow{6}*{\makecell[l]{Llama\\ 33B}}  & temp (-) cont  & 0.0000000 & T & 0.0000003 & T & 0.1534319 & F & 0.0000000 & T & 0.0000000 & T & 0.0002244 & T  \\ 
        ~ & temp (-) stop  & 0.0007359 & T & 0.0048547 & T & 0.1797405 & F & 0.0000023 & T & 0.0000002 & T & 0.0008789 & T  \\ 
        ~ & cont (-) stop & 0.0000000 & T & 0.0000003 & T & 0.0204911 & T & 0.0000000 & T & 0.0002626 & T & 0.4319440 & F  \\ 
        ~ & temp\_cont (-) cont\_stop  & 0.0001365 & T & 0.0788756 & F & 0.0032131 & T & 0.0000000 & T & 0.0000098 & T & 0.0003242 & T  \\ 
        ~ & temp\_stop (-) cont\_stop & 0.0006045 & T & 0.0609501 & F & 0.0165374 & T & 0.0000011 & T & 0.0000009 & T & 0.0003821 & T  \\ 
        ~ & temp\_cont (-) temp\_stop & 0.0012936 & T & 0.3583931 & F & 0.1415489 & F & 0.0001034 & T & 0.0009055 & T & 0.3979685 & F \\ 

    \bottomrule
    \end{tabular}
    }
\caption{The pair-wise t-test significance results.  ``T'' means True while ``F'' means False. In this table, ``temp'' means only keeping temp, which is zero-shot + \textsc{temp}. ``temp\_cont'' means ablating the stopword token representations, which is Standard ICL $-$ \textsc{stop}.}

    \label{tab:t_test}
\end{table*}

\section{Template used for the random string experiments}
\label{app:random}
In this section, we present all the in-context learning templates used for the random experiments in Section~\ref{sec:experiment_id}. 
In the \textbf{Random$_{\rm fixed}$} scenario, the ${\mathbf T}^{\rm in}$ and ${\mathbf T}^{\rm out}$ are consistent across all demonstrations. For the \textbf{Random$_{\rm nonfixed}$} scenario, we employ different random string templates for each demonstration. We use 5 random string templates for each setting, shown in Table~\ref{tab:randomICLtemplate}, Table~\ref{tab:randomICLtemplate2}, Table~\ref{tab:randomICLtemplate3}, Table~\ref{tab:randomICLtemplate4}, and Table~\ref{tab:randomICLtemplate5}. The results in Section~\ref{sec:experiment_id} are averaged over the results with all the different random string templates.

\section{Supplemental Experiments for the Repetition Characteristic}
\label{app:repetitive}
In Section~\ref{sec:repeition}, we examine the repetition characteristic of performance-critical tokens with random template tokens, which could not be general enough since random string tokens are less used in real-world applications. Hence, we conduct another set of experiments in this section, using template tokens with lexical meanings to test the characteristic of repetition. 

These experiments includes two sets of comparisons shown in Table~\ref{tab:repetitive_example1} and Table~\ref{tab:repetitive_example2}. The first set of templates uses meaningful, normal words but exhibits less lexical similarity to the task. The second set of templates is more closely related to the task. All comparisons are made between non-repetitive and repetitive cases.

The results presented in Table~\ref{tab:repetition_supplement} show that, when random strings without lexical meanings are not used, the repetitive patterns can also enhance the final performances and help encode the task within the representations of template tokens, proving our claim that repetition is an important characteristic of performance-critical tokens.



\begin{table*}[t]

  \centering
  \resizebox{0.7\linewidth}{!}{
    \begin{tabular}{p{15.5cm}}
    \toprule
\multicolumn{1}{c}{Template 1}
    \\
    \midrule     
Classify the news articles into the categories of World, Sports, Business, and Technology. \\
\\
\textbf{dog:} First class to the moon. London - British airline magnate Richard Branson announced a plan on Monday for the world's first commercial space flights, saying "thousands" of fee-paying astronauts could be sent into orbit in the near future. \\
\textbf{cat:} Technology\\
\\
\textbf{juice:} Amazon's Holiday Pi. Leave it to Amazon.com (Nasdaq: AMZN). Apparently, the holiday season could be a rich opportunity to addict more users to Amazon's A9. \\
\textbf{wine:} Technology\\
\\
\textbf{sleep:} Will historic flight launch space tourism?. Regardless, space competitions are poised to become big business.\\
\textbf{wake: }Technology\\
\\
\textbf{bunny:} SMART-1 makes lunar orbit. The SMART-1 probe has entered its lunar orbit, and the history books as the first European mission to have done so. Professor David Southwood, director of science for the European Space Agency (ESA), said: "Europe ... \\
\textbf{easter:} \\

    \midrule
\multicolumn{1}{c}{Template 2} \\
    \midrule

    Classify the news articles into the categories of World, Sports, Business, and Technology.\\
\\
\textbf{dog:} First class to the moon. London - British airline magnate Richard Branson announced a plan on Monday for the world's first commercial space flights, saying "thousands" of fee-paying astronauts could be sent into orbit in the near future.\\
\textbf{cat:} Technology\\
\\
\textbf{dog:} Amazon's Holiday Pi. Leave it to Amazon.com (Nasdaq: AMZN). Apparently, the holiday season could be a rich opportunity to addict more users to Amazon's A9.\\
\textbf{cat:} Technology\\
\\
\textbf{dog:} Will historic flight launch space tourism?. Regardless, space competitions are poised to become big business.\\
\textbf{cat: }Technology \\
\\
\textbf{dog:} SMART-1 makes lunar orbit. The SMART-1 probe has entered its lunar orbit, and the history books as the first European mission to have done so. Professor David Southwood, director of science for the European Space Agency (ESA), said: "Europe... \\ 
\textbf{cat:}  \\



    \bottomrule
    \end{tabular}
    }
    \caption{A 3-shot example sampled from AGNews dataset using Template 1 and Template 2.}
    \label{tab:repetitive_example1}
\end{table*}

\begin{table*}[t]

  \centering
  \resizebox{0.7\linewidth}{!}{
    \begin{tabular}{p{15.5cm}}
    \toprule
\multicolumn{1}{c}{Template 3}
    \\
    \midrule     
Classify the news articles into the categories of World, Sports, Business, and Technology.\\
\\
\textbf{article:} First class to the moon. London - British airline magnate Richard Branson announced a plan on Monday for the world's first commercial space flights, saying "thousands" of fee-paying astronauts could be sent into orbit in the near future.\\
\textbf{answer:} Technology\\
\\
\textbf{input:} Amazon's Holiday Pi. Leave it to Amazon.com (Nasdaq: AMZN). Apparently, the holiday season could be a rich opportunity to addict more users to Amazon's A9.\\
\textbf{output:} Technology\\
\\
\textbf{text:} Will historic flight launch space tourism?. Regardless, space competitions are poised to become big business.\\
\textbf{label:} Technology\\
\\
\textbf{sentence:} SMART-1 makes lunar orbit. The SMART-1 probe has entered its lunar orbit, and the history books as the first European mission to have done so. Professor David Southwood, director of science for the European Space Agency (ESA), said: "Europe... \\
\textbf{result:} \\

    \midrule
\multicolumn{1}{c}{Template 4} \\
    \midrule

Classify the news articles into the categories of World, Sports, Business, and Technology.\\
\\
\textbf{article:} First class to the moon. London - British airline magnate Richard Branson announced a plan on Monday for the world's first commercial space flights, saying "thousands" of fee-paying astronauts could be sent into orbit in the near future.\\
\textbf{answer:} Technology\\
\\
\textbf{article:} Amazon's Holiday Pi. Leave it to Amazon.com (Nasdaq: AMZN). Apparently, the holiday season could be a rich opportunity to addict more users to Amazon's A9.\\
\textbf{answer:} Technology\\
\\
\textbf{article: }Will historic flight launch space tourism?. Regardless, space competitions are poised to become big business.\\
\textbf{answer: }Technology\\
\\
\textbf{article:} SMART-1 makes lunar orbit. The SMART-1 probe has entered its lunar orbit, and the history books as the first European mission to have done so. Professor David Southwood, director of science for the European Space Agency (ESA), said: "Europe... \\
\textbf{answer:} \\



    \bottomrule
    \end{tabular}
    }
    \caption{A 3-shot example sampled from AGNews dataset using Template 3 and Template 4.}
    \label{tab:repetitive_example2}
\end{table*}

\begin{table*}[t]

  \centering
  \resizebox{0.55\linewidth}{!}{
    \begin{tabular}{llrrr|r}
    \toprule
     Models & Setting & AGNews & DBPedia &   TREC  & $\triangle$Avg. \\
    \midrule
   \multirow{4}*{\makecell[l]{OpenLlama\\ 3B}} 

   & Template 1 & 33.56 & 9.72 & 5.84 & 16.37 \\
   & Template 2 & \textbf{55.56} & \textbf{61.44} & \textbf{22.36} & \textbf{46.45}  \\
   \cmidrule(r{4pt}){2-6}
   & Template 3  &  48.24 & 50.16 & 24.84 & 41.08 \\
      & Template 4  & \textbf{69.64} & \textbf{63.20} & \textbf{20.96} & \textbf{51.27} \\

   \midrule
    \multirow{4}*{\makecell[l]{Llama\\ 7B}}  

   & Template 1 & 6.00 & 0.24 & 12.92 & 6.39  \\
   & Template 2 & \textbf{26.52} & \textbf{51.20} & \textbf{25.32} & \textbf{34.35} \\
   \cmidrule(r{4pt}){2-6}
   & Template 3  &  19.80 & 62.28 & 1.88 & 27.99 \\
      & Template 4  & \textbf{36.44} & \textbf{64.68} & \textbf{18.68} & \textbf{39.93} \\

   \midrule
    \multirow{4}*{\makecell[l]{Llama\\ 13B}}  

   & Template 1 & 7.40 & 0.00 & 5.68 & 4.36  \\
   & Template 2 & \textbf{46.68} & \textbf{71.40} & \textbf{35.80} &\textbf{ 51.29} \\
   \cmidrule(r{4pt}){2-6}
   & Template 3  &  15.60 & \textbf{76.36} & 4.96 & 32.31 \\
      & Template 4  & \textbf{49.08} & 75.56 & \textbf{23.80} & \textbf{49.48} \\

   \midrule

   \multirow{4}*{\makecell[l]{Llama 2\\ 7B}} 
 
   & Template 1 & 52.84 & 12.08 & 35.60 & 33.51  \\
   & Template 2 & \textbf{75.60} & \textbf{82.80} & \textbf{56.04} & \textbf{71.48 } \\
\cmidrule(r{4pt}){2-6}
   & Template 3  & 32.96 & 76.56 & 7.04 & 38.85 \\
      & Template 4  &  \textbf{70.16} & \textbf{80.96} & \textbf{58.24} & \textbf{69.79}  \\

   \midrule
    \multirow{4}*{\makecell[l]{Llama 2\\ 13B}}  
   & Template 1 & 8.28 & 0.04 & 2.68 & 3.67  \\
   & Template 2 &\textbf{19.80} & \textbf{44.52} & \textbf{13.92} & \textbf{26.08} \\
   \cmidrule(r{4pt}){2-6}
   & Template 3  & 5.84 & 59.88 & 1.72 & 22.48 \\
      & Template 4  &  \textbf{28.60} & \textbf{65.04} & \textbf{12.64} & \textbf{35.43} \\

   \midrule
    \multirow{4}*{\makecell[l]{Mistral\\ 7B}}  
    
   & Template 1 & 27.68 & 0.20 & 17.96 & 15.28  \\
   & Template 2 &   \textbf{67.48 }&\textbf{ 67.60} & \textbf{31.20} & \textbf{55.43} \\
   \cmidrule(r{4pt}){2-6}
   & Template 3  &   2.64 & 47.68 & 4.04 & 18.12 \\
      & Template 4  & \textbf{59.12} & \textbf{70.64} & \textbf{39.12} & \textbf{56.29} \\

    \bottomrule
    \end{tabular}
    }
\caption{The accuracy results of the repetitive supplemental experiments.}

    \label{tab:repetition_supplement}
\end{table*}
\section{Supplemental Experiments for the Structural Cue Characteristic}
\label{app:text_formatting0}
In this section, we describe a set of supplemental experiments, which support the characteristic of structural cues from the perspective of representation-level ablation.
An intuitive method to verify the effect of the structural cues would be using the same random strings to replace ${\mathbf T}^{\rm in}$ and ${\mathbf T}^{\rm out}$, making it harder for a model to parse the structure of the text. However, this would bring the factor of repetition into the process, potentially confounding the results. Hence,
we instead design a one-shot \textbf{Random$_{\rm fixed}$} experiment. The one-shot \textbf{Random$_{\rm fixed}$} setting allows us to control both the characteristics of lexical meaning and repetition since the templates are made up of random strings and there is only one training demonstration. With these two characteristics controlled, we use the masking ablation method from Section~\ref{sec:ablation_representation} to confirm to what extent these random string tokens can function effectively as delimiters between inputs and outputs in ICL prompts. Specifically, we include results from the Zero-shot $+$ \textsc{temp}$_{\rm 1\text{-}shot}^\text{random}$ and Zero-shot $+$ ``:''$_{\rm 1\text{-}shot}^\text{random}$ scenarios, as well as the standard results of one-shot \textbf{Random$_{\rm fixed}$}, for a more comprehensive analysis, shown in Table~\ref{tab:structural}. Examples of all the different model variants are shown in Appendix~\ref{app:text_formatting}.

\begin{table*}
  \centering

  \resizebox{0.7\linewidth}{!}{
    \begin{tabular}{llrrrrrr|r}
    \toprule

    Models & Settings & AGNews & SST2  & TREC & DBPedia & RTE & CB & Avg. \\

        \midrule
    \multirow{3}*{\makecell[l]{OpenLlama\\ 3B}}
    & One-shot Random$_{\rm fixed}$ &47.5 & 51.8 & 32.6 & 19.4 & 51.8 & 42.4 & 40.9 \\
    &Zero-shot$+$\textsc{temp}$_{\rm 1\text{-}shot}^\text{random}$ &39.5 & 49.8 & 27.7 & 13.3 & 49.8 & 44.9 & 37.5 \\
    & Zero-shot$+$``:''$_{\rm 1\text{-}shot}^\text{random}$  &  \underline{31.5} & \underline{35.9} & \underline{23.8} & \underline{8.0} & \underline{35.9} & \underline{33.8} & \underline{28.2}  \\
    \midrule
    \multirow{3}*{\makecell[l]{Llama\\ 7B}}
    & One-shot Random$_{\rm fixed}$&3.9 & 16.9 & \underline{3.5} & 9.6 & 16.9 & 10.4 & 10.2  \\
    & Zero-shot$+$\textsc{temp}$_{\rm 1\text{-}shot}^\text{random}$ &  \underline{2.1} & 15.5 & 7.6 & 3.7 & 15.5 & \underline{5.4}  & 8.3 \\
    &  Zero-shot$+$``:''$_{\rm 1\text{-}shot}^\text{random}$ & 3.6 & \underline{7.5} & 14.6 & \underline{3.0} & \underline{7.5} & 6.8 & \underline{7.2}  \\
    \midrule
    \multirow{3}*{\makecell[l]{Llama\\ 13B}}
    & One-shot Random$_{\rm fixed}$ & 46.1 & 47.5 & \underline{25.0} & 50.8 & 47.5 & 21.4 & 39.7 \\
    & Zero-shot$+$\textsc{temp}$_{\rm 1\text{-}shot}^\text{random}$ &29.2 & 48.9 & 36.1 & 35.7 & 48.9 & \underline{14.0} & 35.5\\
    & Zero-shot$+$``:''$_{\rm 1\text{-}shot}^\text{random}$ &  \underline{14.3} & \underline{22.4} & 25.4 & \underline{22.5} & \underline{22.4} & 28.9 & \underline{22.7} \\
    \midrule
    \multirow{3}*{\makecell[l]{Llama\\ 33B}}
    & One-shot Random$_{\rm fixed}$&  69.7 & 53.0 & 37.8 & 72.8 & 53.0 & \underline{37.6} & 54.0 \\
    & Zero-shot$+$\textsc{temp}$_{\rm 1\text{-}shot}^\text{random}$ &  61.2 & 56.3 & 41.1 & 69.2 & 56.3 & 43.0 & 54.5 \\
    &  Zero-shot$+$``:''$_{\rm 1\text{-}shot}^\text{random}$  & \underline{43.3} & \underline{41.8} & \underline{37.4} & \underline{65.0} & \underline{41.8} & 39.5  & \underline{44.8}  \\

    \bottomrule
    \end{tabular}
}
    \caption{One-shot representation masking experiments conducted to verify if structural template formats could influence the effectiveness of the performance-critical tokens.  ${\mathbf D}^{\rm out}$ is preserved in all the settings. The results showing the greatest decrease during the ablation are underlined.}
    \label{tab:structural}
\end{table*}

We observe that 
adding the attention to random template token representations in the one-shot setting often leads to performance increases while 
masking the attention to the template tokens and only attending to ``:'' $ + {\mathbf D}^{\rm out}$ leads to performance decreases. This indicates that the presence of these tokens is critical to maintaining task performance. With all other characteristics being controlled, this leads us to believe that the delimiting nature of template tokens is likely to be an important characteristic in their role as performance-critical tokens.

\section{Discussion about the Characteristic of Structural Cue}
\label{app:text_formatting}
As discussed in Section~\ref{sec:text_formattingj}, we view structural cues as the textual and structural cues present in the prompt allowing the model to distinguish between the different parts of the ICL demonstration. We believe that performance-critical tokens naturally play this role since the same types of tokens are likely to delimit pretraining text (e.g., html, markdown, etc.). An example of how we believe performance-critical tokens naturally delimit an ICL prompt is shown in Table~\ref{tab:textformatting_example}, sampled from the SST2 dataset tested in our experiments. We bold and place in brackets the role of each section of the prompt as well as what types of tokens it contains.

During LLM pre-training, it is very likely that the model has seen text formatted in a similar way (e.g. in html, plain text with headings), in which the LLM would learn to recognize and store information in the representation of these formatting tokens. A recent study also provides supporting facts from the perspective of pretraining for this~\citep{chen2024parallel}.

One possible way to examine if these structural cues are a key characteristic is to use the same template text for the input demonstration and the output label, disturbing the structure of the prompt and making it difficult to recognize the input and the output (e.g.: input: [demonstration]\textbackslash n input: [label]). However, this would bring the confounding factor of the repetition and lexical meaning when we use multiple demonstrations. Even if we only use one example, the two same templates (“input” and “input”) could form a repetition.  We therefore choose to use a one-shot Random$_{\rm fixed}$ scenario to avoid that for the experiments in Appendix~\ref{app:text_formatting0}. In this case, there are no repetition or lexical meaning confounds. 

Zero-shot $+$ \textsc{temp}$_{\rm 1\text{-}shot}^\text{random}$ and Zero-shot $+$ ``:''$_{\rm 1\text{-}shot}^\text{random}$
To investigate whether these random string tokens are working as performance-critical tokens, given that they only serve as providing structural cues, we applied the representation-level ablation to see the model’s performance when the test examples have or do not have access to the representations of these random string tokens, comparing the performance among [one-shot Random$_{\rm fixed}$], [Zero-shot $+$ \textsc{temp}$_{\rm 1\text{-}shot}^\text{random}$] and [Zero-shot $+$ ``:''$_{\rm 1\text{-}shot}^\text{random}$]. [Zero-shot $+$ \textsc{temp}$_{\rm 1\text{-}shot}^\text{random}$] and [Zero-shot $+$ ``:''$_{\rm 1\text{-}shot}^\text{random}$] are all the representation-level ablation models based on one-shot Random$_{\rm fixed}$, where the templates in these settings are all random strings, shown in Table~\ref{tab:textformatting_example}. 

The results in Table~\ref{tab:structural} show that in this setting, the model could still store the task-related information in the representations of the random string tokens, shown by the performance drop when removing their representations. There is nothing else for the model to recognize these random strings and store the information in their representations except that these tokens serve as delimiters to inform the model distinguishing the different parts of the prompt.


\begin{table}[t]

  \centering
  \resizebox{0.99\linewidth}{!}{
    \begin{tabular}{p{15.5cm}}
    \toprule
 
        \midrule
\multicolumn{1}{c}{Standard ICL}
    \\
    \midrule     
Classify the reviews into the categories of Positive and Negative. \textit{\textbf{[instruction]}} \\
\\
Review:~~\textbf{\textit{[delimiter: template]}}
\\
Peppered with witty dialogue and inventive moments.~~\textbf{\textit{[demonstration: content + stopword]}} \\
{Answer:}~~\textbf{\textit{[delimiter: template]}} \\
{Positive}~~\textbf{\textit{[label]}} \\
    \midrule
        \midrule
\multicolumn{1}{c}{One-shot Random$_{\rm fixed}$}
    \\
    \midrule     
Classify the reviews into the categories of Positive and Negative.~~\textbf{\textit{[instruction]}}  \\
\\
dsafjkldafdsajk:~~\textbf{\textit{[delimiter: random template 1]}} \\
Peppered with witty dialogue and inventive moments.~~\textbf{\textit{[demonstration]}} \\
reqwiorewsdafjl:~~\textbf{\textit{[delimiter: random template 2]}} \\
Positive~~\textbf{\textit{[label]}}  \\

    \midrule
       \midrule
    \multicolumn{1}{c}{Zero-shot$+$\textsc{temp}$_{\rm 1\text{-}shot}^\text{random}$}
    \\
    \midrule     
Classify the reviews into the categories of Positive and Negative.~~\textbf{\textit{[instruction]}}  \\
\\
dsafjkldafdsajk:~~\textbf{\textit{[delimiter: random template 1]}} \\
(m)(m)(m) ... (m)~~\textbf{\textit{[masked demonstration]}} \\
reqwiorewsdafjl:~~\textbf{\textit{[delimiter: random template 2]}} \\
Positive~~\textbf{\textit{[label]}}  \\

     \midrule
        \midrule
\multicolumn{1}{c}{Zero-shot$+$``:''$_{\rm 1\text{-}shot}^\text{random}$ }
    \\
    \midrule     
Classify the reviews into the categories of Positive and Negative.~~\textbf{\textit{[instruction]}}  \\
\\
(m):~~\textbf{\textit{[delimiter: random template 1]}} \\
(m)(m)(m) ... (m)~~\textbf{\textit{[masked demonstration]}} \\
(m):~~\textbf{\textit{[delimiter: random template 2]}} \\
Positive~~\textbf{\textit{[label]}}  \\

    \midrule


    \bottomrule
    \end{tabular}
    }
    \caption{An example, sampled from the SST2 dataset tested in our experiments, of the structural cue characteristic of performance-critical tokens and how they serve as delimiters of the text prompts, where (m) means that this token is masked.}
    \label{tab:textformatting_example}
\end{table}

\begin{table}[t]

  \centering
  \resizebox{0.99\linewidth}{!}{
    \begin{tabular}{lll}
    \toprule
    Datasets & Notations & Examples \\
    \midrule     
   \multicolumn{3}{c}{Random$_{\rm fixed}$}\\
   \midrule
    \multirow{2}*{\makecell[l]{CB \& RTE}}
    & ${\mathbf T}^{\rm in}$ & fdafdasjklfdadf: \{${\mathbf D}^{\rm inA}$\}\textbackslash n zcxvnmxcjkfdas: \{${\mathbf D}^{\rm inB}$\}\textbackslash n \\
    & ${\mathbf T}^{\rm out}$ & reqwiorewsdafjl: \{${\mathbf D}^{\rm out}$\}\textbackslash n\textbackslash n \\
    \midrule
    \multirow{2}*{\makecell[l]{Other tasks}} 
    & ${\mathbf T}^{\rm in}$ & dsafjkldafdsajk: \{${\mathbf D}^{\rm in}$\}\textbackslash n \\
    & ${\mathbf T}^{\rm out}$ & reqwiorewsdafjl: \{${\mathbf D}^{\rm out}$\}\textbackslash n\textbackslash n \\
   \midrule
   \multicolumn{3}{c}{Random$_{\rm nonfixed}$}\\
   \midrule
    \multirow{10}*{\makecell[l]{CB \& RTE}}
    & ${\mathbf T}^{\rm in}_1$ & fdafdasjklfdadf: \{${\mathbf D}^{\rm inA}$\}\textbackslash n zcxvnmxcjkfdas: \{${\mathbf D}^{\rm inB}$\}\textbackslash n \\
    & ${\mathbf T}^{\rm out}_1$ & xiadfjdsalgfweqrjl: \{${\mathbf D}^{\rm out}$\}\textbackslash n\textbackslash n \\
        & ${\mathbf T}^{\rm in}_2$ & gfhdajkgfhdasfj: \{${\mathbf D}^{\rm inA}$\}\textbackslash n cvxhlkdadsajfk: \{${\mathbf D}^{\rm inB}$\}\textbackslash n \\
    & ${\mathbf T}^{\rm out}_2$ & yufoufgaddavfdnsl: \{${\mathbf D}^{\rm out}$\}\textbackslash n\textbackslash n \\
        & ${\mathbf T}^{\rm in}_3$ & rrqetrizxcsdafq: \{${\mathbf D}^{\rm inA}$\}\textbackslash n vncmxasdgfadsl: \{${\mathbf D}^{\rm inB}$\}\textbackslash n \\
    & ${\mathbf T}^{\rm out}_3$ & afdgvcxjlzxnvxzla: \{${\mathbf D}^{\rm out}$\}\textbackslash n\textbackslash n \\
        & ${\mathbf T}^{\rm in}_4$ & mvfvxadfawewqro: \{${\mathbf D}^{\rm inA}$\}\textbackslash n lkajsdfopsadfp: \{${\mathbf D}^{\rm inB}$\}\textbackslash n \\
    & ${\mathbf T}^{\rm out}_4$ & fgsgfskjvcdafds: \{${\mathbf D}^{\rm out}$\}\textbackslash n\textbackslash n \\
            & ${\mathbf T}^{\rm in}_t$ & sdsajfjdsaczvvv: \{${\mathbf D}^{\rm inA}$\}\textbackslash n hkljfdiabasdfj: \{${\mathbf D}^{\rm inB}$\}\textbackslash n \\
    & ${\mathbf T}^{\rm out}_t$ & dafhglajfdvcaol: \{${\mathbf D}^{\rm out}$\}\textbackslash n\textbackslash n \\
    \midrule
    \multirow{10}*{\makecell[l]{Other tasks}} 
    & ${\mathbf T}^{\rm in}_1$ & dsafjkldaasdfjkl: \{${\mathbf D}^{\rm in}$\}\textbackslash n \\
    & ${\mathbf T}^{\rm out}_1$ & xiadfjdsalgfweqrjl: \{${\mathbf D}^{\rm out}$\}\textbackslash n\textbackslash n \\
    & ${\mathbf T}^{\rm in}_2$ & ewqroudajfsdafq: \{${\mathbf D}^{\rm in}$\}\textbackslash n \\
    & ${\mathbf T}^{\rm out}_2$ & yufoufgaddavfdnsl: \{${\mathbf D}^{\rm out}$\}\textbackslash n\textbackslash n \\
        & ${\mathbf T}^{\rm in}_3$ & eqdashcxzlreqguio: \{${\mathbf D}^{\rm in}$\}\textbackslash n \\
    & ${\mathbf T}^{\rm out}_3$ & afdgvcxjlzxnvxzla: \{${\mathbf D}^{\rm out}$\}\textbackslash n\textbackslash n \\
        & ${\mathbf T}^{\rm in}_4$ & cxzvadeqrczxdsa: \{${\mathbf D}^{\rm in}$\}\textbackslash n \\
    & ${\mathbf T}^{\rm out}_4$ & fgsgfskjvcdafds: \{${\mathbf D}^{\rm out}$\}\textbackslash n\textbackslash n \\
        & ${\mathbf T}^{\rm in}_t$ & vcxnkfgahvczxkl: \{${\mathbf D}^{\rm in}$\}\textbackslash n \\
    & ${\mathbf T}^{\rm out}_t$ & dafhglajfdvcaol: \{${\mathbf D}^{\rm out}$\}\textbackslash n\textbackslash n \\
       \midrule
  \multicolumn{3}{c}{Swap}\\
   \midrule
    \multirow{2}*{\makecell[l]{CB \& RTE}}
    & ${\mathbf T}^{\rm in}$ & Answer: \{${\mathbf D}^{\rm inA}$\}\textbackslash n Hypothesis: \{${\mathbf D}^{\rm inB}$\}\textbackslash n \\
    & ${\mathbf T}^{\rm out}$ & Premise: \{${\mathbf D}^{\rm out}$\}\textbackslash n\textbackslash n \\
    \bottomrule
    \end{tabular}
    }
\caption{Example \#1 of the ICL template used in all of our random experiments.}
    \label{tab:randomICLtemplate}
\end{table}

\begin{table}[htbp]

  \centering
  \resizebox{0.99\linewidth}{!}{
    \begin{tabular}{lll}
    \toprule
    Datasets & Notations & Examples \\
    \midrule     
   \multicolumn{3}{c}{Random$_{\rm fixed}$}\\
   \midrule
    \multirow{2}*{\makecell[l]{CB \& RTE}}
    & ${\mathbf T}^{\rm in}$ & eszycidpyopumzg: \{${\mathbf D}^{\rm inA}$\}\textbackslash n sgrlobvqgthjpwz: \{${\mathbf D}^{\rm inB}$\}\textbackslash n \\
    & ${\mathbf T}^{\rm out}$ & zbyygcrmzfnxlsu: \{${\mathbf D}^{\rm out}$\}\textbackslash n\textbackslash n \\
    \midrule
    \multirow{2}*{\makecell[l]{Other tasks}} 
    & ${\mathbf T}^{\rm in}$ & eszycidpyopumzg: \{${\mathbf D}^{\rm in}$\}\textbackslash n \\
    & ${\mathbf T}^{\rm out}$ & zbyygcrmzfnxlsu: \{${\mathbf D}^{\rm out}$\}\textbackslash n\textbackslash n \\
   \midrule
   \multicolumn{3}{c}{Random$_{\rm nonfixed}$}\\
   \midrule
    \multirow{10}*{\makecell[l]{CB \& RTE}}
    & ${\mathbf T}^{\rm in}_1$ & eszycidpyopumzg: \{${\mathbf D}^{\rm inA}$\}\textbackslash n sgrlobvqgthjpwz: \{${\mathbf D}^{\rm inB}$\}\textbackslash n \\
    & ${\mathbf T}^{\rm out}_1$ & zbyygcrmzfnxlsu: \{${\mathbf D}^{\rm out}$\}\textbackslash n\textbackslash n \\
        & ${\mathbf T}^{\rm in}_2$ & cwknayjkywwvpty: \{${\mathbf D}^{\rm inA}$\}\textbackslash n muzprouhvtidhqe: \{${\mathbf D}^{\rm inB}$\}\textbackslash n \\
    & ${\mathbf T}^{\rm out}_2$ & lnlgffeurextxme: \{${\mathbf D}^{\rm out}$\}\textbackslash n\textbackslash n \\
        & ${\mathbf T}^{\rm in}_3$ & pdnizszmpkfjzvo: \{${\mathbf D}^{\rm inA}$\}\textbackslash n ujulhuzkkqlfwkl: \{${\mathbf D}^{\rm inB}$\}\textbackslash n \\
    & ${\mathbf T}^{\rm out}_3$ & gflemobnbdjngii: \{${\mathbf D}^{\rm out}$\}\textbackslash n\textbackslash n \\
        & ${\mathbf T}^{\rm in}_4$ & gvsrxbdoxmpablo: \{${\mathbf D}^{\rm inA}$\}\textbackslash n ujulhuzkkqlfwkl: \{${\mathbf D}^{\rm inB}$\}\textbackslash n \\
    & ${\mathbf T}^{\rm out}_4$ & gflemobnbdjngii: \{${\mathbf D}^{\rm out}$\}\textbackslash n\textbackslash n \\
            & ${\mathbf T}^{\rm in}_t$ & gvsrxbdoxmpablo: \{${\mathbf D}^{\rm inA}$\}\textbackslash n xipddzrshrhprrb: \{${\mathbf D}^{\rm inB}$\}\textbackslash n \\
    & ${\mathbf T}^{\rm out}_t$ & npkxdzaipdpkbrs: \{${\mathbf D}^{\rm out}$\}\textbackslash n\textbackslash n \\
    \midrule
    \multirow{10}*{\makecell[l]{Other tasks}} 
    & ${\mathbf T}^{\rm in}_1$ & eszycidpyopumzg: \{${\mathbf D}^{\rm in}$\}\textbackslash n \\
    & ${\mathbf T}^{\rm out}_1$ & zbyygcrmzfnxlsu: \{${\mathbf D}^{\rm out}$\}\textbackslash n\textbackslash n \\
    & ${\mathbf T}^{\rm in}_2$ & cwknayjkywwvpty: \{${\mathbf D}^{\rm in}$\}\textbackslash n \\
    & ${\mathbf T}^{\rm out}_2$ & lnlgffeurextxme: \{${\mathbf D}^{\rm out}$\}\textbackslash n\textbackslash n \\
        & ${\mathbf T}^{\rm in}_3$ & pdnizszmpkfjzvo: \{${\mathbf D}^{\rm in}$\}\textbackslash n \\
    & ${\mathbf T}^{\rm out}_3$ & gflemobnbdjngii: \{${\mathbf D}^{\rm out}$\}\textbackslash n\textbackslash n \\
        & ${\mathbf T}^{\rm in}_4$ & gvsrxbdoxmpablo: \{${\mathbf D}^{\rm in}$\}\textbackslash n \\
    & ${\mathbf T}^{\rm out}_4$ & npkxdzaipdpkbrs: \{${\mathbf D}^{\rm out}$\}\textbackslash n\textbackslash n \\
        & ${\mathbf T}^{\rm in}_t$ & dgldzypdptzcekq: \{${\mathbf D}^{\rm in}$\}\textbackslash n \\
    & ${\mathbf T}^{\rm out}_t$ & xobxfpnzsfzipol: \{${\mathbf D}^{\rm out}$\}\textbackslash n\textbackslash n \\
    \bottomrule
    \end{tabular}
    }
\caption{Example \#2 of the ICL template used in all of our random experiments.}
    \label{tab:randomICLtemplate2}
\end{table}

\begin{table}[htbp]

  \centering
  \resizebox{0.99\linewidth}{!}{
    \begin{tabular}{lll}
    \toprule
    Datasets & Notations & Examples \\
    \midrule     
   \multicolumn{3}{c}{Random$_{\rm fixed}$}\\
   \midrule
    \multirow{2}*{\makecell[l]{CB \& RTE}}
    & ${\mathbf T}^{\rm in}$ & bcclfxzvjitgtbs: \{${\mathbf D}^{\rm inA}$\}\textbackslash n evtlfrwvtfmjtns: \{${\mathbf D}^{\rm inB}$\}\textbackslash n \\
    & ${\mathbf T}^{\rm out}$ & qtnheeipeustcwn: \{${\mathbf D}^{\rm out}$\}\textbackslash n\textbackslash n \\
    \midrule
    \multirow{2}*{\makecell[l]{Other tasks}} 
    & ${\mathbf T}^{\rm in}$ & bcclfxzvjitgtbs: \{${\mathbf D}^{\rm in}$\}\textbackslash n \\
    & ${\mathbf T}^{\rm out}$ & qtnheeipeustcwn: \{${\mathbf D}^{\rm out}$\}\textbackslash n\textbackslash n \\
   \midrule
   \multicolumn{3}{c}{Random$_{\rm nonfixed}$}\\
   \midrule
    \multirow{10}*{\makecell[l]{CB \& RTE}}
    & ${\mathbf T}^{\rm in}_1$ & bcclfxzvjitgtbs: \{${\mathbf D}^{\rm inA}$\}\textbackslash n evtlfrwvtfmjtns: \{${\mathbf D}^{\rm inB}$\}\textbackslash n \\
    & ${\mathbf T}^{\rm out}_1$ & qtnheeipeustcwn: \{${\mathbf D}^{\rm out}$\}\textbackslash n\textbackslash n \\
        & ${\mathbf T}^{\rm in}_2$ & ymupnggvmbnoobq: \{${\mathbf D}^{\rm inA}$\}\textbackslash n rrrnpgbmmgqymky: \{${\mathbf D}^{\rm inB}$\}\textbackslash n \\
    & ${\mathbf T}^{\rm out}_2$ & xleuwtyqnnfgzjx: \{${\mathbf D}^{\rm out}$\}\textbackslash n\textbackslash n \\
        & ${\mathbf T}^{\rm in}_3$ & pdnizszmpkfjzvo: \{${\mathbf D}^{\rm inA}$\}\textbackslash n qlfulxzxwfnwbum: \{${\mathbf D}^{\rm inB}$\}\textbackslash n \\
    & ${\mathbf T}^{\rm out}_3$ & jpnvgbnjjlawqfo: \{${\mathbf D}^{\rm out}$\}\textbackslash n\textbackslash n \\
        & ${\mathbf T}^{\rm in}_4$ & mfkqxjoxtpmzdrs: \{${\mathbf D}^{\rm inA}$\}\textbackslash n yyzdeayigwzjosn: \{${\mathbf D}^{\rm inB}$\}\textbackslash n \\
    & ${\mathbf T}^{\rm out}_4$ & pdsqooqrhvydszp: \{${\mathbf D}^{\rm out}$\}\textbackslash n\textbackslash n \\
            & ${\mathbf T}^{\rm in}_t$ & rerlkjfvlvyzpmc: \{${\mathbf D}^{\rm inA}$\}\textbackslash n iuumpcsevursgqe: \{${\mathbf D}^{\rm inB}$\}\textbackslash n \\
    & ${\mathbf T}^{\rm out}_t$ & tuaqblysbipihsv: \{${\mathbf D}^{\rm out}$\}\textbackslash n\textbackslash n \\
    \midrule
    \multirow{10}*{\makecell[l]{Other tasks}} 
    & ${\mathbf T}^{\rm in}_1$ & bcclfxzvjitgtbs: \{${\mathbf D}^{\rm in}$\}\textbackslash n \\
    & ${\mathbf T}^{\rm out}_1$ & qtnheeipeustcwn: \{${\mathbf D}^{\rm out}$\}\textbackslash n\textbackslash n \\
    & ${\mathbf T}^{\rm in}_2$ & ymupnggvmbnoobq: \{${\mathbf D}^{\rm in}$\}\textbackslash n \\
    & ${\mathbf T}^{\rm out}_2$ & xleuwtyqnnfgzjx: \{${\mathbf D}^{\rm out}$\}\textbackslash n\textbackslash n \\
        & ${\mathbf T}^{\rm in}_3$ & pdwunmjronsmuvu: \{${\mathbf D}^{\rm in}$\}\textbackslash n \\
    & ${\mathbf T}^{\rm out}_3$ & jpnvgbnjjlawqfo: \{${\mathbf D}^{\rm out}$\}\textbackslash n\textbackslash n \\
        & ${\mathbf T}^{\rm in}_4$ & mfkqxjoxtpmzdrs: \{${\mathbf D}^{\rm in}$\}\textbackslash n \\
    & ${\mathbf T}^{\rm out}_4$ & pdsqooqrhvydszp: \{${\mathbf D}^{\rm out}$\}\textbackslash n\textbackslash n \\
        & ${\mathbf T}^{\rm in}_t$ & rerlkjfvlvyzpmc: \{${\mathbf D}^{\rm in}$\}\textbackslash n \\
    & ${\mathbf T}^{\rm out}_t$ & tuaqblysbipihsv: \{${\mathbf D}^{\rm out}$\}\textbackslash n\textbackslash n \\
    \bottomrule
    \end{tabular}
    }
\caption{Example \#3 of the ICL template used in all of our random experiments.}
    \label{tab:randomICLtemplate3}
\end{table}
\begin{table}[htbp]

  \centering
  \resizebox{0.99\linewidth}{!}{
    \begin{tabular}{lll}
    \toprule
    Datasets & Notations & Examples \\
    \midrule     
   \multicolumn{3}{c}{Random$_{\rm fixed}$}\\
   \midrule
    \multirow{2}*{\makecell[l]{CB \& RTE}}
    & ${\mathbf T}^{\rm in}$ & hsreltpusctapir: \{${\mathbf D}^{\rm inA}$\}\textbackslash n woxwxgwctxdumok: \{${\mathbf D}^{\rm inB}$\}\textbackslash n \\
    & ${\mathbf T}^{\rm out}$ & prlhxooromawkcp: \{${\mathbf D}^{\rm out}$\}\textbackslash n\textbackslash n \\
    \midrule
    \multirow{2}*{\makecell[l]{Other tasks}} 
    & ${\mathbf T}^{\rm in}$ & hsreltpusctapir: \{${\mathbf D}^{\rm in}$\}\textbackslash n \\
    & ${\mathbf T}^{\rm out}$ & prlhxooromawkcp: \{${\mathbf D}^{\rm out}$\}\textbackslash n\textbackslash n \\
   \midrule
   \multicolumn{3}{c}{Random$_{\rm nonfixed}$}\\
   \midrule
    \multirow{10}*{\makecell[l]{CB \& RTE}}
    & ${\mathbf T}^{\rm in}_1$ & hsreltpusctapir: \{${\mathbf D}^{\rm inA}$\}\textbackslash n woxwxgwctxdumok: \{${\mathbf D}^{\rm inB}$\}\textbackslash n \\
    & ${\mathbf T}^{\rm out}_1$ & prlhxooromawkcp: \{${\mathbf D}^{\rm out}$\}\textbackslash n\textbackslash n \\
        & ${\mathbf T}^{\rm in}_2$ & cbptgaytithxayh: \{${\mathbf D}^{\rm inA}$\}\textbackslash n bhxgcstisqmfnpz: \{${\mathbf D}^{\rm inB}$\}\textbackslash n \\
    & ${\mathbf T}^{\rm out}_2$ & mvpvoeuvgczfemz: \{${\mathbf D}^{\rm out}$\}\textbackslash n\textbackslash n \\
        & ${\mathbf T}^{\rm in}_3$ & htkbzfizxwpeqrm: \{${\mathbf D}^{\rm inA}$\}\textbackslash n felxgmjeuabznwd: \{${\mathbf D}^{\rm inB}$\}\textbackslash n \\
    & ${\mathbf T}^{\rm out}_3$ & glfwilpyrwnsujg: \{${\mathbf D}^{\rm out}$\}\textbackslash n\textbackslash n \\
        & ${\mathbf T}^{\rm in}_4$ & frskoasvqybxcob: \{${\mathbf D}^{\rm inA}$\}\textbackslash n bkepuhnckdaqmhx: \{${\mathbf D}^{\rm inB}$\}\textbackslash n \\
    & ${\mathbf T}^{\rm out}_4$ & ljttiywadveyzah: \{${\mathbf D}^{\rm out}$\}\textbackslash n\textbackslash n \\
            & ${\mathbf T}^{\rm in}_t$ & dfpqndhxehhtser: \{${\mathbf D}^{\rm inA}$\}\textbackslash n bvucjofrggmmcsh: \{${\mathbf D}^{\rm inB}$\}\textbackslash n \\
    & ${\mathbf T}^{\rm out}_t$ & koesxfmmjjjjvmp: \{${\mathbf D}^{\rm out}$\}\textbackslash n\textbackslash n \\
    \midrule
    \multirow{10}*{\makecell[l]{Other tasks}} 
    & ${\mathbf T}^{\rm in}_1$ & hsreltpusctapir: \{${\mathbf D}^{\rm in}$\}\textbackslash n \\
    & ${\mathbf T}^{\rm out}_1$ & prlhxooromawkcp: \{${\mathbf D}^{\rm out}$\}\textbackslash n\textbackslash n \\
    & ${\mathbf T}^{\rm in}_2$ & cbptgaytithxayh: \{${\mathbf D}^{\rm in}$\}\textbackslash n \\
    & ${\mathbf T}^{\rm out}_2$ & mvpvoeuvgczfemz: \{${\mathbf D}^{\rm out}$\}\textbackslash n\textbackslash n \\
        & ${\mathbf T}^{\rm in}_3$ & htkbzfizxwpeqrm: \{${\mathbf D}^{\rm in}$\}\textbackslash n \\
    & ${\mathbf T}^{\rm out}_3$ & glfwilpyrwnsujg: \{${\mathbf D}^{\rm out}$\}\textbackslash n\textbackslash n \\
        & ${\mathbf T}^{\rm in}_4$ & frskoasvqybxcob: \{${\mathbf D}^{\rm in}$\}\textbackslash n \\
    & ${\mathbf T}^{\rm out}_4$ & ljttiywadveyzah: \{${\mathbf D}^{\rm out}$\}\textbackslash n\textbackslash n \\
        & ${\mathbf T}^{\rm in}_t$ & dfpqndhxehhtser: \{${\mathbf D}^{\rm in}$\}\textbackslash n \\
    & ${\mathbf T}^{\rm out}_t$ & koesxfmmjjjjvmp: \{${\mathbf D}^{\rm out}$\}\textbackslash n\textbackslash n \\
    \bottomrule
    \end{tabular}
    }
\caption{Example \#4 of the ICL template used in all of our random experiments.}
    \label{tab:randomICLtemplate4}
\end{table}
\begin{table}[htbp]

  \centering
  \resizebox{0.99\linewidth}{!}{
    \begin{tabular}{lll}
    \toprule
    Datasets & Notations & Examples \\
    \midrule     
   \multicolumn{3}{c}{Random$_{\rm fixed}$}\\
   \midrule
    \multirow{2}*{\makecell[l]{CB \& RTE}}
    & ${\mathbf T}^{\rm in}$ & hjdxmpeccamrjzy: \{${\mathbf D}^{\rm inA}$\}\textbackslash n agxyhmkawezafde: \{${\mathbf D}^{\rm inB}$\}\textbackslash n \\
    & ${\mathbf T}^{\rm out}$ & ndxtrwvqugyygku: \{${\mathbf D}^{\rm out}$\}\textbackslash n\textbackslash n \\
    \midrule
    \multirow{2}*{\makecell[l]{Other tasks}} 
    & ${\mathbf T}^{\rm in}$ & hjdxmpeccamrjzy: \{${\mathbf D}^{\rm in}$\}\textbackslash n \\
    & ${\mathbf T}^{\rm out}$ & ndxtrwvqugyygku: \{${\mathbf D}^{\rm out}$\}\textbackslash n\textbackslash n \\
   \midrule
   \multicolumn{3}{c}{Random$_{\rm nonfixed}$}\\
   \midrule
    \multirow{10}*{\makecell[l]{CB \& RTE}}
    & ${\mathbf T}^{\rm in}_1$ & hjdxmpeccamrjzy: \{${\mathbf D}^{\rm inA}$\}\textbackslash n agxyhmkawezafde: \{${\mathbf D}^{\rm inB}$\}\textbackslash n \\
    & ${\mathbf T}^{\rm out}_1$ & ndxtrwvqugyygku: \{${\mathbf D}^{\rm out}$\}\textbackslash n\textbackslash n \\
        & ${\mathbf T}^{\rm in}_2$ & mcsgenpkdwsfknc: \{${\mathbf D}^{\rm inA}$\}\textbackslash n egnqobhzvxjhsxh: \{${\mathbf D}^{\rm inB}$\}\textbackslash n \\
    & ${\mathbf T}^{\rm out}_2$ & ijkdikcmiskofsg: \{${\mathbf D}^{\rm out}$\}\textbackslash n\textbackslash n \\
        & ${\mathbf T}^{\rm in}_3$ & cmaqcvtdkemdauv: \{${\mathbf D}^{\rm inA}$\}\textbackslash n oslzaygbefxlwqt: \{${\mathbf D}^{\rm inB}$\}\textbackslash n \\
    & ${\mathbf T}^{\rm out}_3$ & mumrjhndwmidwmj: \{${\mathbf D}^{\rm out}$\}\textbackslash n\textbackslash n \\
        & ${\mathbf T}^{\rm in}_4$ & cgmylzvslxmojvq: \{${\mathbf D}^{\rm inA}$\}\textbackslash n tlwxsjmnfkolffl: \{${\mathbf D}^{\rm inB}$\}\textbackslash n \\
    & ${\mathbf T}^{\rm out}_4$ & mitaowjyibjwwol: \{${\mathbf D}^{\rm out}$\}\textbackslash n\textbackslash n \\
            & ${\mathbf T}^{\rm in}_t$ & pvockachyflybtk: \{${\mathbf D}^{\rm inA}$\}\textbackslash n wtjqmtwxbnpyqbp: \{${\mathbf D}^{\rm inB}$\}\textbackslash n \\
    & ${\mathbf T}^{\rm out}_t$ & ydediotfezhfnbx: \{${\mathbf D}^{\rm out}$\}\textbackslash n\textbackslash n \\
    \midrule
    \multirow{10}*{\makecell[l]{Other tasks}} 
    & ${\mathbf T}^{\rm in}_1$ & hsreltpusctapir: \{${\mathbf D}^{\rm in}$\}\textbackslash n \\
    & ${\mathbf T}^{\rm out}_1$ & prlhxooromawkcp: \{${\mathbf D}^{\rm out}$\}\textbackslash n\textbackslash n \\
    & ${\mathbf T}^{\rm in}_2$ & cbptgaytithxayh: \{${\mathbf D}^{\rm in}$\}\textbackslash n \\
    & ${\mathbf T}^{\rm out}_2$ & mvpvoeuvgczfemz: \{${\mathbf D}^{\rm out}$\}\textbackslash n\textbackslash n \\
        & ${\mathbf T}^{\rm in}_3$ & htkbzfizxwpeqrm: \{${\mathbf D}^{\rm in}$\}\textbackslash n \\
    & ${\mathbf T}^{\rm out}_3$ & glfwilpyrwnsujg: \{${\mathbf D}^{\rm out}$\}\textbackslash n\textbackslash n \\
        & ${\mathbf T}^{\rm in}_4$ & frskoasvqybxcob: \{${\mathbf D}^{\rm in}$\}\textbackslash n \\
    & ${\mathbf T}^{\rm out}_4$ & ljttiywadveyzah: \{${\mathbf D}^{\rm out}$\}\textbackslash n\textbackslash n \\
        & ${\mathbf T}^{\rm in}_t$ & dfpqndhxehhtser: \{${\mathbf D}^{\rm in}$\}\textbackslash n \\
    & ${\mathbf T}^{\rm out}_t$ & koesxfmmjjjjvmp: \{${\mathbf D}^{\rm out}$\}\textbackslash n\textbackslash n \\
    \bottomrule
    \end{tabular}
    }
\caption{Example \#5 of the ICL template used in all of our random experiments.}
    \label{tab:randomICLtemplate5}
\end{table}

\end{document}